\documentclass[11pt]{article}

\usepackage{fullpage}
\usepackage{graphicx}
\usepackage{amsfonts}
\usepackage{booktabs}
\usepackage{float}
\usepackage{amsmath, amssymb, latexsym, amscd, amsthm, epsfig}
\usepackage{multirow}
\usepackage{array}
\usepackage{tabu}
\usepackage{verbatim}
\usepackage[table]{xcolor}
\usepackage{adjustbox}
\usepackage{subcaption}
\usepackage{makeidx}
\usepackage{tikz}
 \usepackage{hyperref}

\title{\textbf{Collapsing of dimensionality} }
\author{     \vspace{0.5cm}   \\  { Marco Gori, Marco Maggini, Alessandro Rossi}   \\  \vspace{0.3cm} \\ {  Department of Information Engineering and Mathematical Sciences, } \\ \vspace{0.1cm} \\ {University of Siena, Italy  }           }
\date{July 1, 2015}

\def\A{\mathbf A}
\def\B{\mathbf B}

\def\E{\mathbf E}

\def\R{\mathbf{R}}
\def\M{\mathbf M}
\def\N{\mathbf N}

\def\a{\mathbf a}
\def\a{\mathbf b}

\def\f{\mathbf f}
\def\g{\mathbf g}

\def\tt{\mathbf t}

\def\x{\mathbf x}
\def\y{\mathbf y}
\def\n{\mathbf{\eta}}

\def\cG{\mathcal G}
\def\cI{\mathcal I}

\def\cK{\mathcal K}
\def\cL{\mathcal L}

\def\cN{\mathcal N}

\def\cS{\mathcal S}

\def\cU{\mathcal U}
\def\cV{\mathcal V}

\def\bbN{\mathbb{N}}
\def\bbR{\mathbb{R}}

\def\Rg{R_G}	
\def\Eg{E_G}	
\def\rg{r}		

\def\by{\bar{f}}

\def\inn{\! \in \!}

\def\ug{\! = \!}
\newcommand{\norm}[1]{\left\lVert#1\right\rVert}

\def\a{\alpha}
\def\b{\beta}
\def\ee{\epsilon}

\def\g{\gamma}
\def\l{\lambda}

\def\q{\tau}
\def\t{\theta}

\def\RefFig#1{Figure~\ref{#1}}
\def\RefTab#1{Table~\ref{#1}}

\def\RefSec#1{Section~\ref{#1}}

\begin{document}

\maketitle

\vskip3cm

\tableofcontents

\newpage

\section{Introduction}

We study the behavior of function $f$ which work on a spatio-temporal domain. 
We try to drop the direct dependence on the spatial input space.
Starting from the work of~\cite{papini}, 
we can substitute the research of the function $f(t,x(t))$ with the research of $\by \ug f \circ \pi $, 
where $\pi$ is a map such that $\pi(t) \ug (t,x(t))$.
Indeed, find $f$ as a stationary point of the functional:
\begin{equation}\label{PHI}
\Phi(f) = \l \int_{\mathcal{M}} (Pf)^2 \psi ds + \mu \int_{\mathcal{M}} f^2 \psi ds +  \sum_{i \ug 1}^N \psi (p_i) |  f(p_i) - y_i |^2
\end{equation}
where $p_i$ are points in the domain $\mathcal{M}$ of $f$,
is equivalent to find $\by$ as stationary point of:
\begin{equation}\label{PHI2}
\bar{\Phi}(\by) = \l \int_{0}^T (P \by)^2 \psi \sqrt{1 + |x(t)'|^2} dt + \mu \int_{0}^T {\by}^2 \psi \sqrt{1 + |x(t)'|^2} dt +  \sum_{i \ug 1}^N \psi (t_i) | \by(t_i) - y_i |^2
\end{equation}
where we pose $\psi \ug \psi \circ \pi$.
By the application of the Eulero-Lagrange equation, derived in the framework of distribution, we obtain the condition:
\begin{equation}\label{first}
\l \, P^* (\psi \, \sqrt{1 + |x(t)'|^2} \,  P \by) + \mu \, \psi  \, \sqrt{1 + |x(t)'|^2} \,  \by + \sum_{i \ug 1}^N \psi (t_i) \left[ \by (t_i) - y_i \right] \delta (t-t_i) = 0 
\end{equation}
we pose $b \ug b(x(t)) \ug \sqrt{1 + |x(t)'|^2} $ and $ U \ug U(t) \ug \sum_{i \ug 1}^N \psi (t_i) \left[ \by (t_i) - y_i \right] \delta (t-t_i)$. 
We want to analyze the behavior of (\ref{first}) when $P$ is a linear differential operator and $\psi (t)$ is a monotonically positive increasing function. 
The external supervisions are provided at discrete time instants $t_i, i \in \N$ by specifying a target value $y_{i}$ for the function $\by$ to be learned.  
The supervised pairs $(x(t_i),y_i)$ are collected into the set $\cL = \left\{(x(t_i),y_i)\right\}_{i \in \N}$ 
and data are supposed to be cyclical in time with period $T$.
To deal with a manageable resolution of (\ref{first}) we choose $\psi (t) \ug \frac{e^{\t t}}{\sqrt{1 + |x(t)'|^2}} \ug \frac{\varphi}{b}$ , $ \varphi \ug \varphi(t) \ug e^{\t t}$ .
 We can now concisely write (\ref{first}) as:

\begin{equation}\label{diss}
 P^* (\varphi \,  P \by) + \frac{\mu}{\l}  \, \varphi \,  \by + \frac{1}{\l} U = 0 
\end{equation}

\subsection{First Order Operator}\label{FOT}
When $P \ug \a_0 + \a_1 D$, where $D \ug \frac{d}{dt}$, we have $P^* \ug \a_0 - \a_1 D$ and then (\ref{first}) became:
$$
\begin{array}{c}
 \displaystyle  \left[ \a_0 - \a_1 D \right] \left(\varphi \, \left[ \a_0 + \a_1 D\right] \by \right) + \frac{\mu}{\l}\,  \varphi \, \by + \frac{1}{\l} U  =  \\
  \vspace{2pt} \\
   \displaystyle  \left[ \a_0 - \a_1 D \right] \left( \a_0 \varphi \by + \a_1 \varphi  D \by \right) + \frac{\mu}{\l}\,  \varphi \, \by + \frac{1}{\l} U  =  \\
  \vspace{2pt} \\
 \displaystyle   \a_0^2 \varphi \by  + \a_0 \a_1 \varphi  D \by  - \a_0 \a_1 D \left[ \varphi \by \right] - \a_1^2 D \left[ \varphi D\by \right] + \frac{\mu}{\l}\,  \varphi \, \by + \frac{1}{\l} U  =  \\
  \vspace{2pt} \\ 
   \displaystyle \a_0^2 \varphi \by  + \a_0 \a_1 \varphi  D \by  - \a_0 \a_1\dot{\varphi} \by - \a_0 \a_1  \varphi D\by - \a_1^2 \dot{\varphi} D\by  - \a_1^2 \varphi D^2\by + \frac{\mu}{\l}\,  \varphi \, \by + \frac{1}{\l} U  =  \\
\end{array}
$$
if we divide by $-\a_1^2 \varphi $ and considering that $\dot{\varphi} = \t e^{\t t} = \t \varphi$ we have:
\begin{equation}\label{dissFO}
D^2 \by \, + \, \t D \by \,  + \, \left[\frac{ \a_0 \a_1 \t - \a_0^2 - \mu / \l}{\a_1^2} \right] \by \, - \, \frac{1}{\l \a_1^2 \varphi } U= 0 
\end{equation}
We can pose:
$$
\b =  \frac{ \a_0 \a_1 \t - \a_0^2 - \mu / \l}{\a_1^2}
$$
and since $\mu \inn \{0\,,1\}$, when $\mu \ug 0 $ we obtain an equation analogous to the cases analyzed in \cite{report}.

We can rewrite (\ref{dissFO}) as a system of two first order linear differential equations:
$$
\dot{\f}(t) = \A \f(t) + \B F(t)
$$
where $ F(t) \ug   \frac{1}{\l \a_1^2 \varphi } \sum_{i \ug 1}^N \psi (t_i) \left[ \by (t_i) - y_i \right] \delta (t-t_i)$ and

$$
\A =
\left[
\begin{array}{cc}
   0      &     1     \\
- \b & - \t \\
\end{array}
\right] \, ,
\quad
\B = \left[
\begin{array}{r} 0 \\ 1 \\ 
\end{array} 
\right]\, ,
\quad
\f = \left[
\begin{array}{c} \by \\  D\by \\ 
\end{array}
\right] .
\medskip
$$
By using the Lagrange formula we have
\begin{equation}
\medskip
\f (t) \ug e^{\A(t-t_0)} \f (t_0) + \int_{t_0}^t e^{\A(t-s)} \cdot \B F(s) ds .
\medskip
\end{equation}
When we consider an equally spaced discretization of time $t_k \ug \q K$ and we assume $t_0 \ug 0$, the general evolution of the system can be computed by
\begin{equation}
\medskip
\f [ K] = \f(\q K)= e^{\A \tau K} \f [0] + \int_{0}^{\q K} e^{\A(\q K-s)} \cdot \B F(s) ds
\medskip
\end{equation}
and at the next step we have 
\begin{eqnarray*}
\medskip
\f [K+1]  & = & e^{\A \tau (K+1)} \f [0] + \int_{0}^{\q (K+1)} e^{\A(\q (K+1)-s)} \cdot \B F(s) ds  \\
              & = & e^{\A \q} \left( e^{\A \tau K} \f [0]  + \int_{0}^{\q K} e^{\A(\q K-s)} \cdot \B F(s) ds \right) \\
              &    & + \int_{\q K}^{\q (K+1)} e^{\A [\q (K+1)-s]} \cdot \B F(s) ds \\
              & = & e^{\A \q} \f [ K]  + \int_{\q K}^{\q (K+1)} e^{\A [ \q (K+1)-s]} \cdot \B F(s) ds \\
              & = & e^{\A \q} \f [ K]  + \int_{\q K}^{\q (K+1)} \left( e^{\A [ \q (K+1)-s]} \cdot \B \frac{1}{\l \a_1^2 \varphi(s) } \sum_{i \ug 1}^N \psi (t_i) \left[ \by (t_i) - y_i \right] \delta (s-t_i) \right) ds \\              
              \medskip
\end{eqnarray*}
Since we can arbitrarily assume that $t_i \ug (i -1)\q + (\q / 2)$, if we pose $t_{\bar{i}}= K \q +  (\q / 2)$ we have:
\begin{eqnarray}
\medskip
\f [K+1]  & = & e^{\A \q} \f [ K]  + e^{\A [ \q (K+1)-t_{\bar{i}} ]} \cdot \B \frac{1}{\l \a_1^2 \varphi(t_{\bar{i}}) } \frac{\varphi(t_{\bar{i}})}{b(t_{\bar{i}})} \left[ \by (t_{\bar{i}}) - y_{\bar{i}} \right]    \nonumber\\
              & = &   e^{\A \q} \f [ K]  + e^{\A \q /2} \cdot \B \frac{1}{\l \a_1^2 b(t_{\bar{i}}) }\left[ \by (t_{\bar{i}}) - y_{\bar{i}} \right]   \label{UpFoFO}        
              \medskip
\end{eqnarray}

\subsection{Second Order Operator}
When $P \ug \a_0 + \a_1 D +\a_2 D^2 $, where $D \ug \frac{d}{dt}$, we have $P^* \ug \a_0 - \a_1 D +\a_2 D^2 $ and from \emph{Eulero-Lagrange} equation we can write a fourth order linear differential equation:

\begin{equation}\label{dissSO}
D^4 \by + \b_3 D^3 \by + \b_2 D^2 \by + \b_1 D \by + \b_0  \by + \, \frac{1}{\l \a_2^2 \varphi } U =0 
\end{equation}
where:
 \begin{equation}\label{coeff4}
 \begin{array}{rcl}
 \b_0 \! & \ug & \frac{\a_0 \a_2 \t^2 - \a_0 \a_1 \t  + \a_0 ^2 + \mu / \l}{a_2^2}  \\
  \b_1 \! & \ug & \frac{\a_1 \a_2 \t^2 + (2 \a_0 \a_2 - \a_1^2)\t}{\a_2^2} \\
 \b_2 \! & \ug & \frac{\a_2^2 \t^2 + \a_1 \a_2 \t + 2 \a_0 \a_2 - \a_1^2}{\a_2^2}\\
 \b_3 \! & \ug & 2\t 
 \end{array}
 \end{equation}
which replicate the case of \cite{report} when $\mu \ug 0$. By transforming again the (\ref{dissSO}) in a system we find the analogous of (\ref{UpFoFO}) :
\begin{equation} \label{UpFoSO} 
\medskip
\f [K+1]  =   e^{\A \q} \f [ K]  + e^{\A [ \q /2]} \cdot \B \frac{1}{\l \a_2^2 b(t_{\bar{i}}) }\left[ \by (t_{\bar{i}}) - y_{\bar{i}} \right]        
\medskip
\end{equation}
\medskip
$$
\A =
\left[
\begin{array}{cccc}
   0      &     1     &     0    &      0   \\
  0      &     0     &     1    &      0   \\
  0      &     0     &     0    &      1   \\
- \b_0 & - \b_1 & - \b_2 & - \b_3 \\
\end{array}
\right] \, ,
\quad
\B = \left[
\begin{array}{r} 0 \\ 0 \\ 0 \\ -1 \\ 
\end{array}
\right].
\medskip
$$
\section{Implementation and initial experiments}\label{implementation}

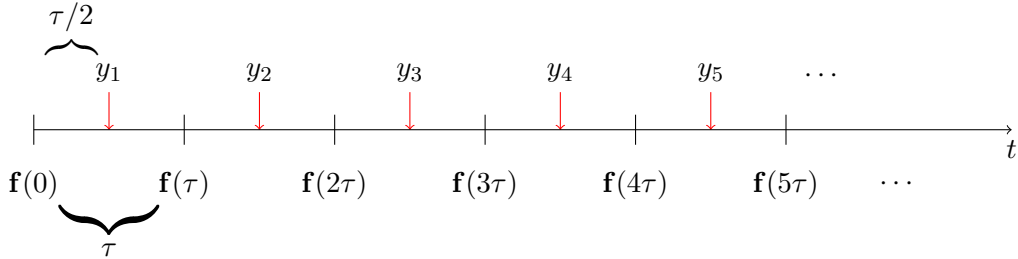
\begin{figure}[h]
\hspace{1cm}
\begin{tikzpicture}

       \foreach \i in {0,...,5} { 
       		\draw (0+2*\i,-.2) -- (0+2*\i, .2);
		 }
		 
	\node [below] at (0,-.4) {$\f(0)$};
	\node [below] at (2,-.4) {$\f(\q)$};
	\node [below] at (11.5,-.5) {$\cdots$};
	
             \foreach \i in {2,...,5} { 
		\node [below] at (0+2*\i,-.4) {$\f(\i \q)$};
		 }
       
       \foreach \i in {0,...,4} { 
       		\draw[red,->] (1+2*\i,0.5) -- (1+2*\i, 0);
		 }
		 
	\foreach \i in {1,...,5} { 
		\node [above] at (-1+2*\i,0.5) {$y_{\i}$};
		 }
	\node [above] at (10.5,.5) {$\cdots$};
		 
       \draw[->] (0,0) -- (13,0);
       
       \node  [below] at (13,0) {$t$};

	\node  at (1,-1.2) {\huge $\underbrace{ }$};
	\node  at (1,-1.6) {$\q$};
	
	\node  at (0.5,1.1) { $\overbrace{ }$};
	\node  at (0.5,1.5) {$\q /2$};
	
\end{tikzpicture}
\caption{Distribution of updating and instants of supervision in time.}\label{data}
\end{figure}

In section \ref{FOT}  we arbitrarily assume that the supervision instants fall in the middle of two updating of the function $\f$. We can see this scheme in Fig.\ref{data}.

Where the initial conditions $\f [0]$ are given. At each step $K\!+\!1$ we need the derivative of the error function 
$E(t) \ug \frac{1}{2} {\left( \by(t) - y \right)}^2$ w.r.t time  
 $\delta E \ug \frac{dE(t)}{dt} \ug [\by(t_{\bar{i}}) - y_{\bar{i}}]$ ,
 evaluated at the supervision instant
  $t_{\bar{i}} \ug \q K + \q/2$ .
  Since the supervisions come at a different moment of the updates of $\by$ (with gap $\q /2$), if we use 
  $\delta E \ug  \left[\f_1[K] - y_{\bar{i}} \right]$ 
  (where $ \f_1[K] \ug \by(\q K) $ ) 
  we commit a little approximation error. 
  To avoid this error we can perform an intermediate updating by calculating the value of the function $\by$ at the point  
  $t_{\bar{i}}$ as 
  $\mathbf{\tilde{f}}[K\!+\!1] = e^{\A \q /2} \: \f[K] $, 
  then evaluate exactly the derivative as 
  $\delta E \ug \left[\mathbf{\tilde{f}}_1[K\!+\!1] - y_{\bar{i}}\right]$, 
  where 
  $\mathbf{\tilde{f}}_1[K\!+\!1]  \ug \by( t_{\bar{i}} )$.

\begin{figure}[h]
\hspace{4cm}
\begin{tikzpicture}

       \draw[->] (0,0) -- (8,0);
        \node  [below] at (8,0) {$t$};
        
	\foreach \i in {0,...,1} { 
       		\draw (4*\i+2,-.2) -- (4*\i+2, .2);
		 }

	\node at (2,-.6) {$\f[K]$};
	 \draw[fill] (4,0) circle [radius=0.05];
	\node  at (4,-.6) {$\mathbf{\tilde{f}}_1[K\!+\!1]$};
	\draw[red,->] (4,0.5) -- (4, 0.1);
	\node  at (4,.6) {$y_{\bar{i}}$};
	\node  at (6,-.6) {$\f[K\!+\!1]$};
	\node  at (5,.3) {{\small $\q /2$}};
	\node  at (3,.3) {{\small $\q /2$}};
	
\end{tikzpicture}
\caption{Distribution of intermediate updating to avoid approximation.}\label{data2}
\end{figure}
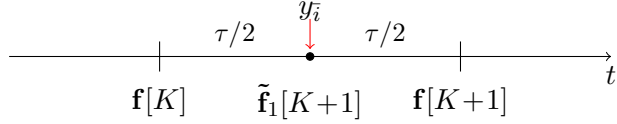

\subsection{1-Dimension artificial function}

%
%

\begin{figure}[H]
\hspace{3cm}
\includegraphics[scale=0.5]{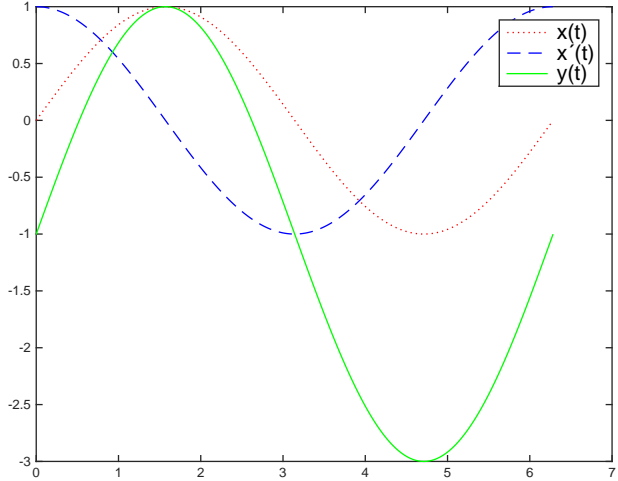}
\caption{Graphs of the input function $x(t)$ with its derivative $x'(t)$ and the target $y(t)$ in the period $\left[ \, 0 \,, \: T \ug 2 \pi \, \right] $.}\label{InpuFunc}
\end{figure}

We start with some experiments when a 1-Dimension input comes from the simple periodic function $x(t)\ug \sin t $. 
At this input we assign the target $y(t)= 2\,x(t)-1$, as we can see in Fig.\ref{InpuFunc}.

We can generate data by choosing the time step $\q$ and then find the total number of supervision as $N \ug T/ \q$.
 Because of the assumption on the distribution trough time of $\f$ and $y_i$, we produce a sampling of $T$ by taking $t_1 \ug \q /2\, , \; t_2 \ug \q \!+\! \q /2 \, , ... , t_{{\small K+1}} \ug \q K \!+\! \q /2, ... $ and obtain a vector
$$
\tt = \left[ \, t_1 \, \cdots \, t_N \, \right] \; , \quad t_N \leq T
$$
\begin{figure}[h]
\hspace{4cm}
\begin{tikzpicture}

       \draw[->] (0,0) -- (9,0);
        \node  [below] at (9,0) {$t$};

        \draw (0,-.2) -- (0, .2);
        \node  [below] at (0,-0.3) {$0$};
        
        \draw[fill] (1,0) circle [radius=0.05];
        \node  [below] at (1,-0.3) {$t_1$};
        
        \draw[fill] (3,0) circle [radius=0.05];
        \node  [below] at (3,-0.3) {$t_2$};
        
        \draw[fill] (7,0) circle [radius=0.05];
        \node  [below] at (7,-0.3) {$t_N$};
        
        \draw (8,-.2) -- (8, .2);
        \node  [below] at (8,-0.3) {$T$};

	\node at (5,-.5) {$\cdots$};
	\node  at (0.5,0.4) { $\overbrace{ }$};
	\node  at (0.5,0.8) {$\q /2$};
		\node  at (2,0.5) { {\Huge $\overbrace{ }$}};
	\node  at (2,0.9) {$\q$};

	\node  at (7.5,0.4) { $\overbrace{ }$};
	\node  at (7.5,0.8) {$\q /2$};
	
\end{tikzpicture}
\end{figure}

and we calculate exactly the data as:
$$
\begin{array}{rcrcccl}
\x		&	 = 		&	[ 	& \sin t_1  & \cdots & \sin t_N & ]		 \\
\dot{\x}	&	 = 		&	[	& \cos t_1 & \cdots & \cos t_N & ]		 \\
\y		&	 = 		&	[  	& 2x_1-1  & \cdots & 2 x_N -1 & ]		 \\
\end{array}
$$

We start our study in the case $\mu \ug 0$ to simplify some calculation. 
In this case ( see  \cite{report} ) we can choose directly the solutions of (\ref{dissFO}) or (\ref{dissSO}) composing the Impulsive Response $g$, instead of select the coefficients $\a_j$ of the differential operator $P$. We assume $\a_k \ug 1$, where $k$ is the order of $P$.

%
%

\subsubsection{First Order Operator}

\begin{figure}[H]
\hspace{-1cm}
\includegraphics[scale=0.5]{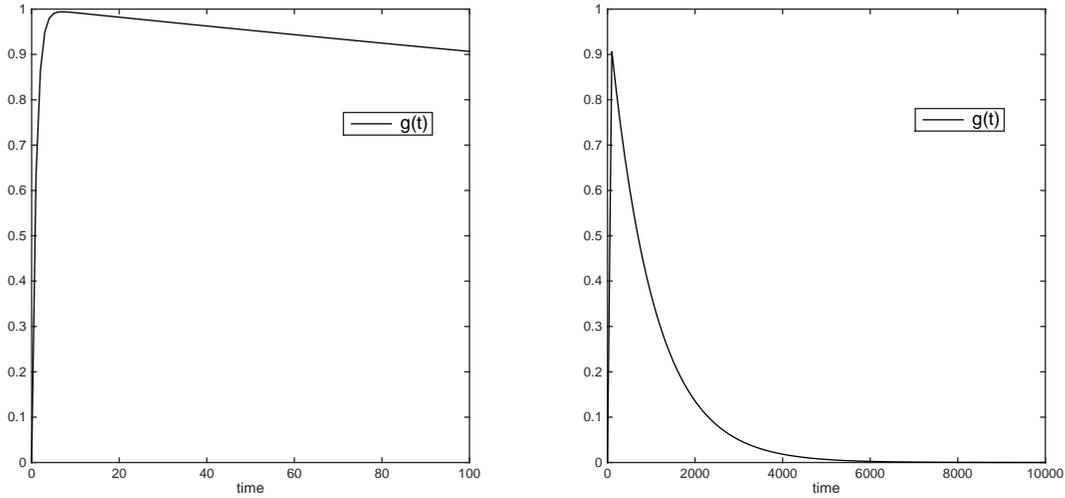}
\caption{Impulsive Response $g$ coming from a differential operator with $\a_0 \ug 0.999\, , \; \a_1 \ug 1$ and $\t \ug 1$.}\label{ImRe1}
\end{figure}

We first try with the solutions $\ell_1 \ug -10^{-3} \, , \; \ell_2 = -0.999$ which correspond of a setting with parameters $\t \ug 1$ and $\a_0 \in \left\{ 0.001\, ,\: 0.999 \right\}$. This lead to the Impulsive Response of Fig.\ref{ImRe1}. We choose null initial conditions $\f[0] \ug \left[ \, 0 \, , \, 0 \, \right]' $.

\begin{figure}[H]
\hspace{0cm}
\includegraphics[scale=0.8]{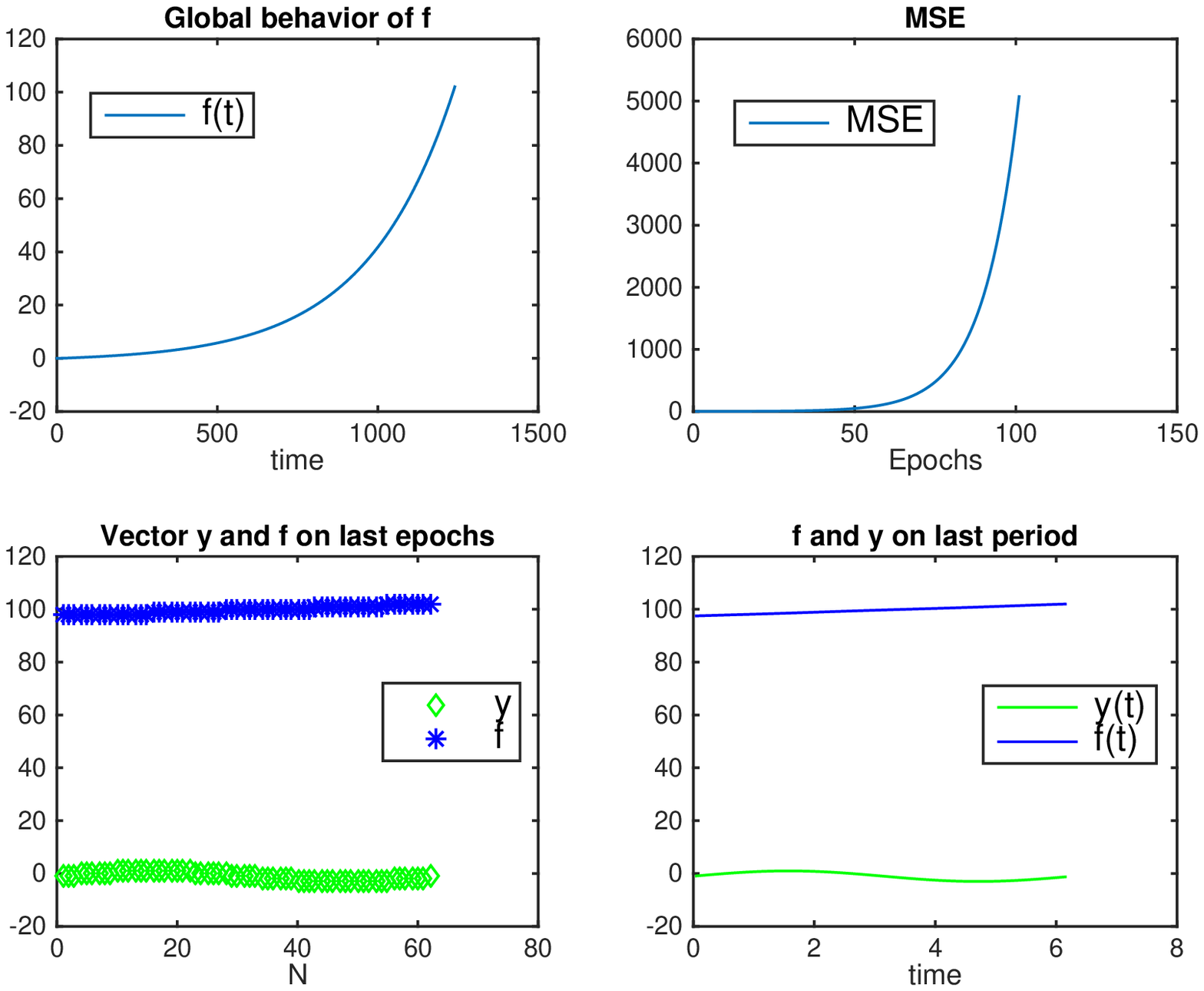}
\caption{Parameters $ \t \ug 1 \, , \; \a_0 \ug 0.999\, , \; \a_1 \ug 1 \, , \; \q \ug 0.1  \, , \; \l  \ug 10^3$ , Epochs:  100.}\label{fo1}
\end{figure}

\begin{figure}[H]
\hspace{0cm}
\includegraphics[scale=0.8]{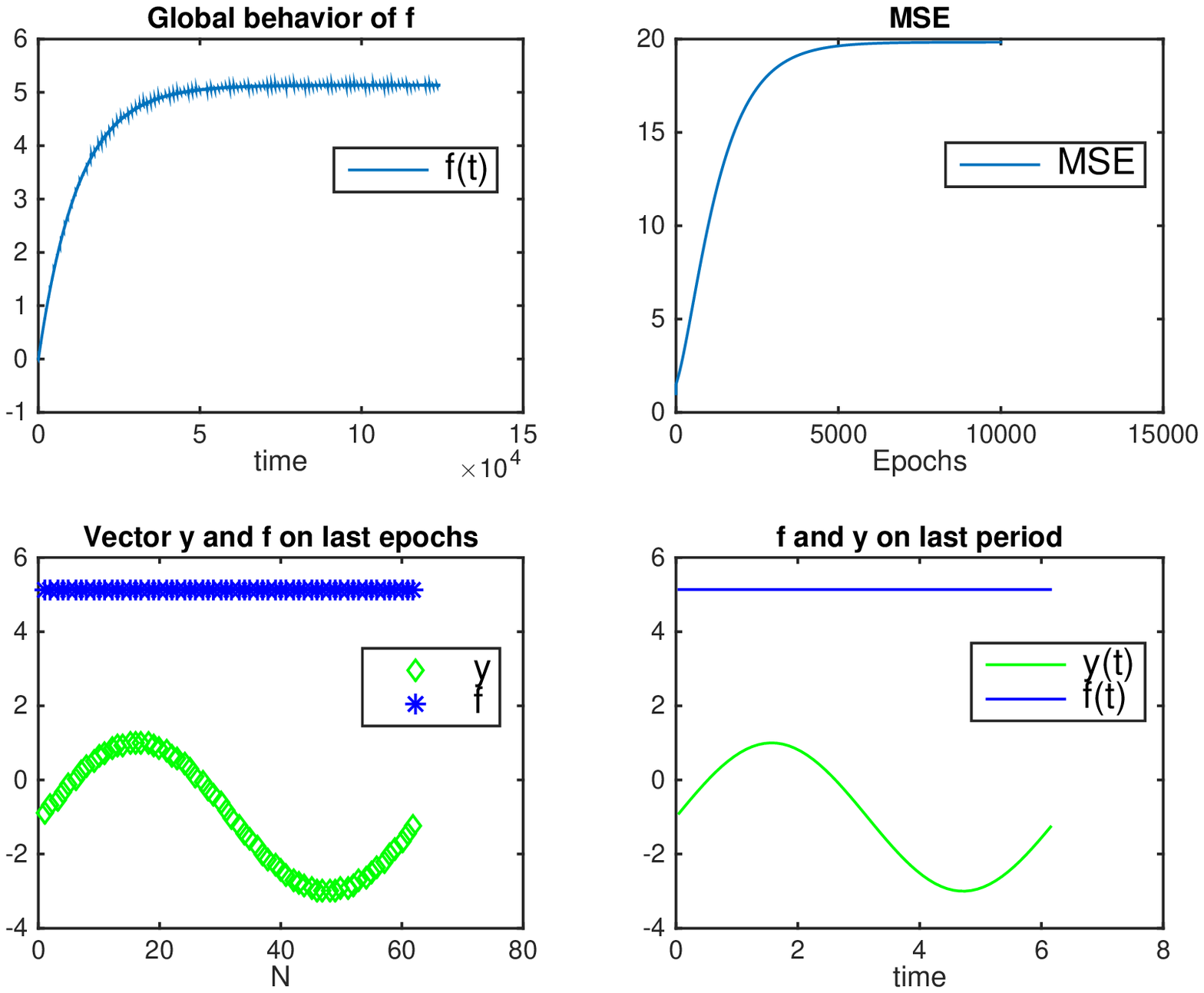}
\caption{Parameters $  \t \ug 1 \, , \; \a_0 \ug 0.999\, , \; \a_1 \ug 1 \, , \; \q \ug 0.1 \, , \; \l  \ug 10^4$ , Epochs: $10^4$ .}\label{fo2}
\end{figure}

We set $\q \ug 0.1$ and we find $N \ug 62$ samples between $0.05$ and $2\pi$. We try different setting but not suitable solutions appear. We find a divergent $\by$ when $\l \ug 10^3 $ as we can see in Fig.\ref{fo1}. 
When we unbalancing the system toward regularization, by enlarging $\l$, we obtain the result of Fig.\ref{fo2}. We try with different values of $\l$ but the behavior is the same. In the plots of each figure we can see:
\begin{itemize}
\item  at the top left, the global behavior of $\by$ versus time of the system;
\item  at the top right, the evaluations of the MSE at each epoch evaluated over all the $N$ points of supervision;
\item  at the bottom left, the comparing of the vector $\y$ and the correspondent evaluations of ${\tilde{f}}_1$ on the instants of supervision in the last epoch of training;
\item  at the bottom right, the plot of $\y$ and the correspondent evaluations of ${\tilde{f}}_1$ on the instants of supervision in the last epoch of training versus the time.
\end{itemize}

\begin{figure}[H]
\hspace{0cm}
\includegraphics[scale=0.8]{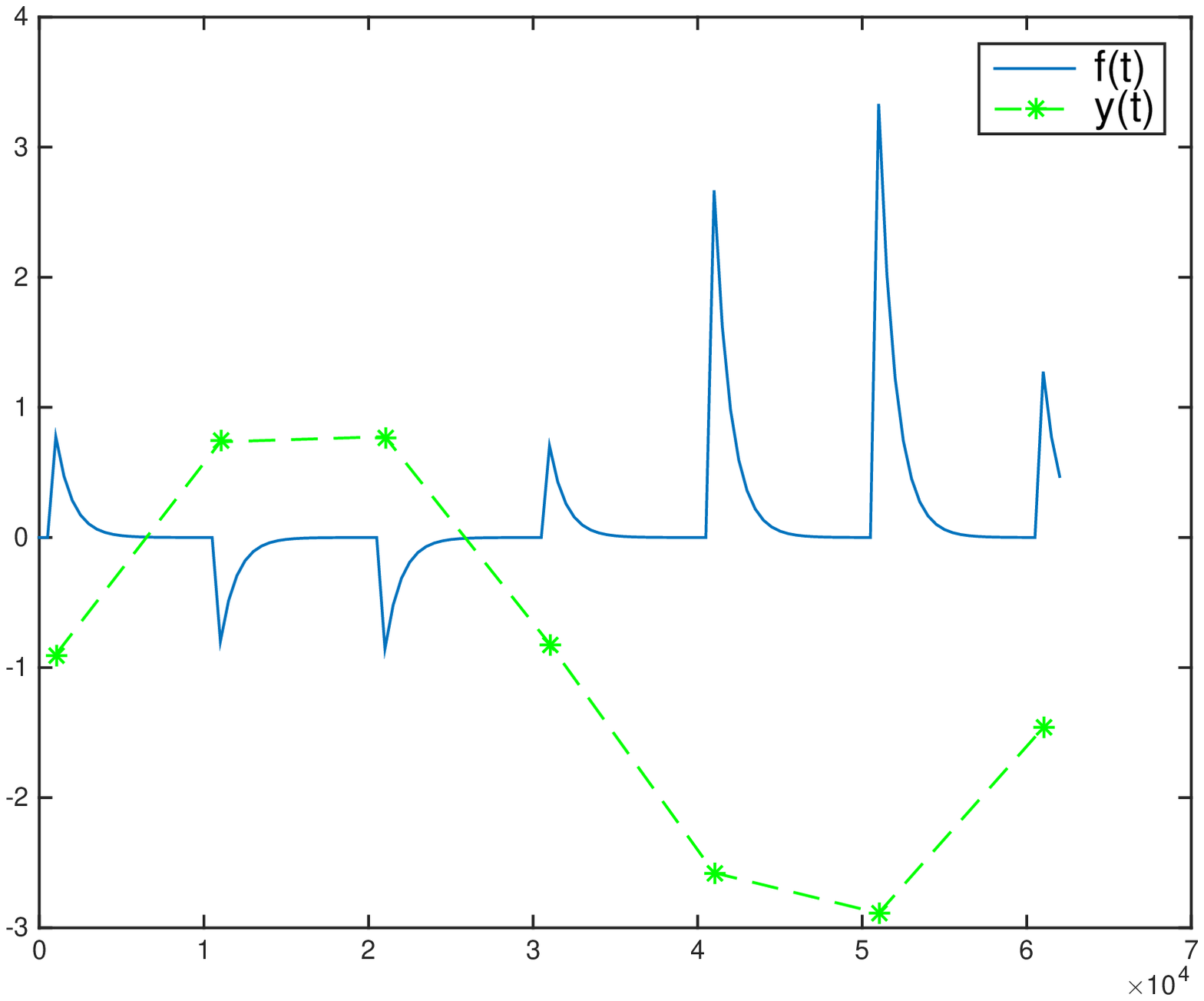}
\caption{Parameters $  \t \ug 1 \, , \; \a_0 \ug 0.999\, , \; \a_1 \ug 1 \, , \; \q \ug 10^3 \, , \; \l  \ug 0.5$ , Epochs: $1$ .}\label{fo2a}
\end{figure}

We obtain similar results both for different values of $\q$ and the differential operator $P$. Indeed, when we use an operator of odd order $k$, the last coefficients of the adjoint operator $P^*$ is negative. This lead to a sign flip in the regularization term of the functional $\Phi$ to be optimize. In this case, our solutions becomes a maximum for the opposite of loss function. This maximum is infinite, unless we impose a strong regularization, by enlarging $\q$ or $\l$. The intuitive idea is that, the regularization impose the function to reach $0$ between to consecutive impulses. We show this case in Fig.\ref{fo2a}, where we decrease the number of impulses and impose a strong regularization.

\begin{figure}[H]
\hspace{0cm}
\includegraphics[scale=0.8]{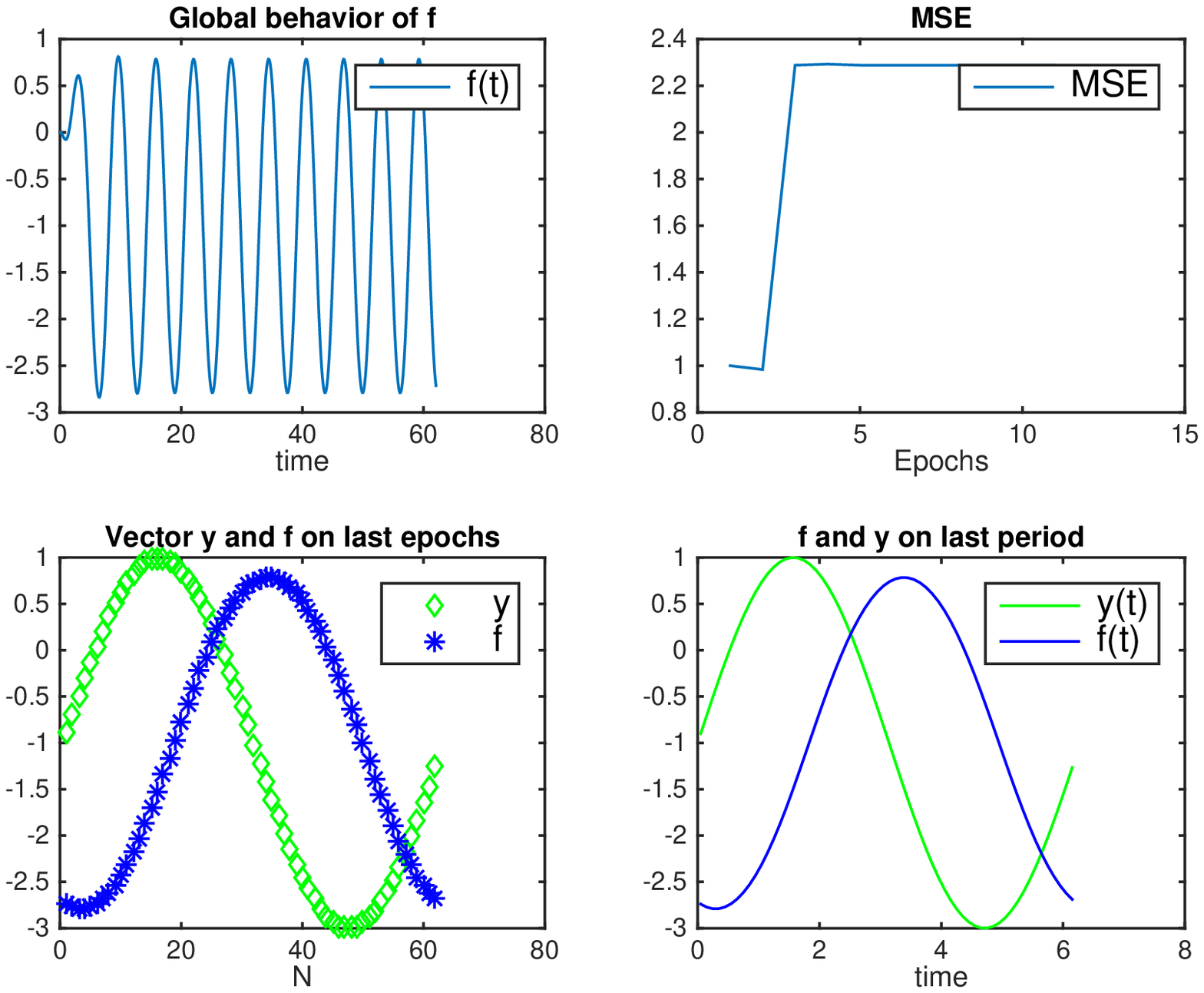}
\caption{Parameters $ \t \ug 1 \, , \; \a_0 \ug 0.999\, , \; \a_1 \ug 1 \, , \;  \q \ug 0.1 \, , \;  \l  \ug -10$ , Epochs:  10 .}\label{fo3}
\end{figure}

Regardless the theory which require $\l >0$, but trying to follow an analogy with \cite{report}, we try some experiments with $\l < 0$. For the same differential operator, we find the behavior of Fig.\ref{fo3}.

\begin{figure}[H]
\hspace{0cm}
\includegraphics[scale=0.8]{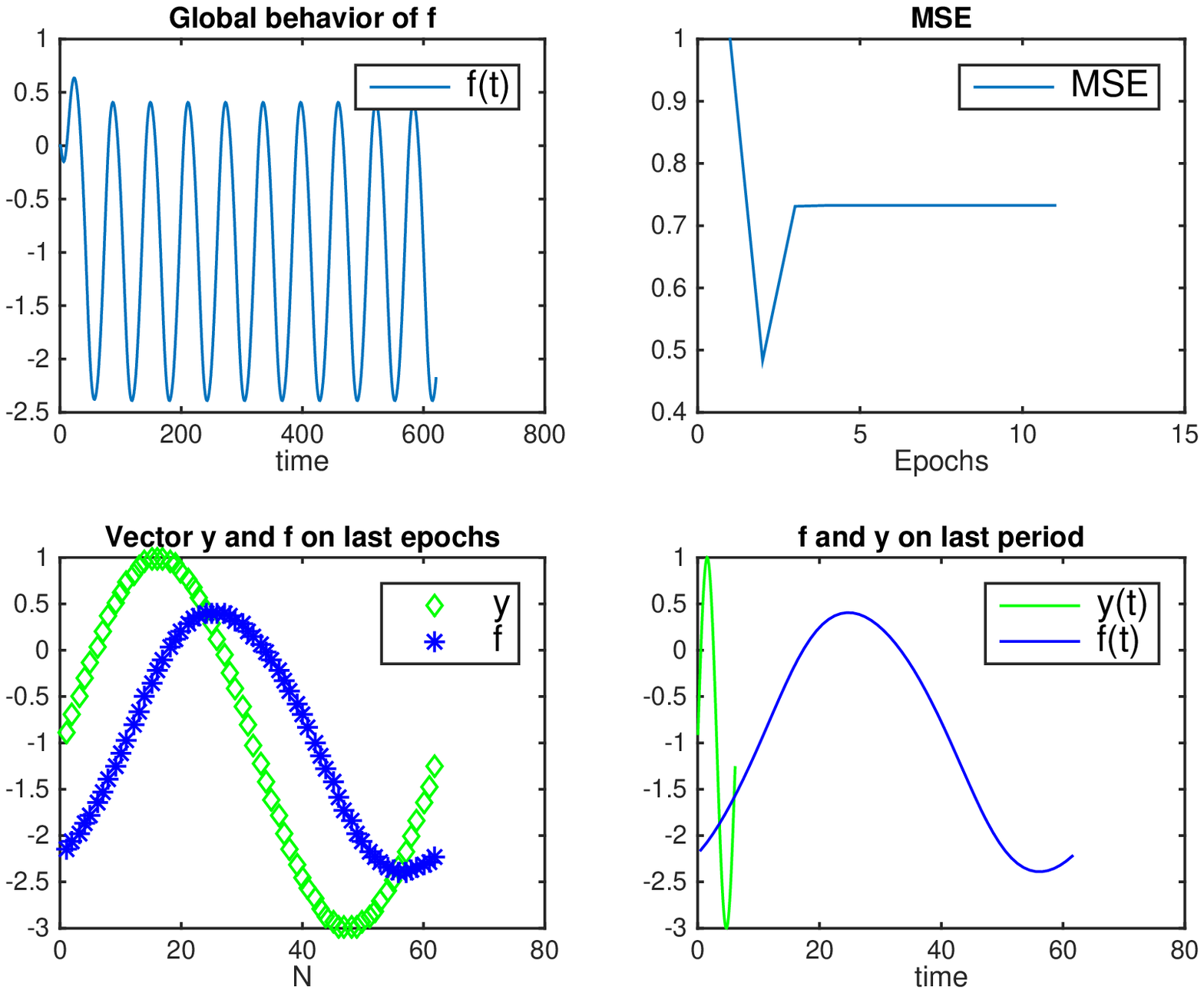}
\caption{Parameters $ \t \ug 1 \, , \; \a_0 \ug 0.999\, , \; \a_1 \ug 1 \, , \; \q \ug 0.1\, , \; \q' \ug 1 \, , \;  \l  \ug -10$ , Epochs: 10  .}\label{fo4}
\end{figure}

We still do not find a good approximation, but we can see that in this case $\by$ respect in some way the amplitude of the oscillation of $y$, but with a longer period. The idea is then to accelerate $\by$ so as to obtain in the points of updating the same value of of the supervisions $y_i$. We update $\by$ with a sampling step $\q' \ug 10 \q \ug 1$, that is:
$$
\f [K+1]  =   e^{\A \q'} \f [ K]  + e^{\A \q' /2} \cdot \B \frac{1}{\l \a_1^2 b(t_{\bar{i}}) }\left[ \by (t_{\bar{i}}) - y_{\bar{i}} \right]               
$$
In this way we drop the real dependence on time between to neighbor samples of $\y$, but we consider $\y$ as a vector of sorted points, together with their correspondent derivatives $ \dot{\x}$ (necessary for the updating). With these assumption we obtain the results of Fig.\ref{fo4}.

\begin{figure}[H]
\hspace{0cm}
\includegraphics[scale=0.8]{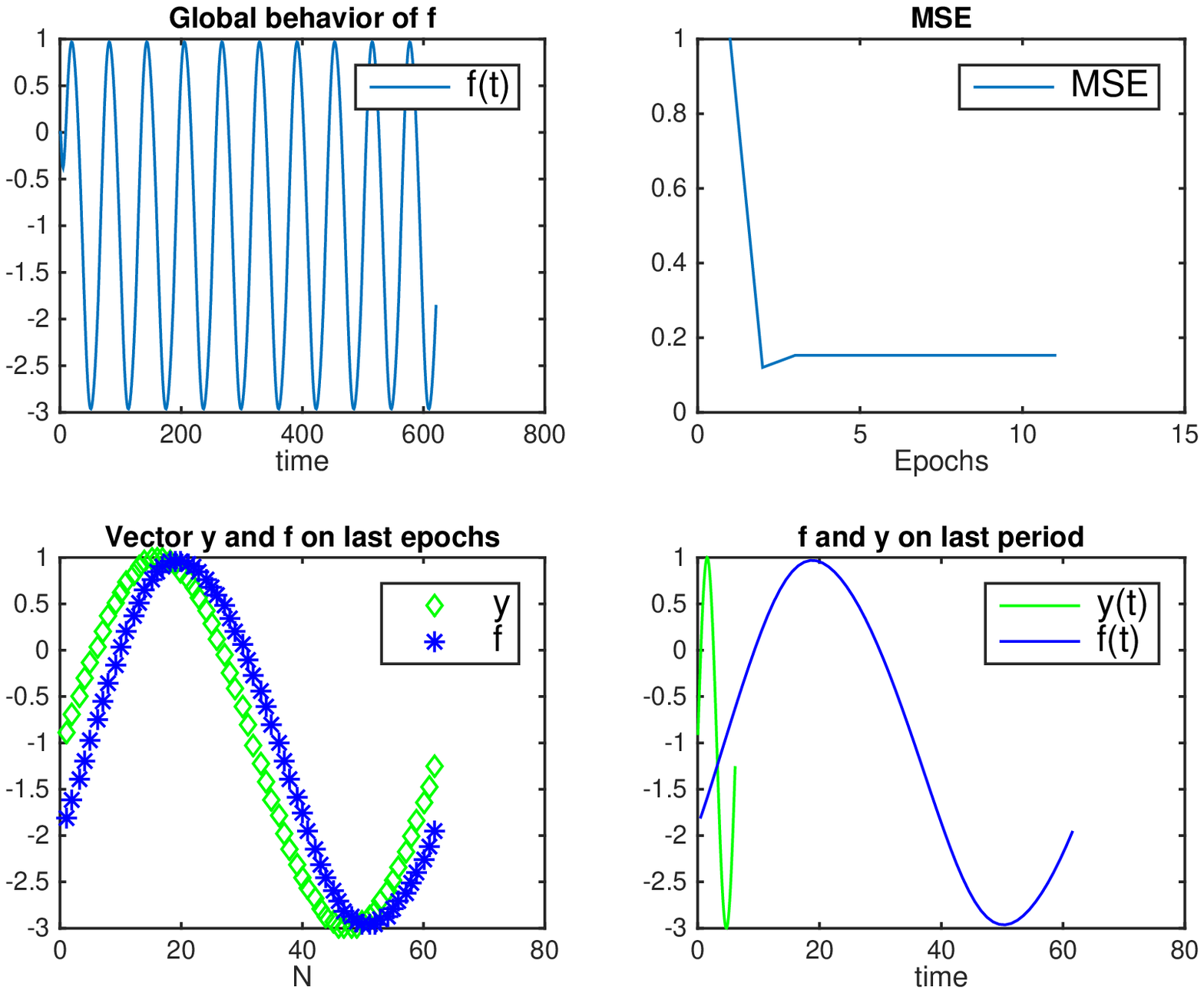}
\caption{Parameters $ \t \ug 1 \, , \; \a_0 \ug 0.999\, , \; \a_1 \ug 1 \, , \;  \q \ug 0.1\, , \; \q' \ug 1 \, , \;  \l  \ug -3$ , Epochs: 10  .}\label{fo5}
\end{figure}

The function $\by$ is still different from $y$, but this time an unbalancing towards the fitting by taking $\l \ug -3$ affect the results. We can see a good interpolation in Fig.\ref{fo5} which became better if we enlarge $N$ (by taking a smaller $\q$).

\begin{figure}[H]
\hspace{-1cm}
\includegraphics[scale=0.5]{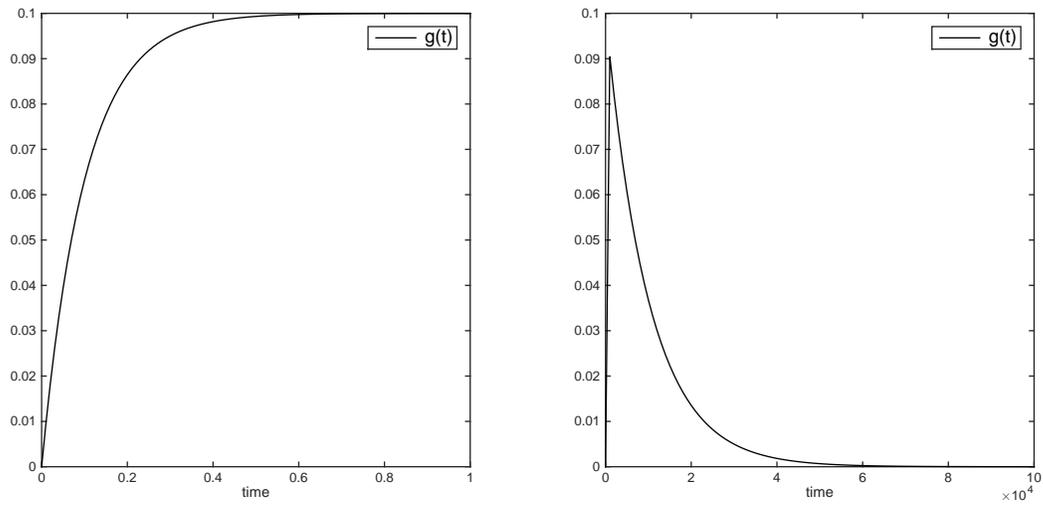}
\caption{Impulsive Response $g$ coming from a differential operator with $\a_0 \ug 9.999\, , \; \a_1 \ug 1$ and $\t \ug 10$.}\label{ImRe1a}
\end{figure}

This acceleration of $\by$ can in some way be linked to an acceleration of the Impulsive Response. The delay of the system w.r.t. data can be related to the delay of the Impulsive Response, which does not promptly reach its maximum in the analyzed cases. As in \cite{report}, we can implement an High Dissipation setting to have a quicker $g$. Again, a large value of $\t$ could make the weight $e^{\t t}$ in the function $\psi$ unmeaningful. But since the memory of the system depends on the smaller solution, we can balance this by taking a solutions close to $0$. For examples, if we take $\t \ug 10$ and $\a_0 \ug 9.999$ we have the solutions $\ell_1 \ug -10^{-4} \, , \; \ell_2 = -9.999$ and Impulsive Response of Fig.\ref{ImRe1a}.

\begin{figure}[H]
\hspace{0cm}
\includegraphics[scale=0.8]{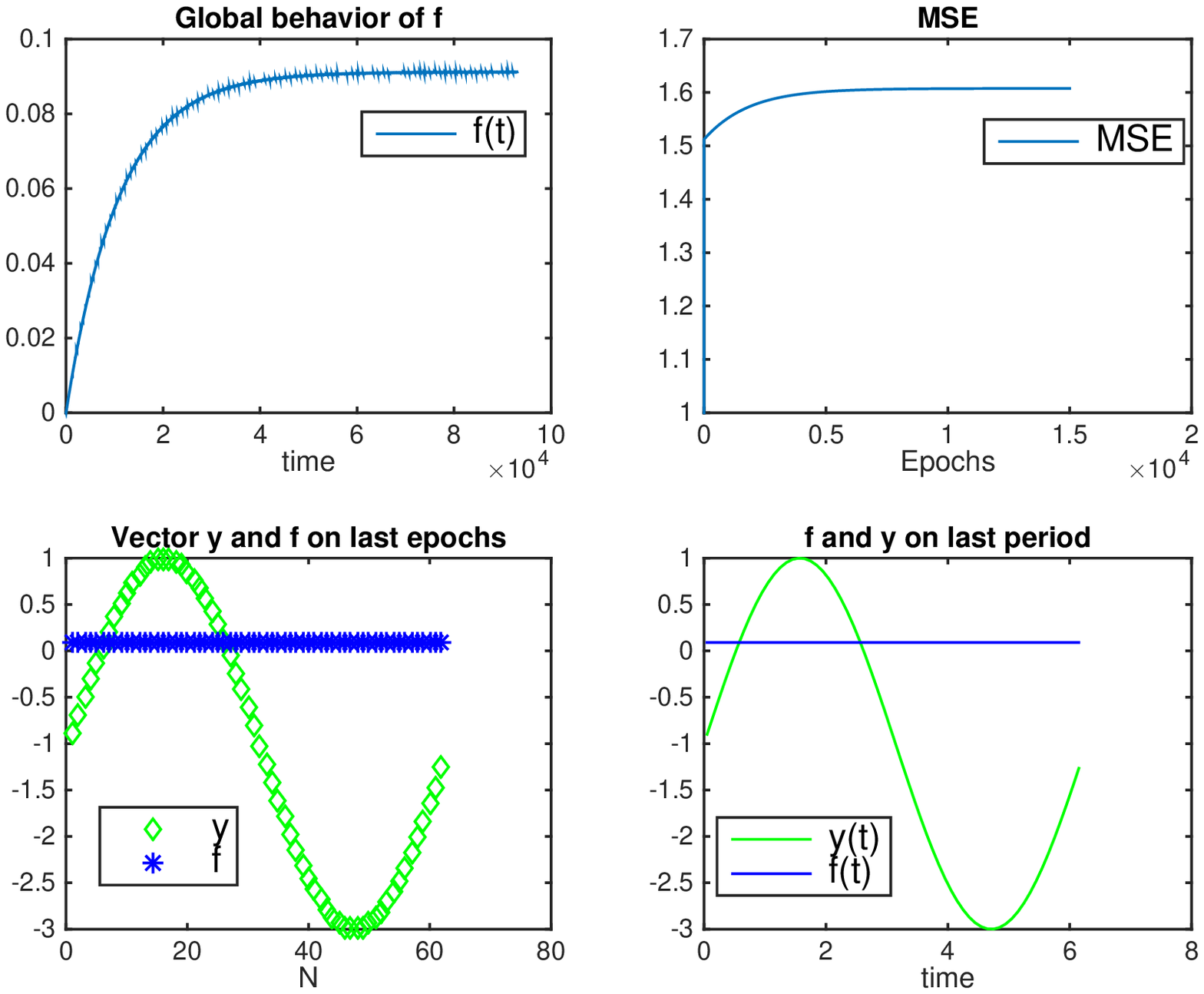}
\caption{Parameters $ \t \ug 10 \, , \; \a_0 \ug 9.999\, , \; \a_1 \ug 1 \, , \;  \q \ug 0.1 \, , \;  \l  \ug 10^5$ , Epochs: $15000$  .}\label{fo6}
\end{figure}

\begin{figure}[H]
\hspace{0cm}
\includegraphics[scale=0.8]{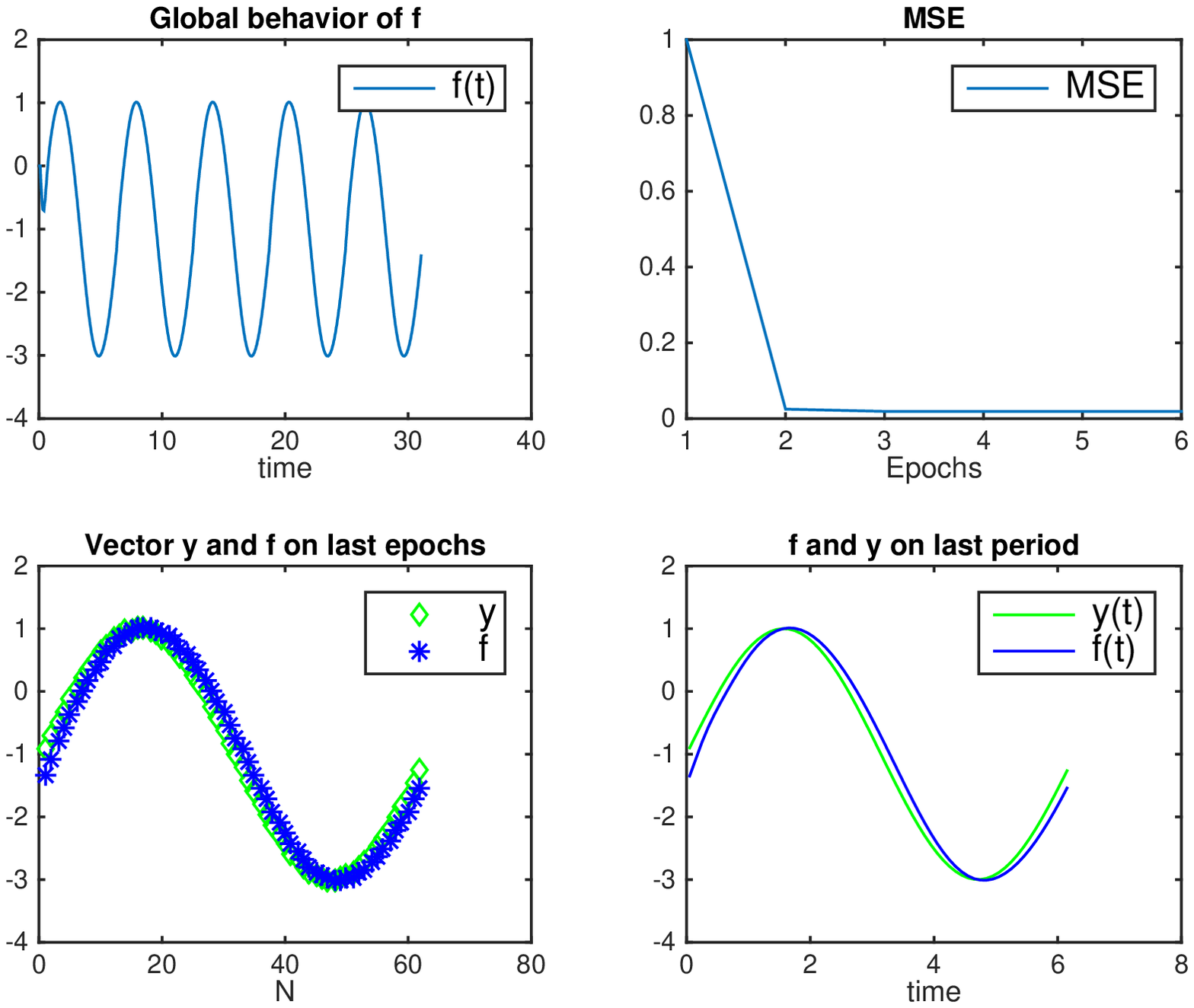}
\caption{Parameters $ \t \ug 10 \, , \; \a_0 \ug 9.999\, , \; \a_1 \ug 1 \, , \;  \q \ug 0.1\, , \;  \l  \ug -0.1$ , Epochs: 5  .}\label{fo7}
\end{figure}

For positive and negative value of $\l$ we have the results of Fig.\ref{fo6},\ref{fo7}.

%
%

\subsubsection{Second Order Operator}

\begin{figure}[H]
\hspace{-1cm}
\includegraphics[scale=0.5]{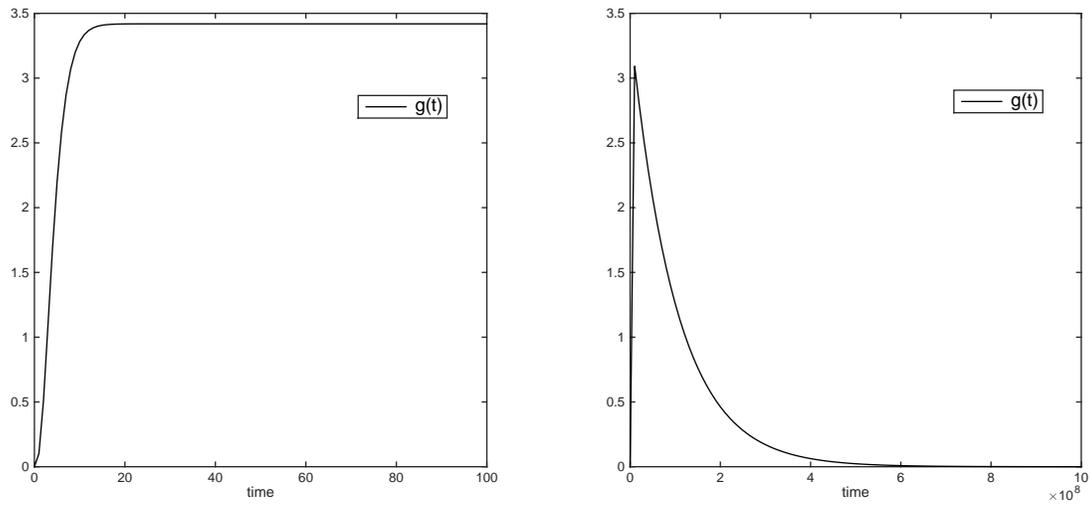}
\caption{Impulsive Response $g$ versus two different time interval, $\a_0 \ug 1\, , \; \a_1 \simeq 1.99 \, , \; \a_2 \ug 1$ and $\t \ug 1$.}\label{ImRe2}
\end{figure}

We start again with the dissipation term $\t \ug 1$ and null initial conditions $\f[0] \ug \left[ \, 0 \, , \, 0 \, , \, 0\, , \, 0 \, \right]' $. We choose the solutions of (\ref{dissSO}) as 
$$
\begin{array}{rll}
\ell_1 	&	=	 -10^{-8} \\ 
\ell_2 	&	=	-0.6 \\ 
\ell_3 	&	=	-0.65 \\ 
\ell_4 	&	=	-0.74999999 \\ 
\end{array}
$$
which generate the impulsive Response $g$ of Fig.\ref{ImRe2}.

\begin{figure}[H]
\hspace{0cm}
\includegraphics[scale=0.8]{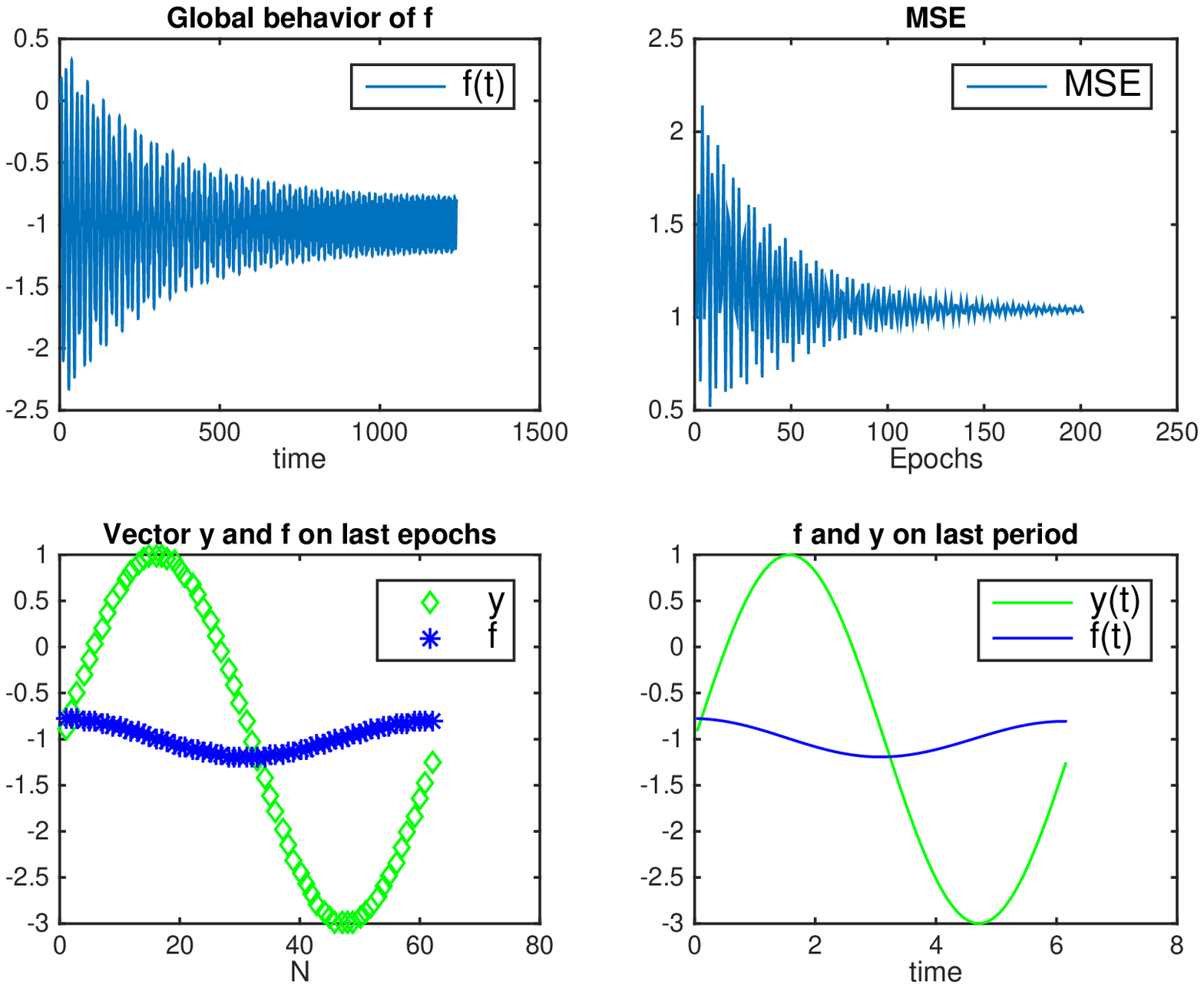}
\caption{Parameters $ \t \ug 1 \, , \; \a_0 \ug 1\, , \; \a_1\simeq 1.99 \, , \; \a_2 \ug 1 \, , \;  \q \ug 0.1 \, , \;  \l  \ug 50$ , Epochs: 200.}\label{so1}
\end{figure}

\begin{figure}[H]
\hspace{0cm}
\includegraphics[scale=0.8]{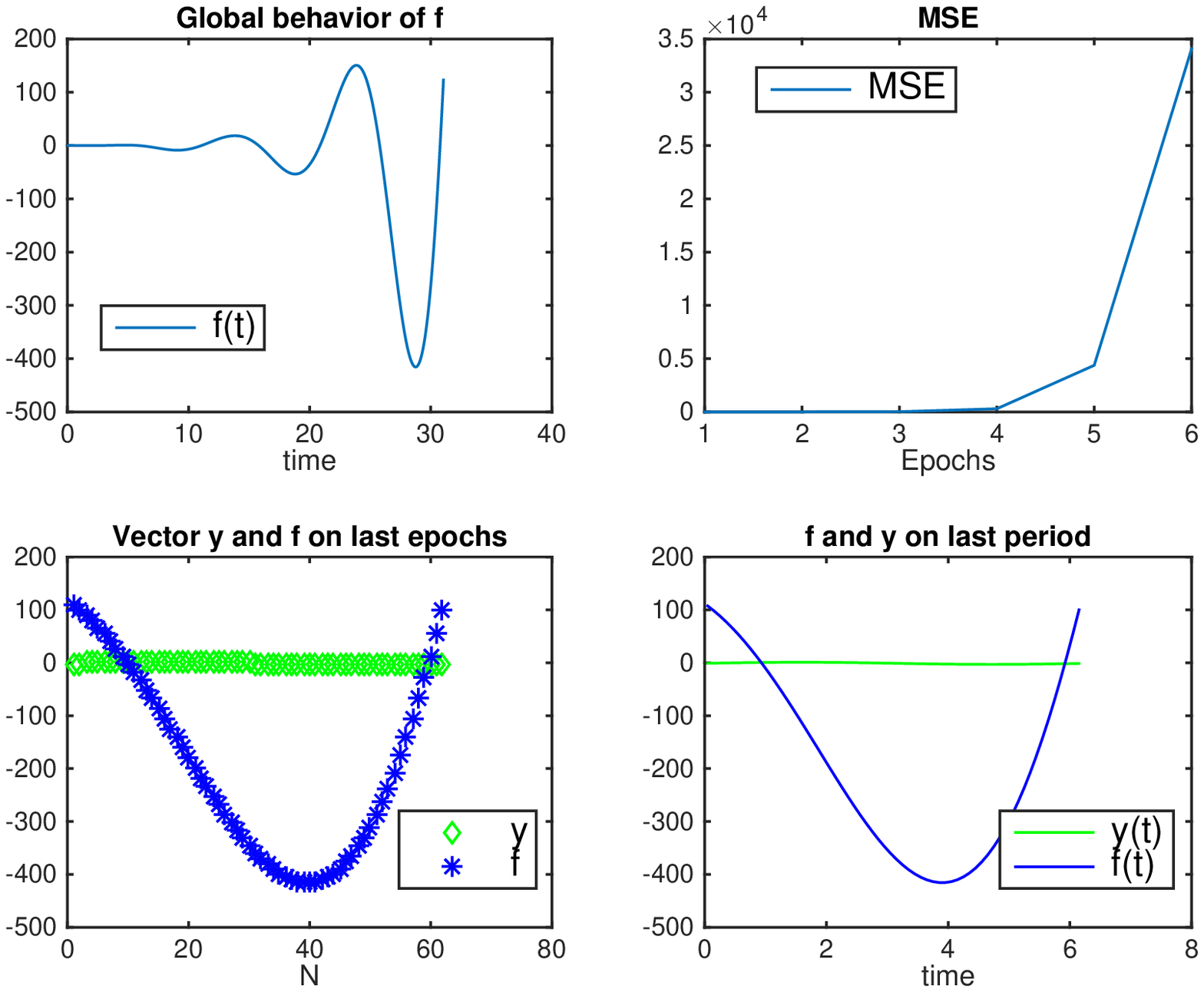}
\caption{Parameters $ \t \ug 1 \, , \; \a_0 \ug 1\, , \; \a_1\simeq 1.99 \, , \; \a_2 \ug 1 \, , \; \q \ug 0.1\, , \;  \l  \ug 10$ , Epochs: 5.}\label{so2}
\end{figure}

Also in this case, we can not find suitable settings. We find the behavior of Fig.\ref{so1} with $\l \ug 50$, whereas the we obtain Fig.\ref{so2} when we reduce regularization with $\l \ug 10 $.

\begin{figure}[H]
\hspace{0cm}
\includegraphics[scale=0.8]{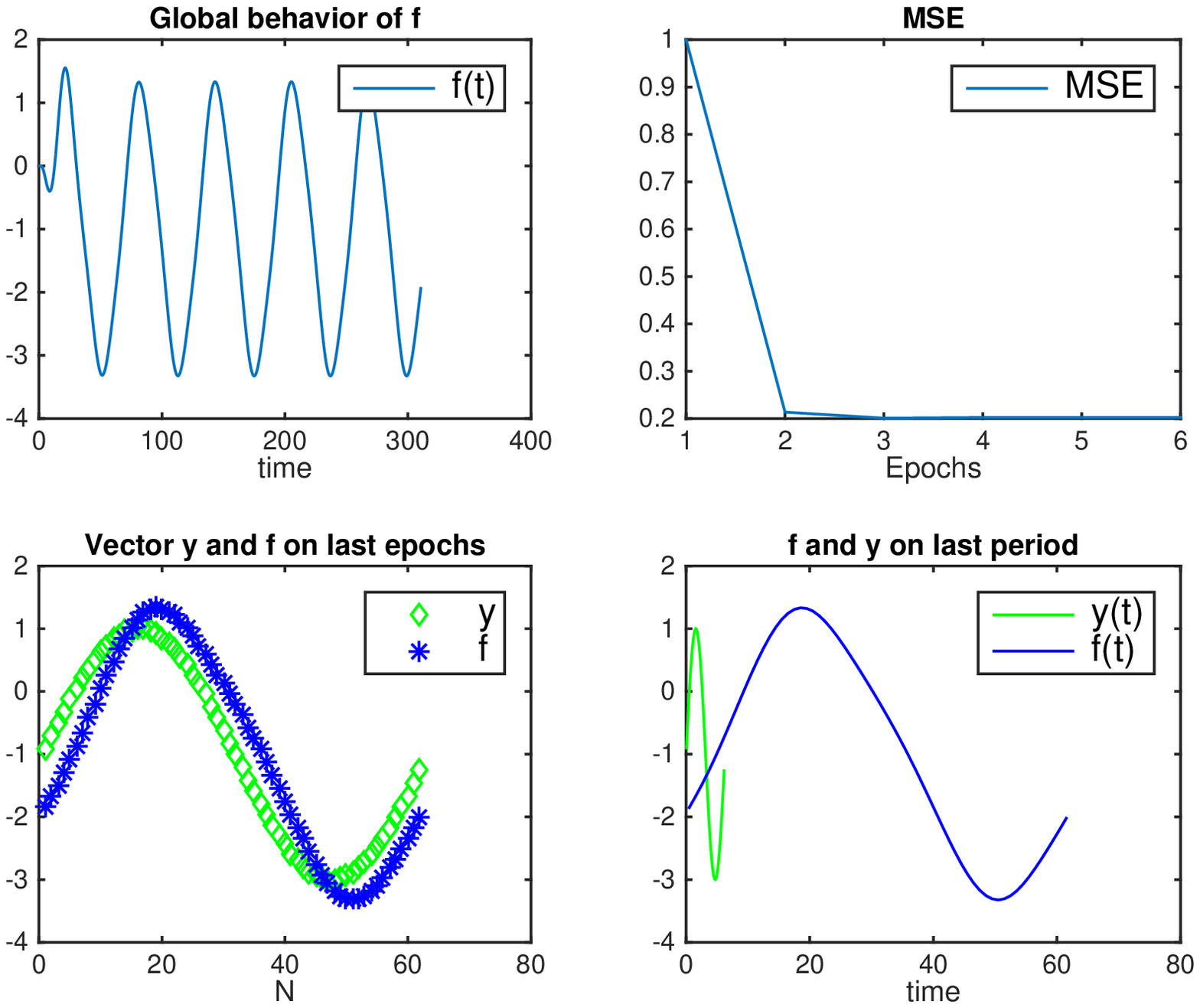}
\caption{Parameters $ \t \ug 1 \, , \; \a_0 \ug 1\, , \; \a_1\simeq 1.99 \, , \; \a_2 \ug 1 \, , \; \q \ug 0.1\, , \; \q' \ug 1 \, , \;  \l  \ug 10$ , Epochs:  5.}\label{so3}
\end{figure}

Again, if we accelerate $\by$ w.r.t. $\y$ we can find a good fitting, as shown in Fig.\ref{so3}.

\begin{figure}[H]
\hspace{0cm}
\includegraphics[scale=0.8]{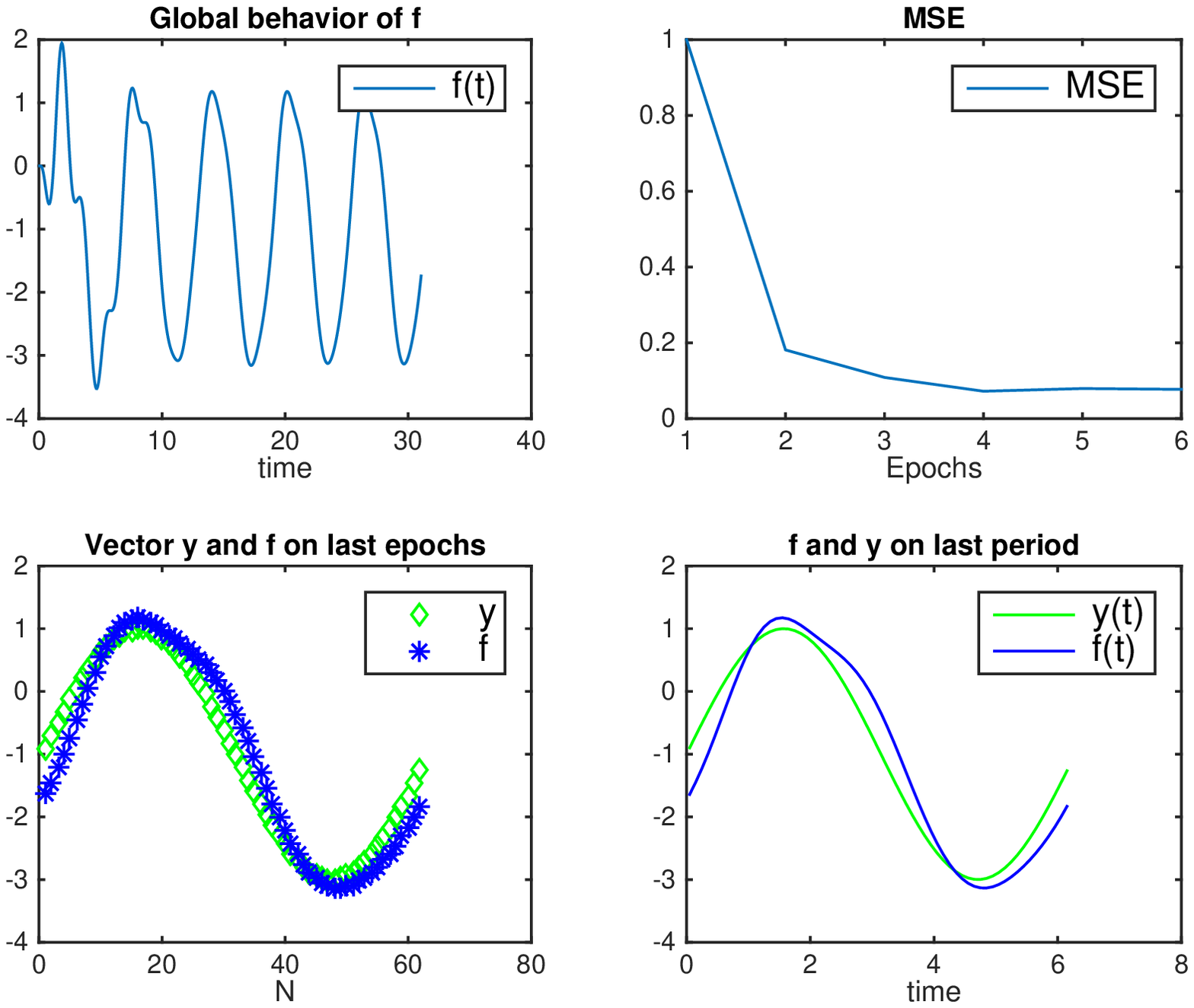}
\caption{Parameters $ \t \ug 10 \, , \;  \q \ug 0.1\, , \; \q' \ug 1 \, , \;  \l  \ug 0.006$ , Second order $P$ , Epochs:  5.}\label{so4}
\end{figure}

Again, when we reduce the delay of the impulse response, for example we take $\t \ug 10$, we have the results of Fig.\ref{so4}

%
%

\subsubsection{Dependence on Initial Conditions}
\begin{figure}[H]
\hspace{0cm}
\includegraphics[scale=0.8]{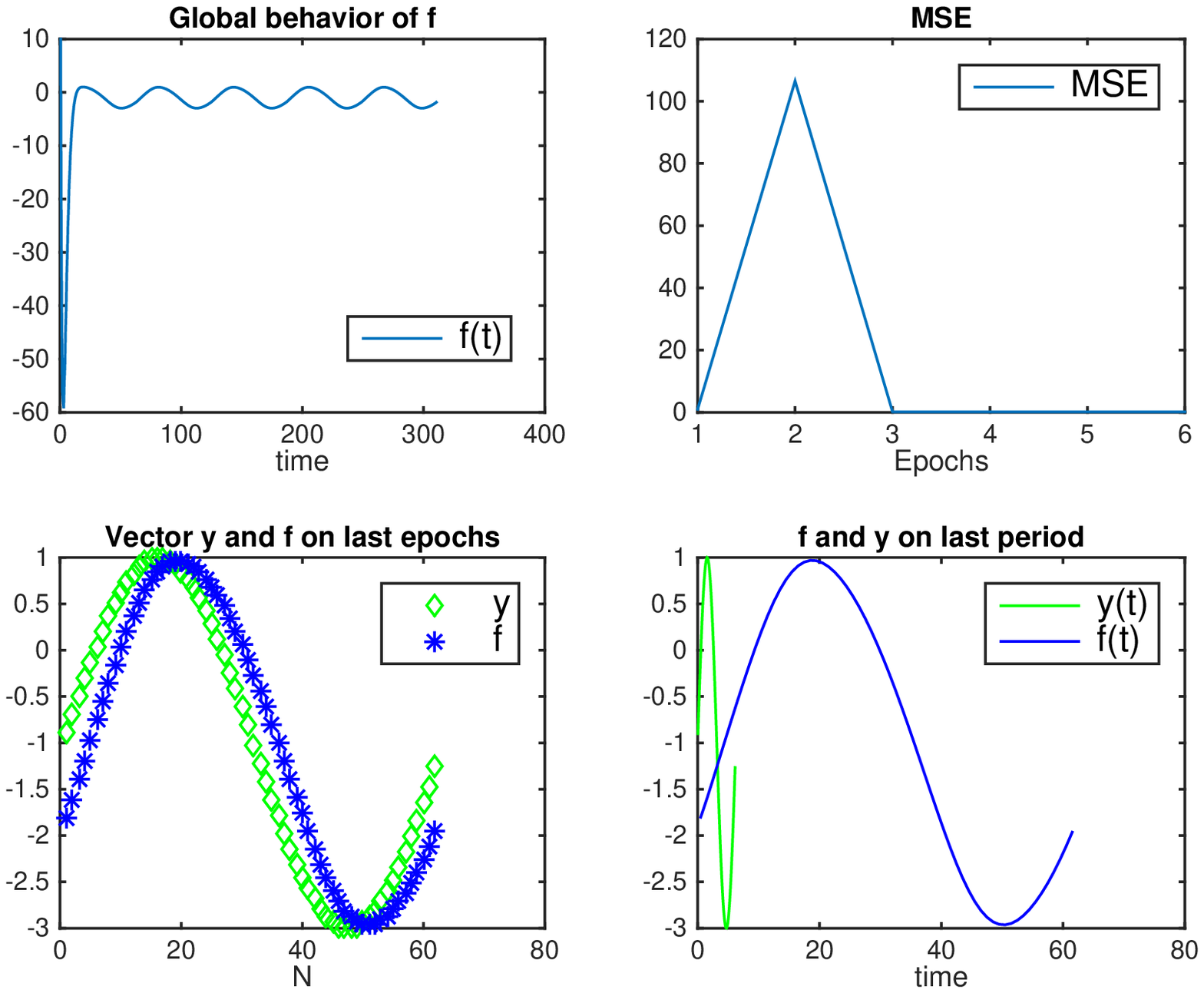}
\caption{Behavior of the system with the same first order operator setting of Fig.\ref{fo5} : $ \t \ug 1 $ , $ \a_0 \ug 1\, , \; \a_1 \ug 1  \, , \; \q \ug 0.1\, , \; \q' \ug 1 \, , \;  \l  \ug -3 $ with Initial Conditions $\f[0] \ug \left[ \, 10 \, , \, -90 \, \right]' $. }\label{IC1}
\end{figure}

\begin{figure}[H]
\hspace{0cm}
\includegraphics[scale=0.8]{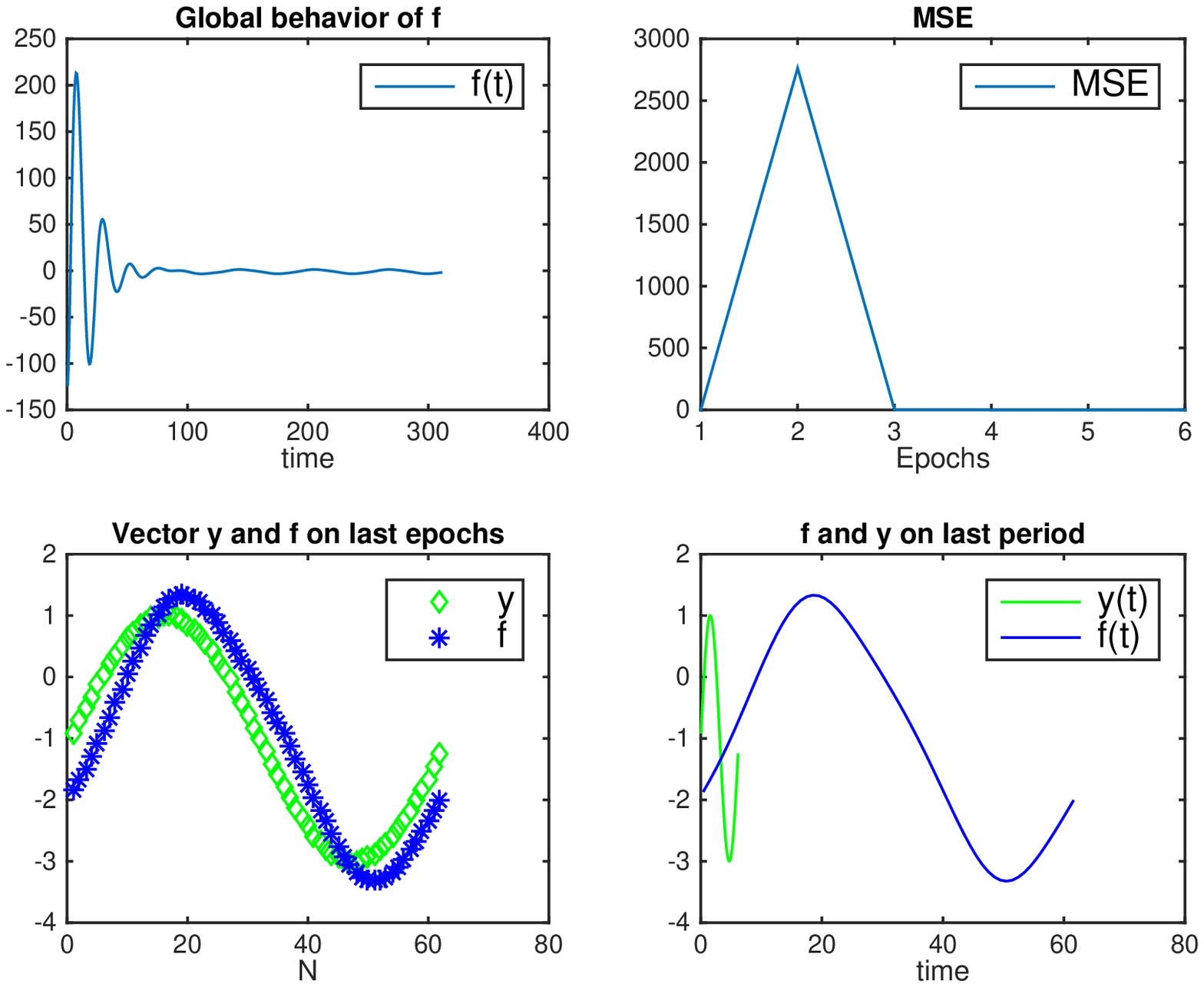}
\caption{Behavior of the system with the same second order operator setting of Fig.\ref{so3} : $ \t \ug 1 \, , \; \a_0 \ug 1\, , \; \a_1\simeq 1.99 \, , \; \a_2 \ug 1 \, , \; \q \ug 0.1\, , \; \q' \ug 1 \, , \;  \l  \ug 10$ with Initial Conditions $\f[0] \ug \left[ \, -123 \, , \, 22 \, , \, 45 \, , \, -10\right]' $. }\label{IC2}
\end{figure}

After this first results we explored briefly the dependence of the system on the Initial Conditions $\f(0) \ug \f [0]$. We only report the experiments in Fig.\ref{IC1},\ref{IC2} for the first and second order operator respectively. Because of this results, we will assume null Initial Conditions from now on.

%
%

\subsubsection{Different Input Function}

\begin{figure}[H]
\hspace{3cm}
\includegraphics[scale=0.5]{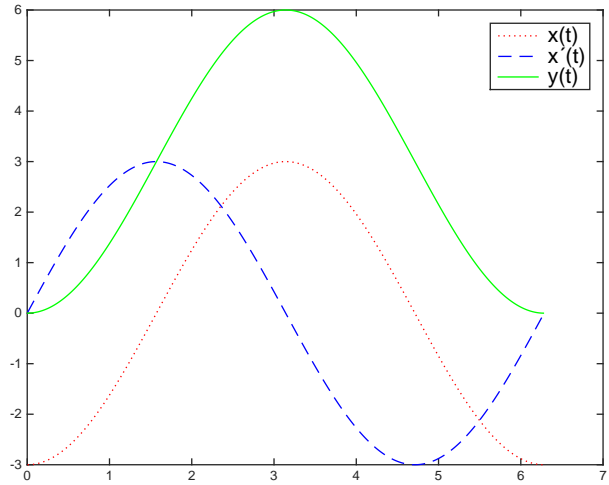}
\caption{Graphs of the input function $x(t)$ with its derivative $x'(t)$ and the target $y(t)$}\label{input2}
\end{figure}

We use a time sampling step $\q \ug 0.01$ (which implies $N\ug 628$) and calculateour data as:
$$
\begin{array}{rcrcccl}
\x		&	 = 		&	[ 	& -3\cos t_1  & \cdots & -3\cos t_N & ]		 \\
\dot{\x}	&	 = 		&	[	& 3\sin t_1 & \cdots & 3\sin t_N & ]		 \\
\y		&	 = 		&	[  	& x_1+3  & \cdots & x_N +3 & ]		 \\
\end{array}
$$
which are shown in Fig.\ref{input2}.

\begin{figure}[H]
\hspace{0cm}
\includegraphics[scale=0.8]{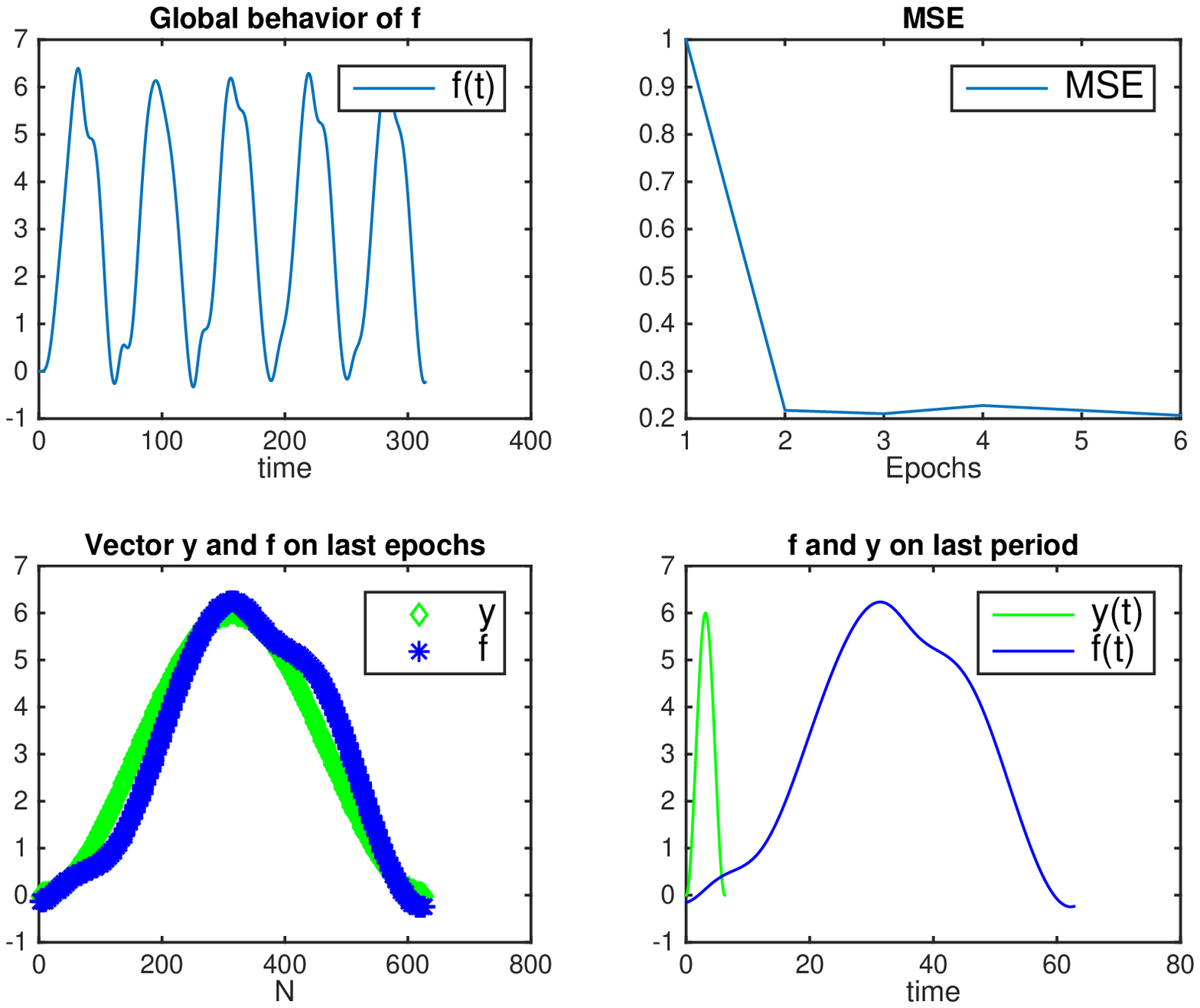}
\caption{Behavior of the system in the same setting of Fig.\ref{so3}: $ \t \ug 1 \, , \; \a_0 \ug 1\, , \; \a_1\simeq 1.99 $ ,  $ \a_2 \ug 1 \, , \; \q \ug 0.01 \, , \; \q' \ug 0.1 \, , \; \l \ug 30$ , Epochs:~5.}\label{newinput}
\end{figure}

\begin{figure}[H]
\hspace{0cm}
\includegraphics[scale=0.8]{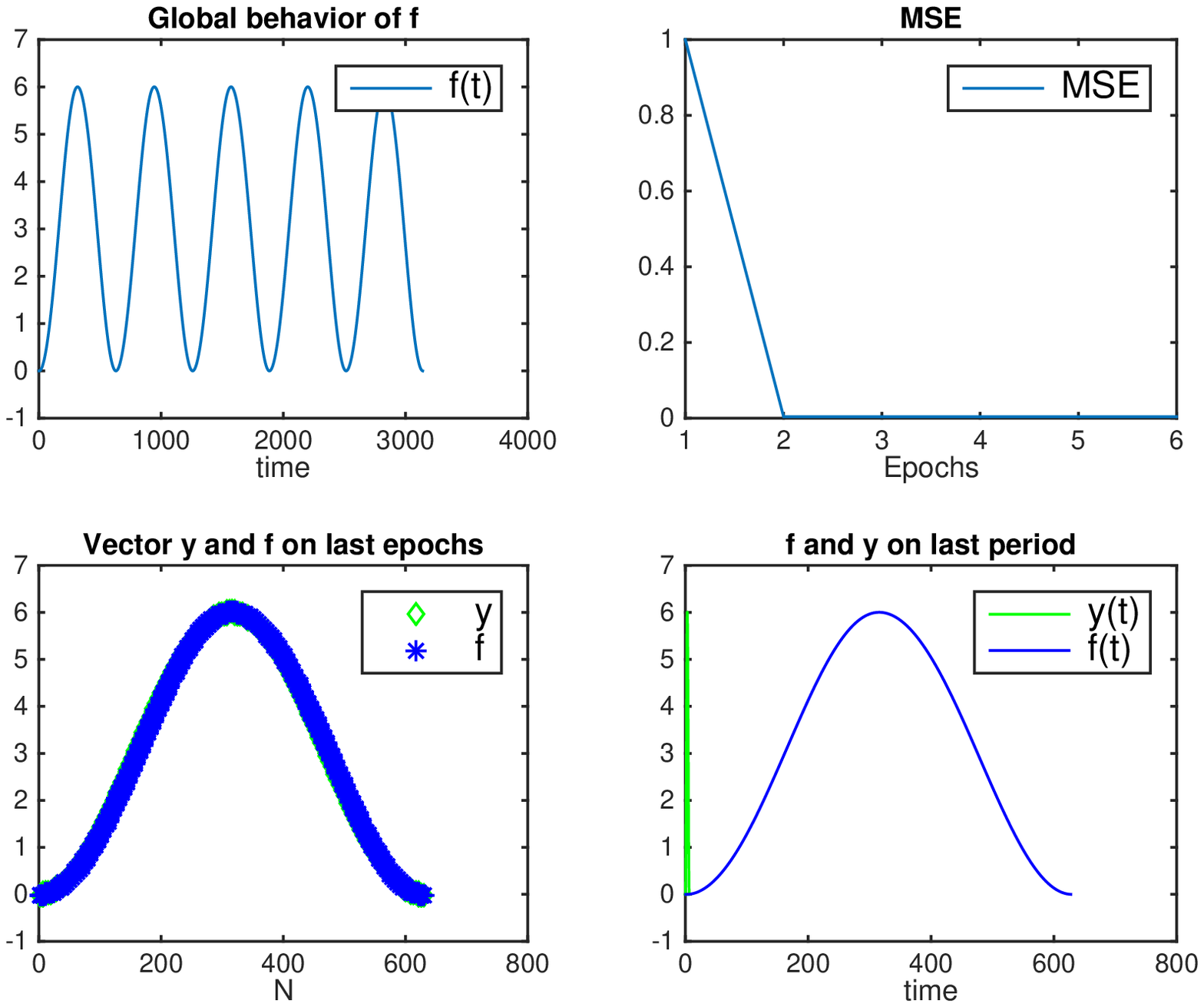}
\caption{Behavior of the system in the same setting of Fig.\ref{so3}: $ \t \ug 1 \, , \; \a_0 \ug 1\, , \; \a_1\simeq 1.99 $ , $ \a_2 \ug 1 \, , \; \q \ug 0.01 \, , \; \q' \ug 1 \, , \; \l \ug 5$ , Epochs:~5.}\label{newinput2}
\end{figure}

We repeat the experiments of the previous section with the same results. In Fig.\ref{newinput},\ref{newinput2} we can see two examples with the same setting of Fig.\ref{so3} with a second order differential operator. In Fig.\ref{newinput} we update $\by$ with $\q' \ug 0.1$ whereas in Fig.\ref{newinput2} we use $\q' \ug 1$.

%
%

\subsubsection{Lack of supervisions}

\begin{figure}[H]
\hspace{-2.5cm}
\includegraphics[scale=0.63]{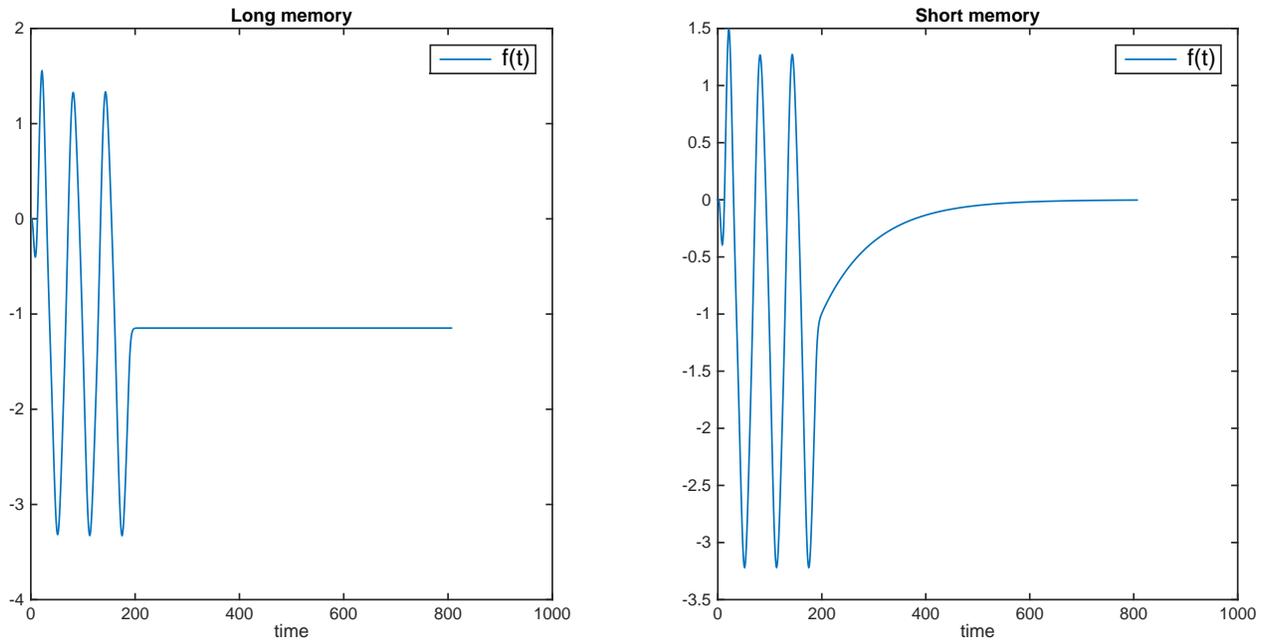}
\caption{Behavior of the system in two different settings , Epochs with supervision:~3, epochs without supervision: ~10. In the left plot a long memory parameters setting. In the right one, a system obtained with less memory. }\label{NoSup}
\end{figure}

In this section we analyze the reaction of the system when, after a period of training which lead to an oscillating stable state, we turn off the supervisions. We set a configuration which can fit data then, after some epochs of training, we let the time goes by without supervision. The idea is that the system should go to $0$ with a decay related to the memory of the Impulsive Response. In Fig.\ref{NoSup} we can see this behavior in two configurations with different memory.

%
%

\subsubsection{Random Data}

\begin{figure}[H]
\hspace{-2.5cm}
\includegraphics[scale=0.53]{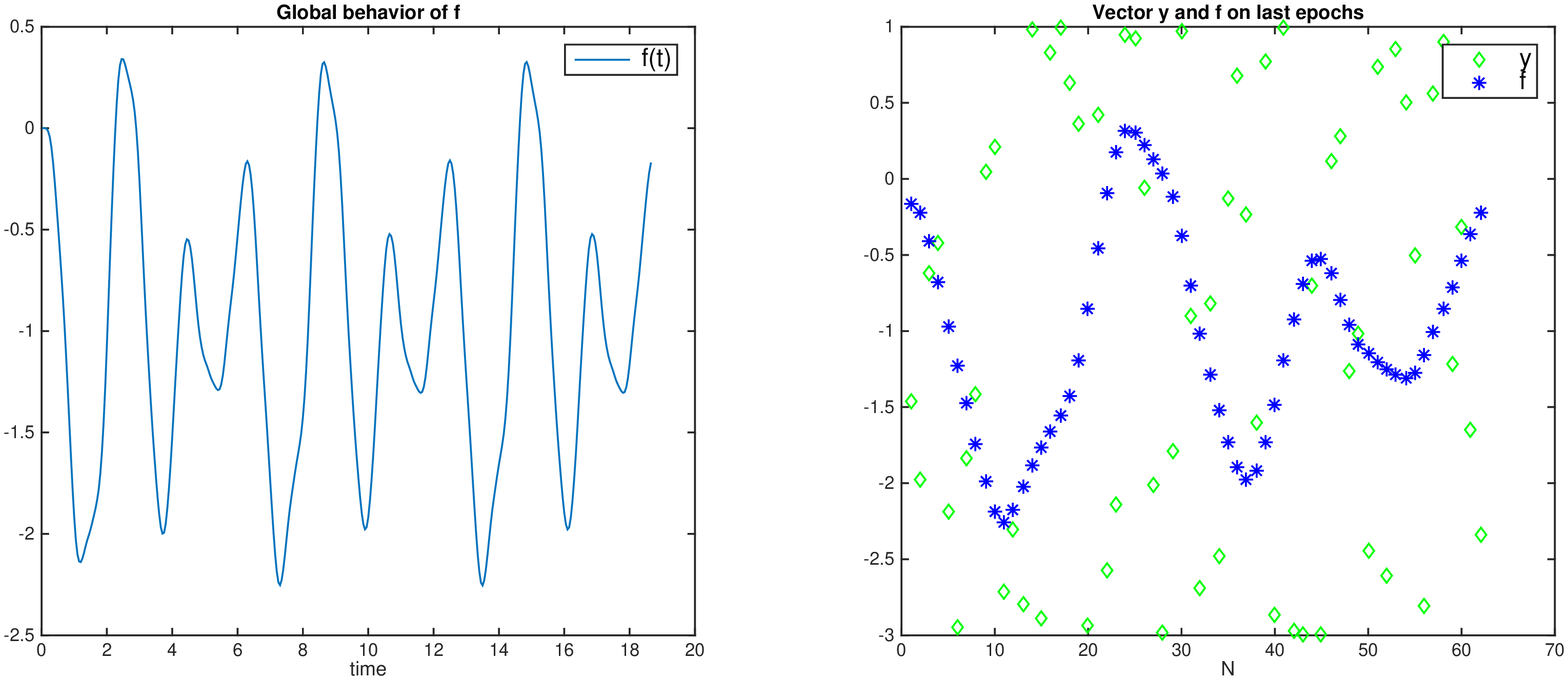}
\caption{Behavior of the system w.r.t. random data, repeated for 3 epochs. }\label{rand1}
\end{figure}

\begin{figure}[H]
\hspace{-2.4cm}
\includegraphics[scale=0.56]{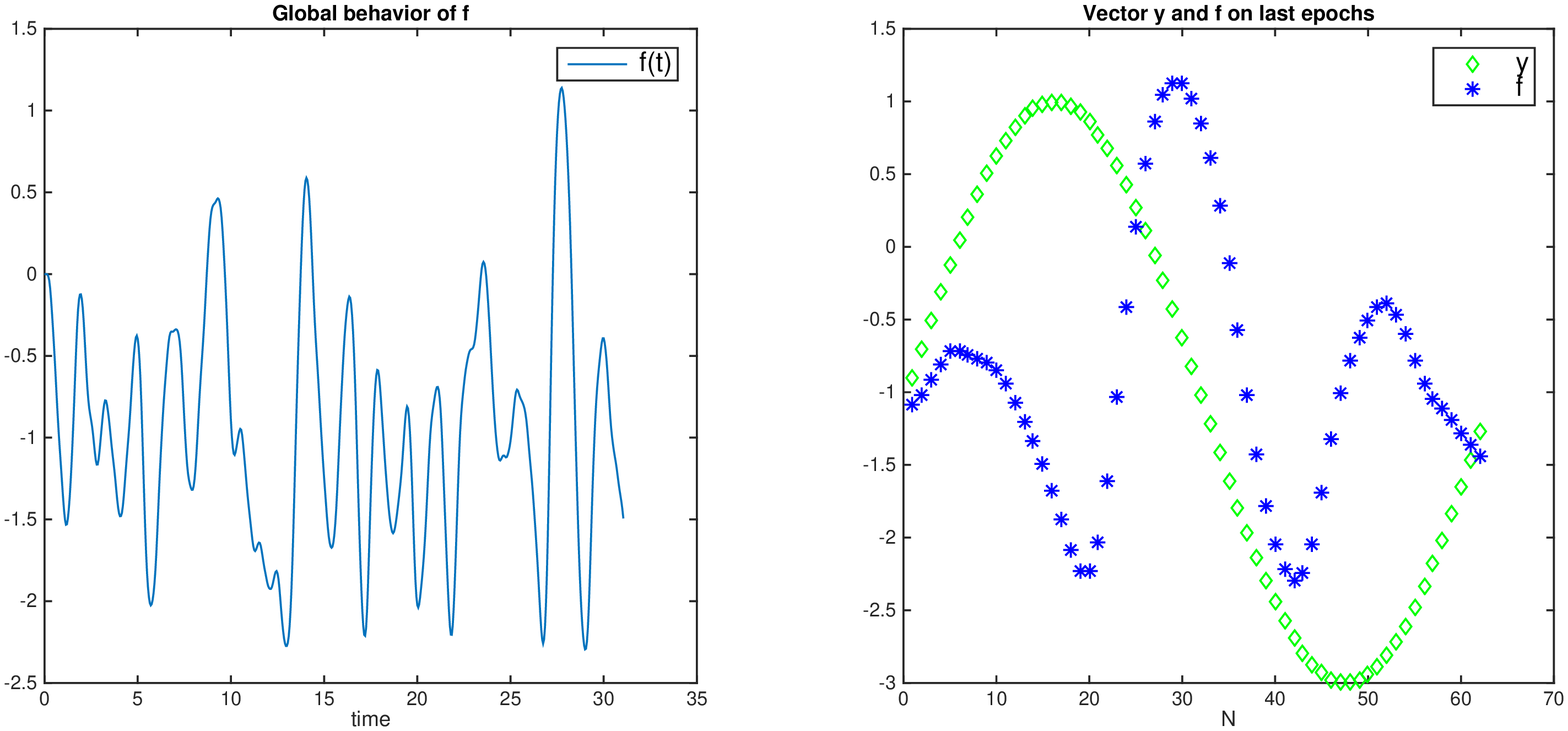}
\caption{Behavior of the system w.r.t. random data, permutation changed each times for 5 epochs. }\label{rand2}
\end{figure}

In this section we drop the time dependence of data. In Fig.\ref{rand1} we can see the response of the system when data are sorted randomly. The data are still cyclical in time, then we stil obtain a cyclical behavior. If we perform a random permutation of data at each epoch of training, we obtain the results of Fig.\ref{rand2}. The derivatives are calculated with the Finite Differences method.

\section{Causality Vs. Non-Causality}\label{integration}

In this section we develop a quick analysis on the computation of our solution.
Until now, we propose a forward updating of the system.
We assume the supervised instance to be distributed in time and the system to have a \emph{Causal} impulsive response, typical of the dynamic system.
This is a common characteristic of the time models, since the events occurred at a certain instant of time can not influence the past of the system.
With this idea, we performed an on-line updating of the function in time.
To reinforce the meaning of this choice, we report a results showed in \cite{papini}.
They prove that, when we restrict the search space the the one of periodic function, 
the solution obtained with the on-line updatings can converge to the one calculated by assuming to already know the events in the period $T$.
The condition which guarantees this convergence can be expressed by
\begin{equation}\label{convcond}		
C \left( 1 + \frac{1}{\l} \right) \left( 1 + \frac{C}{\l} \right)^{N-1} e^{- \b T} < 1
\end{equation}
where $C\geq 1$ is a constant and $\b >0$ is positively related to $\t$. 
As we can see, these conditions are in some way related to the dissipation and regularization terms ($\t$ and $\lambda$ respectively), and to the rate of impulses in $T$.
We shows the computation necessary to exploit this \emph{Global} solutions and checks the hypothesize convergence.
Once we assume to can see the global panorama of the analysed period, 
we compare these solutions with the one coming from the \emph{A-causal} impulsive response,
which intuitively should have the optimum solution (since optimize the function in the past too).
To sort out this comparing, we try to approximate the value of the functional (\ref{PHI2}) in each case.

\subsection{General Solution}

First, we sketch out the system to compute the global solution.
We pose $\nu \ug (-1)^k$ and consider a differential operator $Q$ such that:
\begin{equation}
\l \, P^* (\psi \, \sqrt{1 + |x'|^2} \,  P \by) = (-1)^k \l \psi \sqrt{1 + |x'|^2} \; Q \by
\end{equation}
we have that our solution $\by$ must satisfy:
\begin{equation}
Q \by = -\frac{\nu}{\l} \sum_{i \ug 1}^N \frac{ \by(t_i) - y_i }{\sqrt{1 + |x(t_i)'|^2}} \delta (t-t_i).
\end{equation}
The Green's function of $Q$ (i.e. a function $g$ such that $Q g \ug \delta$) is the same of the operator $P^* (\psi \, \sqrt{1 + |x'|^2} \,  P)$,  than we have: 
\begin{equation}
 \by (t)= \hat{f}(t) - \frac{\nu}{\l} \sum_{i \ug 1}^N \frac{ \by(t_i) - y_i }{\sqrt{1 + |x(t_i)'|^2}} g (t-t_i)
\end{equation}
where $\hat{f} \inn Ker(Q)$ and can be expressed as $\hat{f}(t) = \sum_{l=1}^{2h} c_l e^{\l_l t}$  ($h$ order of $P$). 
The explicit expression of $f(t)$ depends then on the coefficients $c_l$ and $f(t_i)$.
We obtain $N$ conditions by evaluating $f$ at the supervisions points:
\begin{equation}
 \by (t_j)= \sum_{l=1}^{2h} c_l e^{\l_l t_j} -\frac{\nu}{\l} \sum_{i \ug 1}^N \frac{ \by(t_i) - y_i }{\sqrt{1 + |x(t_i)'|^2}} g (t_j-t_i)
\end{equation}
which can be write as 
\begin{equation}
 \frac{\l}{\nu} \delta_{\small{ij}} \by (t_j) - \frac{\l}{\nu} \sum_{l=1}^{2h} c_l e^{\l_l t_j} + \sum_{i \ug 1}^N \frac{ g (t_j-t_i) }{\sqrt{1 + |x(t_i)'|^2}} \by(t_i) = \sum_{i \ug 1}^N \frac{   g (t_j-t_i)}{\sqrt{1 + |x(t_i)'|^2}} y_i.
\end{equation}
The remaining $2h$ conditions are free (for instance we can choose the classic Cauchy's conditions impose some condition on the boundary). 
We obtain a system of the kind
\begin{equation}\label{finalsyst}
 \M \bar{\f} = \left[ \frac{\l}{\nu} \left( I_N + C \right)+ G \right] \bar{\f} =  G \y 
\end{equation}
where $I_N, C, G \in \R^{(N+2h)\times(N+2h)}$,   
$\bar{\f}\ug [ \, \by(t_1), \cdots ,\by(t_N), \,  c1, \cdots ,c_{2h}\, ]'$ , 
$\y \ug [ \, y_1, \cdots ,y_N, \,  0, \cdots ,0\, ]'$, 
$I_N$ is the identity matrix with null elements from $I_{N+1,N+1}$. $C$ and $G$ are two block-matrix:
$$
\begin{array}{cc}
G =
 \left[
\begin{array}{cc}
G^g & 0 \\
G^c & 0 \\
\end{array}
\right]
&,\;
C=
\left[
\begin{array}{cc}
0 & C^g \\
0 & C^c \\
\end{array}
\right].
\\
\end{array}
$$
The matrices $G^g$ and $C^g$ are fixed:
$$
\begin{array}{lcl}
{G}_{ji}^g =  \frac{ g (t_j-t_i) }{\sqrt{1 + |x(t_i)'|^2}} & , & \quad j,i=1,...,N \\
&&\\
{C}_{ji}^g =  -e^{\l_i t_j} & , & \quad j=1,...,N \, , \; i=1,...,2h
\\
\end{array}
$$
The matrices $G^c$ and $C^c$ express the remaining $2h$ conditions.

As already said, because of the assumption in \cite{papini}, we are interested in the case of periodic functions. 
Then we can determine the system by imposing our solution to be periodic with the boundary conditions $f^{(s)}(0) = f^{(s)}(T)$ , $s\ug 0,...,2h-1$:
\begin{equation}\label{system}
\sum_{l=1}^{2h} c_l \l_l^s  -\frac{\nu}{\l} \sum_{i \ug 1}^N \frac{ \by(t_i) - y_i }{\sqrt{1 + |x(t_i)'|^2}} g (-t_i)  = \sum_{l=1}^{2h} c_l \l^s e^{\l_l T} -\frac{\nu}{\l} \sum_{i \ug 1}^N \frac{ \by(t_i) - y_i }{\sqrt{1 + |x(t_i)'|^2}} g (T-t_i)
\end{equation}
\subsection{Examination of convergence}

By solving the system in (\ref{system}), we will find the optimal solution for our problem. 
A fundamental parameter is the function $g$. 
As already said, we obtain the best solution by considering its non-causal version. 
In this section, we try some experiments to verify that the solutions proposed in Section \ref{implementation}, 
converge to the one obtained by finding the coefficients of $f$. 
This solution lead to possibly numerical problems. 
The condition which guarantee the desired convergence is (see \cite{papini}):
\begin{equation}\label{convcond}		
C \left( 1 + \frac{1}{\l} \right) \left( 1 + \frac{C}{\l} \right)^{N-1} e^{- \b T} < 1
\end{equation}
where $C\geq 1$ is a constant and $\b >0$ is positively related to $\t$.
The parameters $\l,\n, N,\, \t,\, T$ can be used to guarantee this convergence and by maintaining a good fitting.
However, the condition number of the matrix $\M$ in (\ref{finalsyst}) impose some constraints coming from the exponential nature of the impulsive response. Relatively slightly changes on $\t$ produce a $g$ which reach too bigger or too small values in the positive axis. Moreover, since in the negative part of its domain the non-causal version of $g$ is divergent, if we enlarge $T$ we quickly reach unmanageable values of $g$. These two facts force to values of $N$ and $\l$ which allow a lousy fitting. Luckily, these problems are avoided with the forward solution studied in Section \ref{implementation} where the parameters were free. For these reasons, we have to limit the comparing in this Section to some non-optimal fitting cases.

\section{Temporal links}

The temporal-oriented approach presented gives an incomplete description of the environment.
Indeed, our agent seems to lose the spatial references of data as time goes by.
This because the function weights the spatial distance between to consecutive instances (via $b(t)$), 
but do not take in account any other information regarding direction or the positions of the previous points. 
In this part, we try to face this problems by saving links among the examples which are similar in terms of spatial-features.
The first idea is to build a ERN, 
representing the concept of \emph{neighborhood} of samples in both temporal and spatial sense. 
We still suppose our agent travels in the input features space continuously in time 
and the output function $\by$ to be updated with the same time sampling step $\q$.
Now, we let the agent takes in accounts of its previous predictions in the correspondent region of features space.
Then, it receives from these informations a sum of impulses in a way comparable to the external supervisions.
To implement this model, we develop time-wise a graph $\cG \ug (\Rg,\Eg)$, 
whose nodes represent points of the input features space.  
At each step, 
The agent looks for a node similar to the current input in $\Rg$, if no one fits, a new node is added.
Then it adds a binary link between the current node and the last node visited, expressing their temporal correlation.
At the same time, it inserts weighted edges on $\Rg$ depending on the distance. 
Then $\by$ is evaluated and saved in the node (updated if the input correspond with an existing vertex).
If external supervision is provided too, this value is saved in the node instead of $\by$.
Once a supervision is saved in a node, its value will be changed only if another supervision comes (and their values averaged).
We express a general time instant as $t_k \ug k \q, \, k\inn \bbN$ and determine the node $\rg_k \inn \Rg$ from $x_k \ug x(t_k)$ as:
\begin{equation}
\displaystyle
\rg_k = 
\left\{
\begin{array}{l}
r_{k-1}  \, , \quad \mbox{ if } \| x_k - r_{k-1} \| \leq \ee_{k-1}  \, , \;  \ee_{k-1} \inn \bbR \vspace{0.3cm}\\
\hspace{-0.1cm}
\begin{array}{ccl} 
\rg^* = &  \arg \min  &   \| x_k - \rg_j \| \, , \quad  \mbox{ if } \exists j: \; \| x_k - \rg_j \| \leq \ee_j  \, , \;  \ee_j \inn \bbR \\
  \vspace{-0.6cm}  \\
   &\mbox{\tiny $r_j \in R_G$ } & \\
   \end{array} \\
x_k \; \mbox{ added in $\Rg$ otherwise} .\\
\end{array}
\right.
\end{equation}
The parameter $\ee$ represent the radius of the spheres of our graph, 
the index $j$ means that it could changes depending on the density of the region around each node.
We indicate the value of $\by(t)$ saved in the node $\rg_k$ as $\by_k$.
The value of $\by (t)$ is update in a similar way of what seen in the previous Sections.
The function is still calculated as the solution of the Eulero-Lagrange equations of a functional of the kind of (\ref{PHI2}), 
with domain $[0,\, T]$ (where $T \ug \q K,\; K \in \bbN$), 
modified to implement these new constraints.
Again, we restrict the analysis to the case $\mu \ug 0$.
The regularization term:
\begin{equation}\label{Kt} 
\cK(\by) = \l \int_{0}^T (P \by)^2 \psi \, b \, dt
\end{equation}
is unchanged. 
The term expressing the temporal correlation, indicated from the binary links between consecutive nodes, can be write as:
\begin{equation}\label{Ut} 
\cU(\by) = \eta \sum_{k=1}^{K}  \left(       \frac{ \psi (t_k)}{\hat{w}_k} \sum_{j \ug 1}^{\left| R_G \right|} \hat{w}_{kj} | \by(t_k) - \by_j |^2       \right)
\end{equation}
where $\hat{w}_{kj}$ represent the number of  times that the node $r_k$ succeeds the node $r_j$. 
Since we assume to  increment by one the weight $\hat{w}_{kj}$ each time that $r_j$ precede $r_k$, 
we add the normalization parameter $\hat{w}_{k}$ among the contributions: 
$ \hat{w}_{k} \ug \max_j \hat{w}_{kj}$. 
The term expressing the spatial correlation, is composed by the contributions of the closest nodes in the features space: 
\begin{equation}\label{Ut} 
\cS(\by) = \g \sum_{k=1}^{K} \left(       \frac{\psi(t_k)}{\rho}  \sum_{s =1}^{\rho} w_{k\iota_s} | \by(t_k) - \by_{\iota_s} |^2       \right)
\end{equation}
where the set $\cN \ug \{ r_{\iota_1},...,r_{\iota_{\rho}}\}\,,\; \rho\inn \bbN_{ \left| R_G \right|}$ express the nodes which we want to consider in the neighborhood, 
sorted according to the distance weights $ w_{k\iota_s} $, 
which could be calculated for example as
$$w_{k \iota_s} = e^{-\frac{\norm{x_k - x_{k\iota_s}}^2}{2\sigma^2}} $$
where $\sigma$ is the mean of the Euclidean distance between samples.
This value changes when new nodes are added to $\cG$, 
but parameters as this one and $\ee$ can be estimated on an initial portion of data and then fixed to speed up computation.
The penalty term in (\ref{PHI2}) expressing the external supervision is the same,
we only have to adapt the index of the summation as:
\begin{equation}\label{Vt} 
\cV(\by) =  \sum_{k \ug1}^N \psi (t_k) | \by(t_k) - y_k |^2 .
\end{equation}
We add the parameters $\l,\, \eta,\, \g$ to tune the global contribution of each terms, 
even if we will assume $\g \ug 1-\eta$.
The new functional $\Phi_{\mathcal{S}}(\by)$ is composed by the summation among these four terms and generates the following Eulero-Lagrange equation:
 \begin{equation}
\begin{array}{ccc}\label{ELst}
\displaystyle
 \l \, P^* (\psi \, b \,  P \by)  + \sum_{k\in \cI} \psi (t_k) \left[ \by (t_k) - y_k \right] \delta (t-t_k)  + & & \\
 \displaystyle
 + \sum_{k=1}^{K} \psi(t_k)  \left( \frac{\eta}{\hat{w}_k} \sum_{j \ug 1}^{\left| R_G \right|} \hat{w}_{kj}  \left[ \by(t_k) - \by_j \right] + \frac{(1-\eta)}{\rho}  \sum_{s=1}^{\rho}  w_{ks} \left[ \by (t_k) - \by_s \right]  \right) \delta (t-t_k) & = & 0 
\end{array}
\end{equation}

\subsection{Implementation of the new model}

To easily solve (\ref{ELst}) we use the same approach of Section \ref{implementation}. 
From a theoretical point of view, 
we assume possible both external supervisions and the propagation of the informations from updating of $\cG$
to happen in the middle of two updating of $\by$. 
This little shift allow us to solve the equation with the Lagrange formula with a forward solution which can be easily evaluated on-line.
Starting from the Cauchy's conditions:
\begin{equation}
\f[0] = [\, \by(0)\, , \; \by' (0) \, , \cdots , \; \by^{(2h-1)}(0) \,]'
\end{equation}
the new value $\f[k+1] \ug \f (\q(k+1)) $ is evaluated by:
\begin{equation} \label{UpFoST} 
\f [k+1]  =   e^{\A \q} \f [ k]  +
 e^{\A [ \q /2]} \cdot \B \frac{1}{\l \a_{2h}^2 b(t_{k}) }  \E[k]     
\end{equation}
where we posed
\footnote{As in \RefSec{implementation} we calculate $\tilde{f}(t_{k})$ by performing an intermediate updating $\tilde{\f}(t_{k}) \ug e^{\A [\q /2]} \f [k]$.}
\begin{equation}\label{Ek}
\E[k] = 
 \left(
 \Delta_{k}+ 
\frac{\eta}{\hat{w}_k} \sum_{j \ug 1}^{\left| R_G \right|} \hat{w}_{kj}  \left[ \tilde{f}(t_k) - \by_j \right] + 
\frac{\g}{\rho}  \sum_{s=1}^{\rho}  w_{ks} \left[ \tilde{f}(t_k) - \by_s \right] 
 \right)  
\end{equation}
and
$$
 \Delta_{k}=
 \left\{
\begin{array}{l}
\tilde{f}(t_k) - y_l \, , \; \mbox{ if } \exists l : (u(t_l),y_l) \mbox{ is provide at } t_l \in \left[ t_k, t_{k+1} \right] \\
0 \mbox{ otherwise}
\end{array}
  \right.
$$

\subsection{Experiments}

\subsubsection{Vowels}
To face classification tasks, we decide to replicate our model for each class. 
These functions share the same graph $\cG$, 
but each class is described by a one dimensional output function, 
so as to evaluate the global predictor at each step as the max over all them. 
The supervisions are expressed as a binary target, 
the function correspondent to the class of the supervised point receives target $1$, 
while the others receives $0$.
To evaluate this model,
we will observe the classification Accuracy achieved on the Test Set, 
once the model as been train with the Training Set.

We analyze the behavior of this kind of structures on a vowels classification task. 
We recorded two different audio tracks with the sequential pronunciation of the five Italian vowels. 
The audio signal was processed to extract $40$ auditory spectral coefficients using the RASTA PLP algorithm\footnote{\texttt{http://labrosa.ee.columbia.edu/matlab/rastamat/}}.
We use a first track (about $20$ second with one utterance of each vowels, $2053$ total samples) to train the model, 
with $5$ supervised instances per class.
A second track of about $20$ seconds, 
with alternative pronunciation of the vowels, 
is used as Test set, with about $250$ labels per class (total of $2144$ samples).

\begin{figure}[H]
\hspace{3cm}
\includegraphics[scale=0.5]{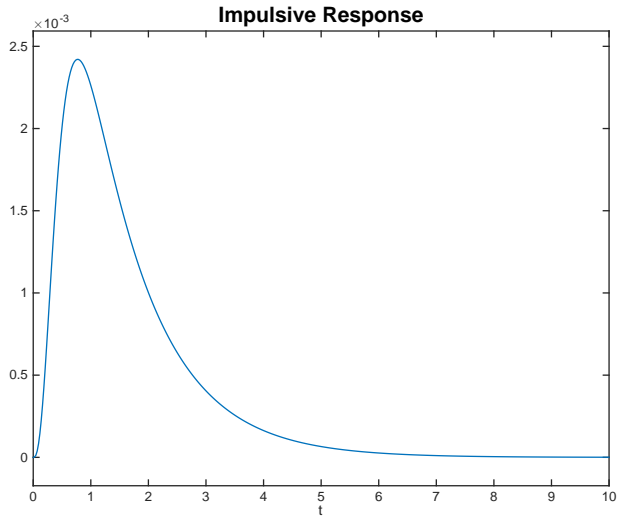}
\caption{Impulsive response used in the experiments of this Section, obtained by $\t \ug 10$, $\a_0 \ug \a_2 \ug 1$ and $\a_1 \simeq 0.99$.}\label{MNIR}
\end{figure}

It is quite obvious that the performances of this system are strongly influenced by the resolution of $\cG$. 
That is, a larger number of nodes allows better prediction performance but also slower computational efficiency.
Then, we skip the study of this parameter, whose effects are known and depend on data.
We fix $\ee \ug 2$, this allow to save about $200$(10\% of total)  nodes in $\cG$.
The differential equation describing the functions is set to give the impulse response of \RefFig{MNIR}, 
which has a quick response and decay interval such that propagates the informations not too much in time, 
In this way the function of a class extinguish after few points, 
unless other impulses of the same kind arrives.

\begin{figure}[htbp]
\resizebox{\textwidth}{.47\textheight}{\hspace{-0.5cm}
\begin{tabular}{ccc}
\includegraphics[scale=0.4]{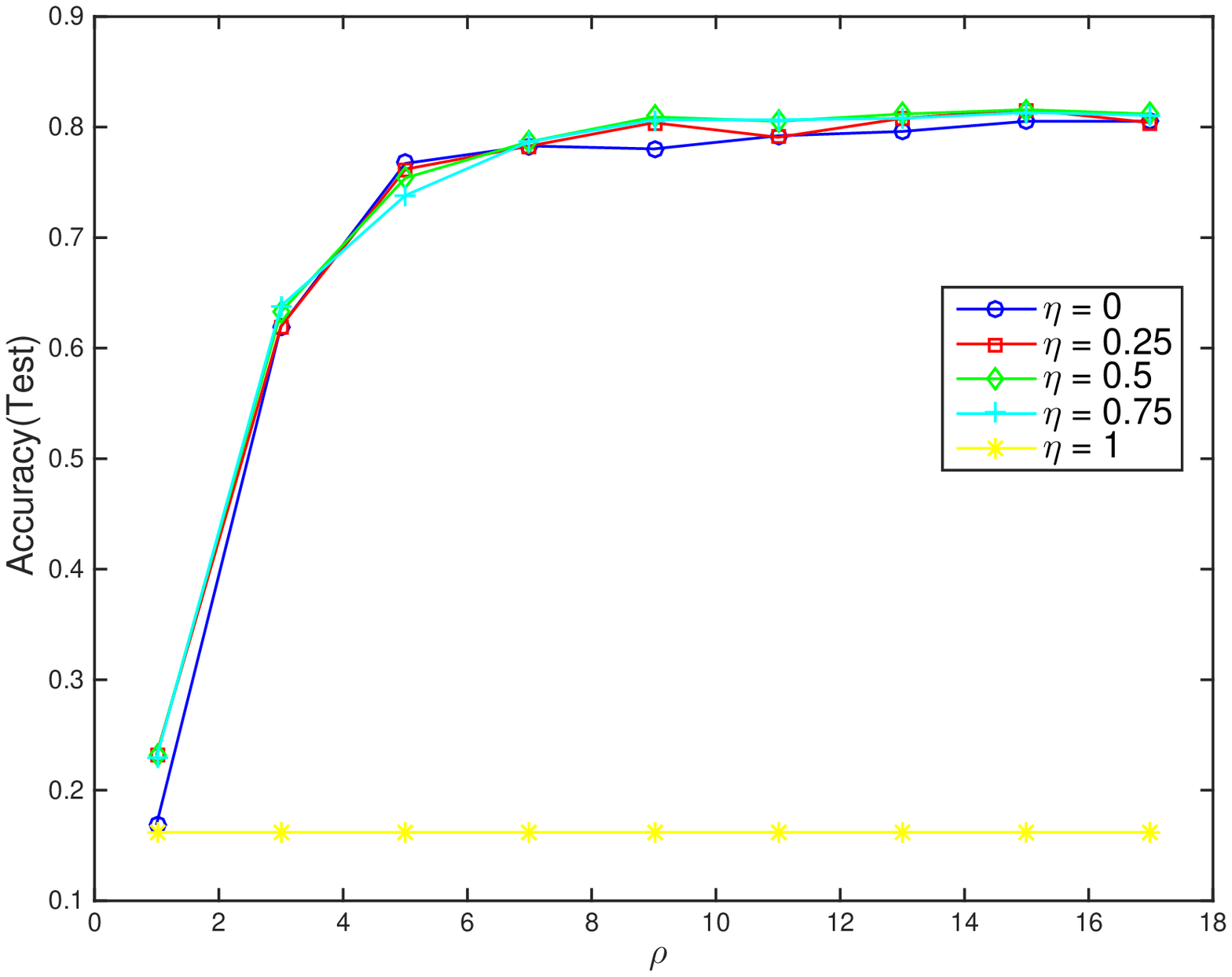} & \includegraphics[scale=0.4]{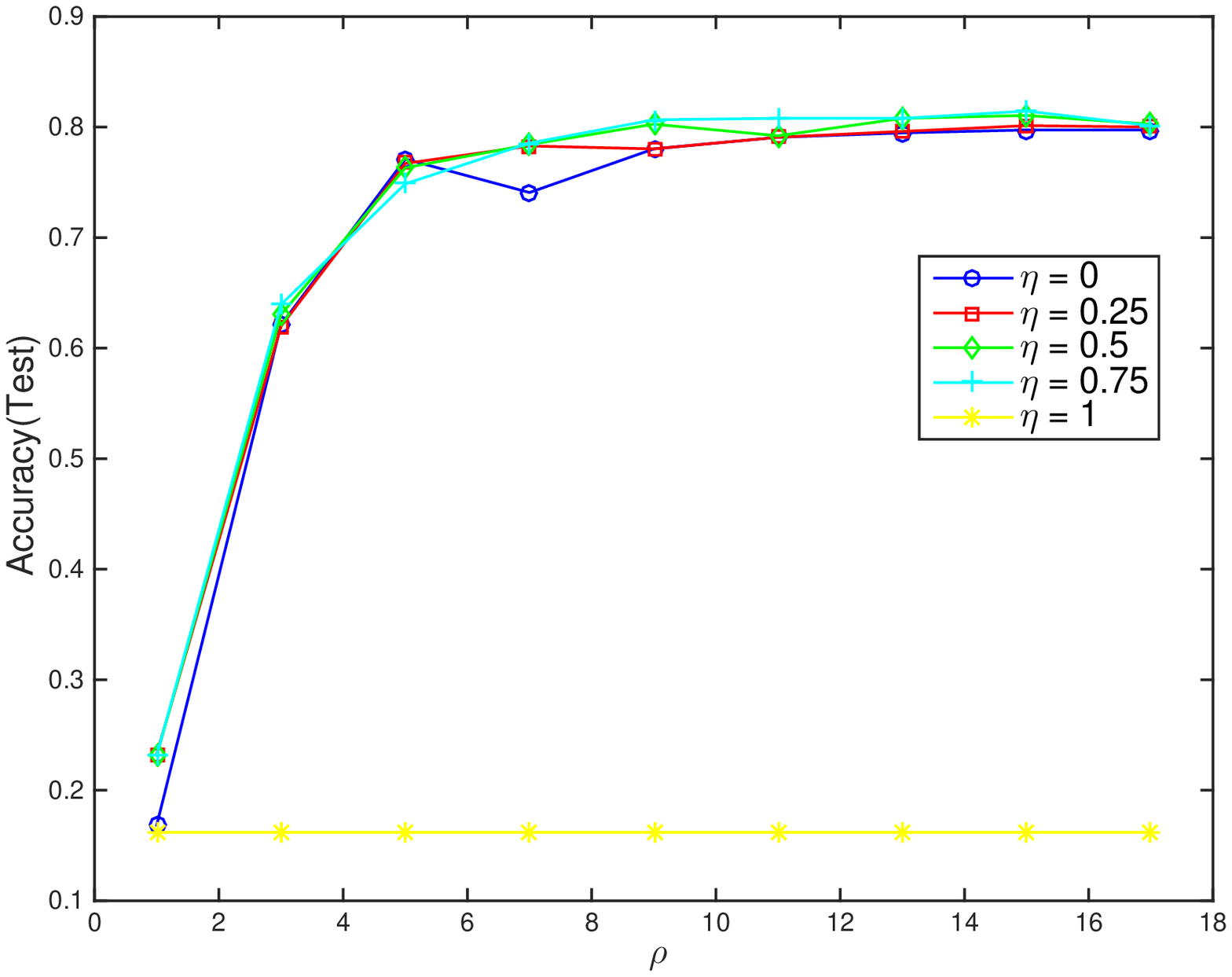} & \includegraphics[scale=0.4]{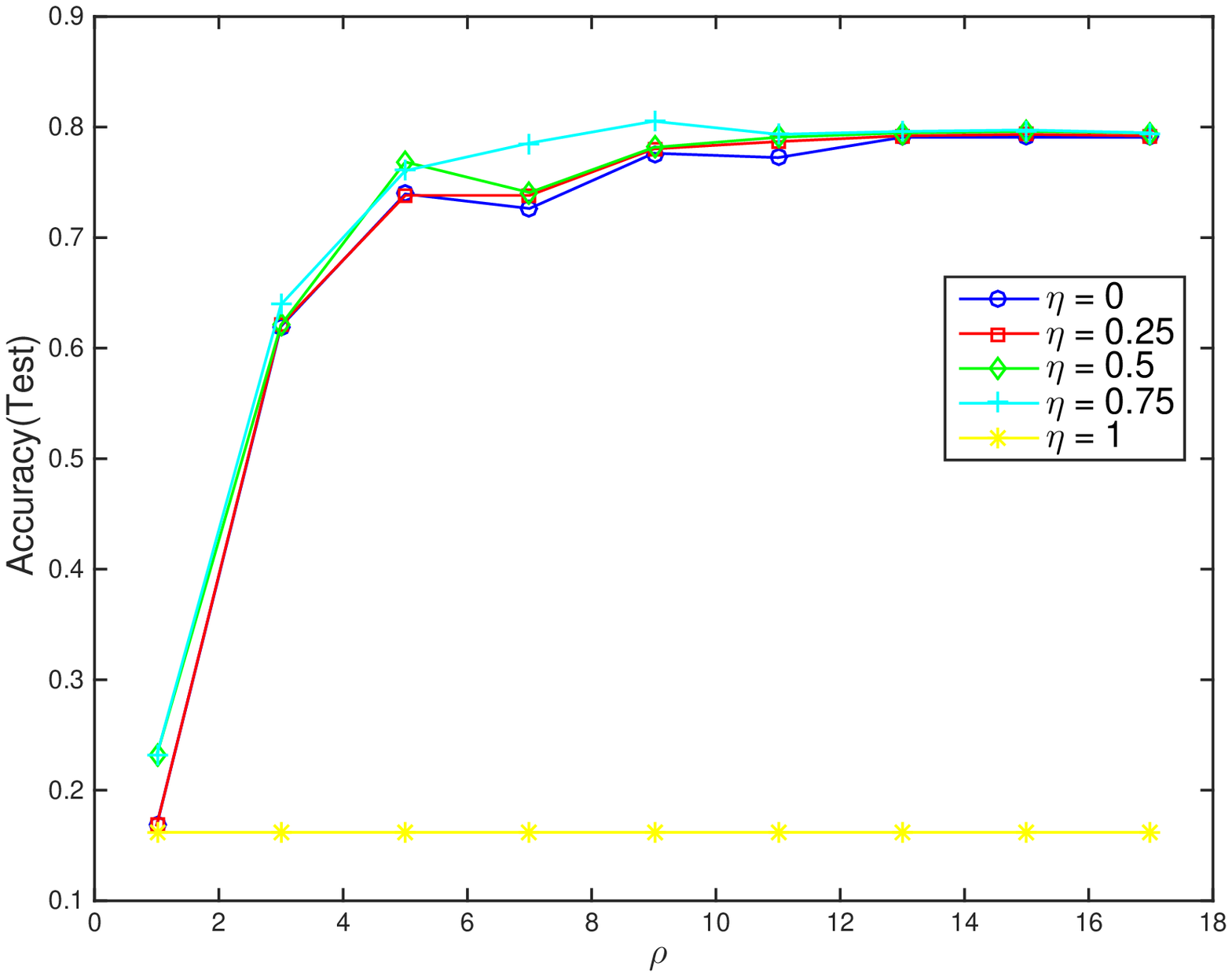}  \\
(a) $\q = 1\, , \; \l = 0.01 $ & (b) $\q = 1\, , \; \l = 0.007 $ &  (c) $\q = 1\, , \; \l = 0.004 $ \\
 \includegraphics[scale=0.4]{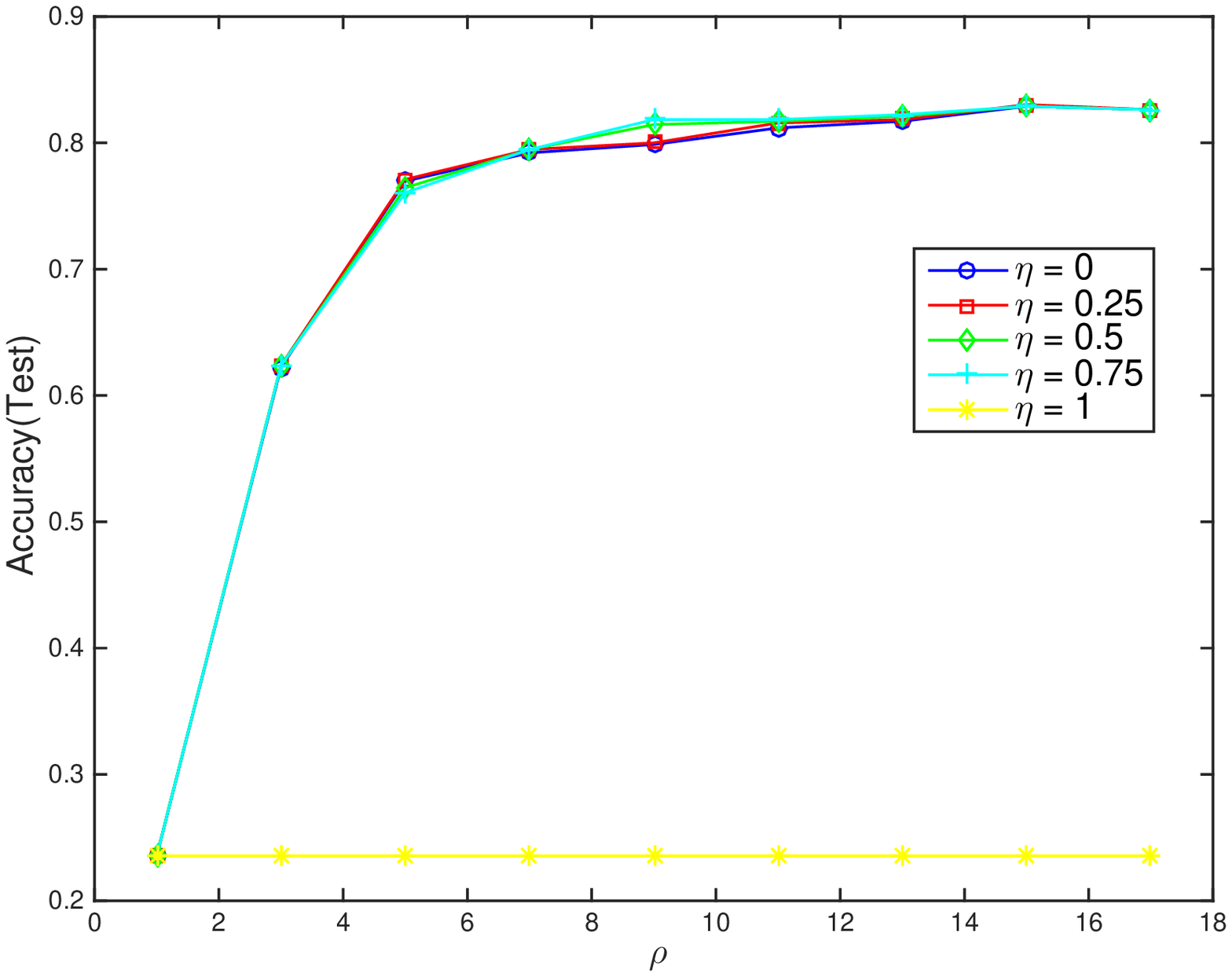} & \includegraphics[scale=0.4]{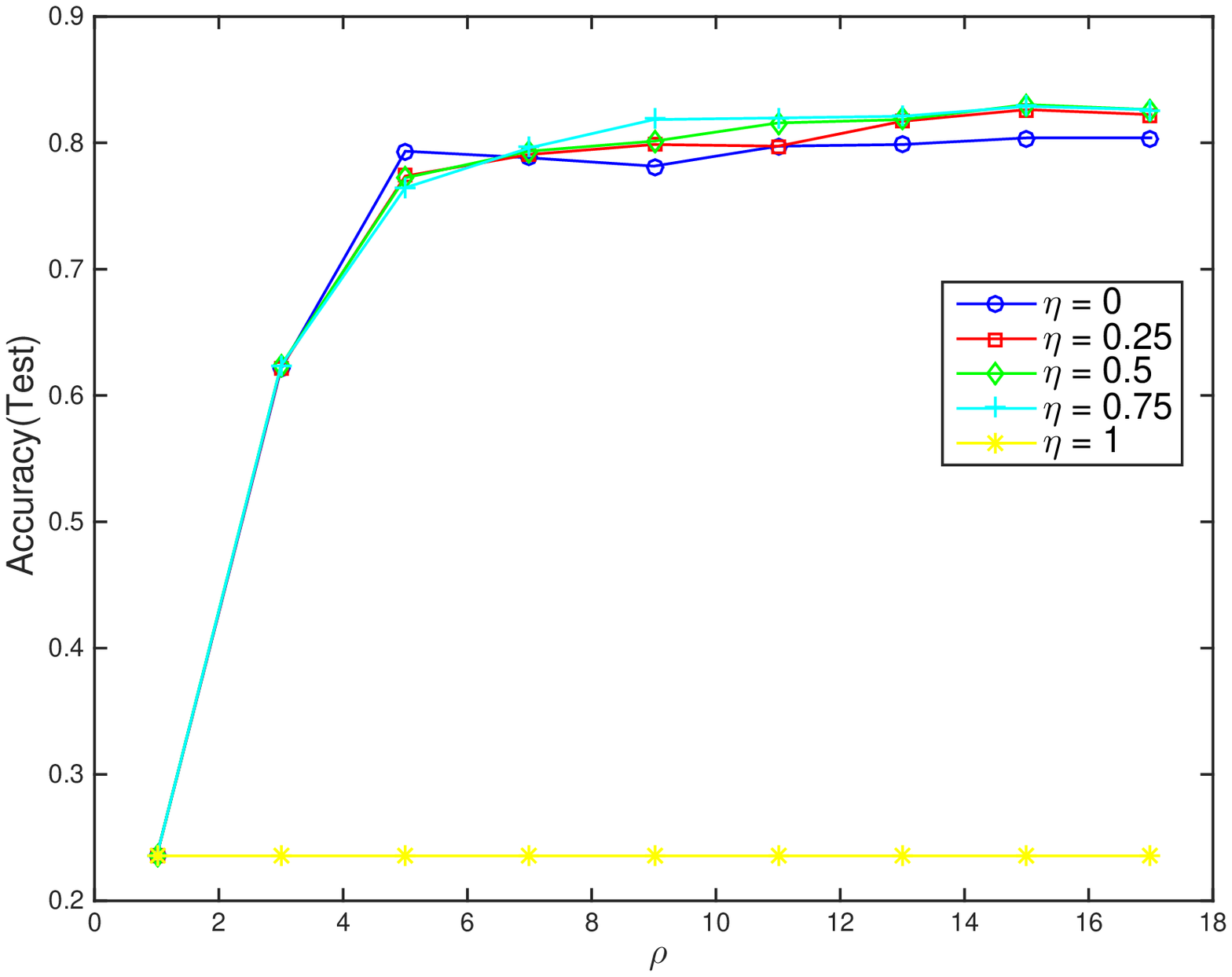} & \includegraphics[scale=0.4]{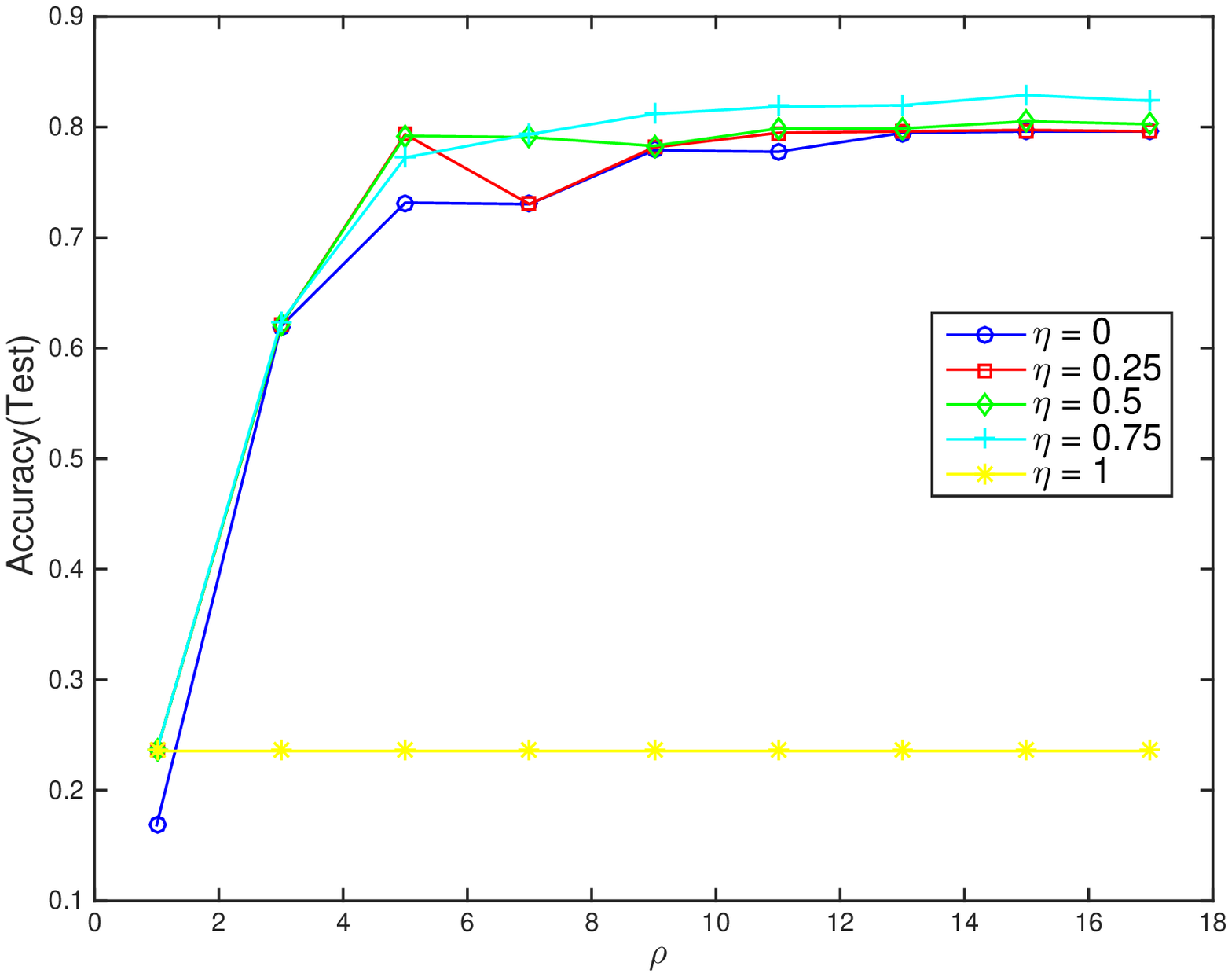}  \\
 (d) $\q = 2\, , \; \l = 0.01 $ & (e) $\q = 2\, , \; \l = 0.007 $ &  (f) $\q = 2\, , \; \l = 0.004 $ \\
 \includegraphics[scale=0.4]{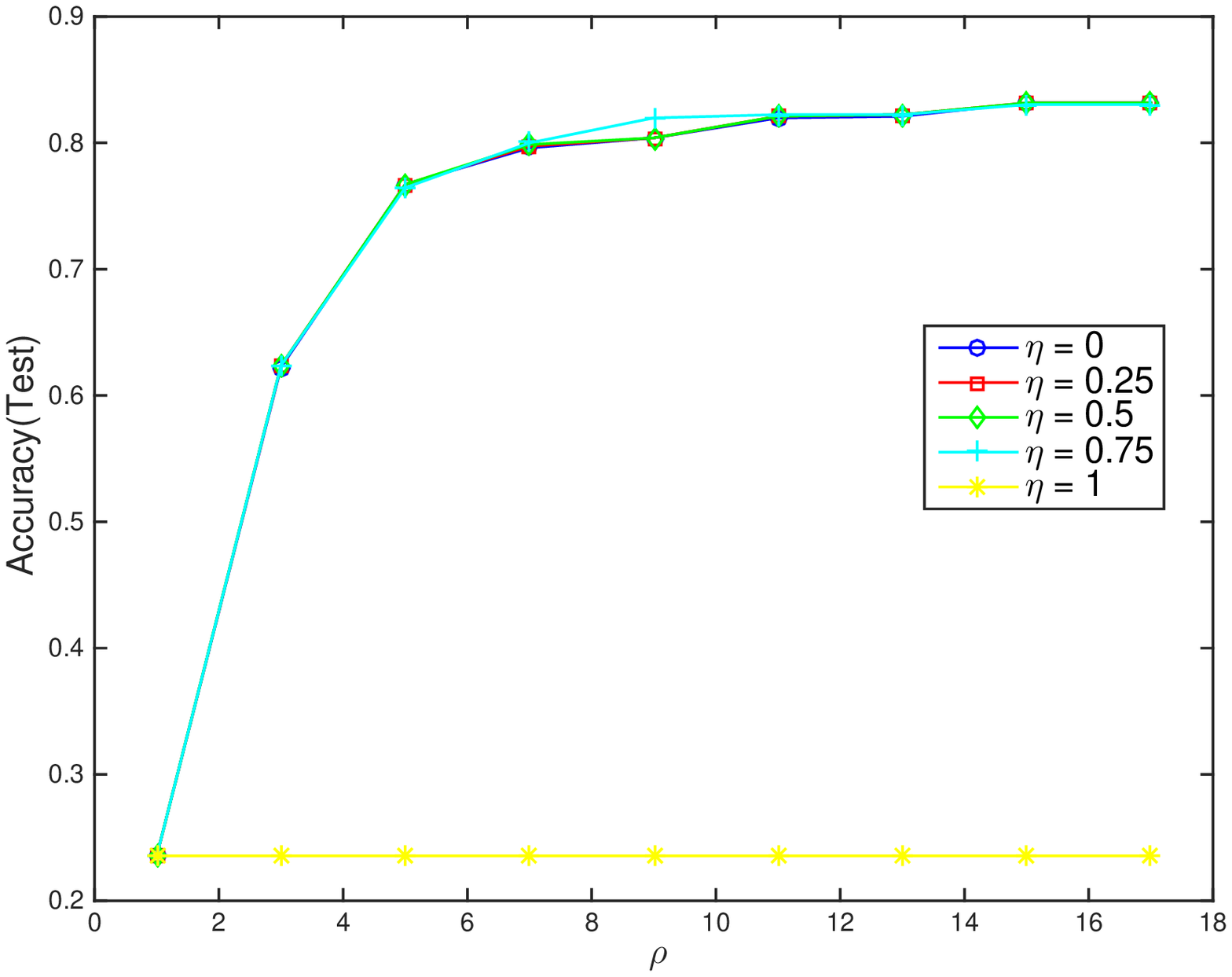} & \includegraphics[scale=0.4]{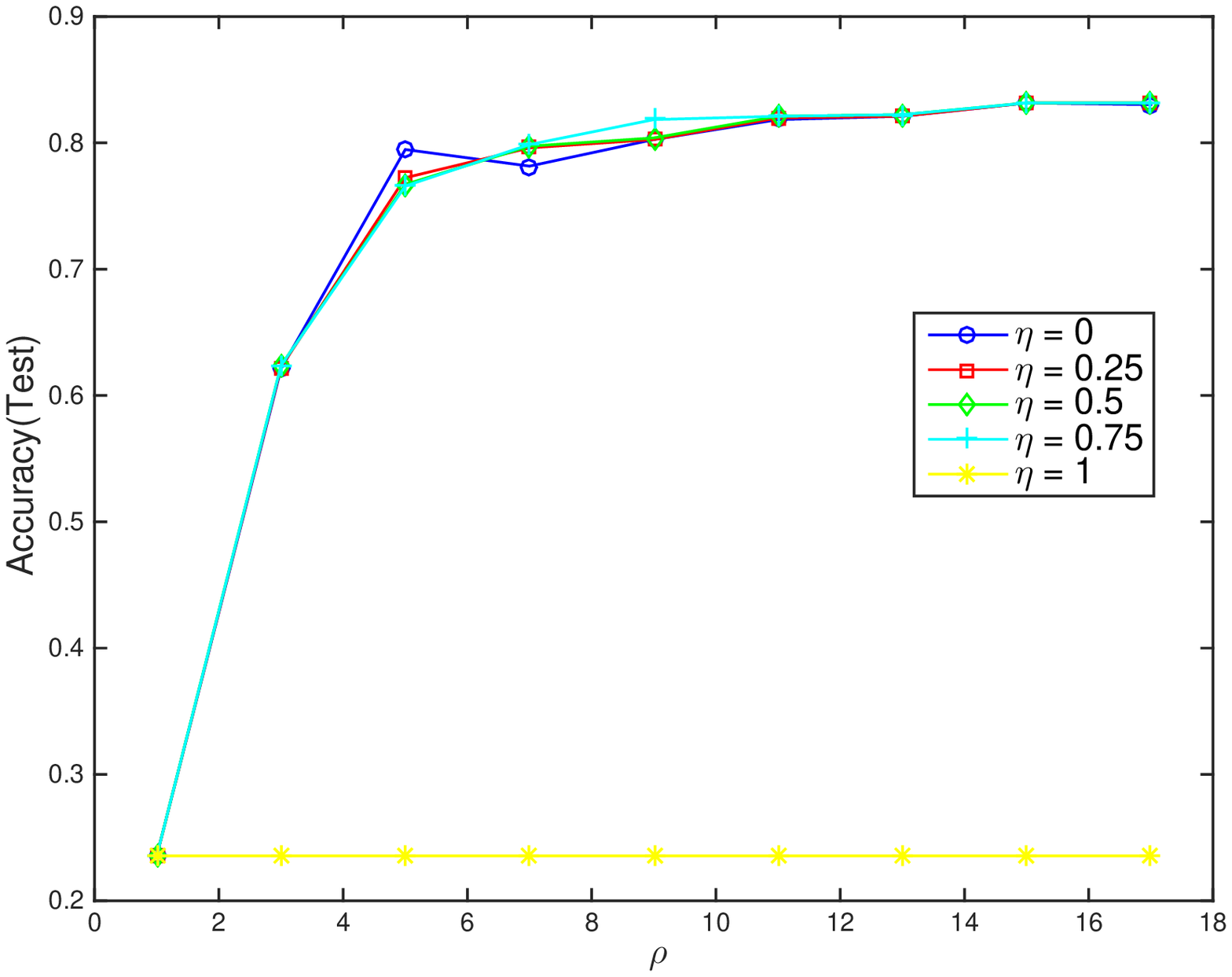} & \includegraphics[scale=0.4]{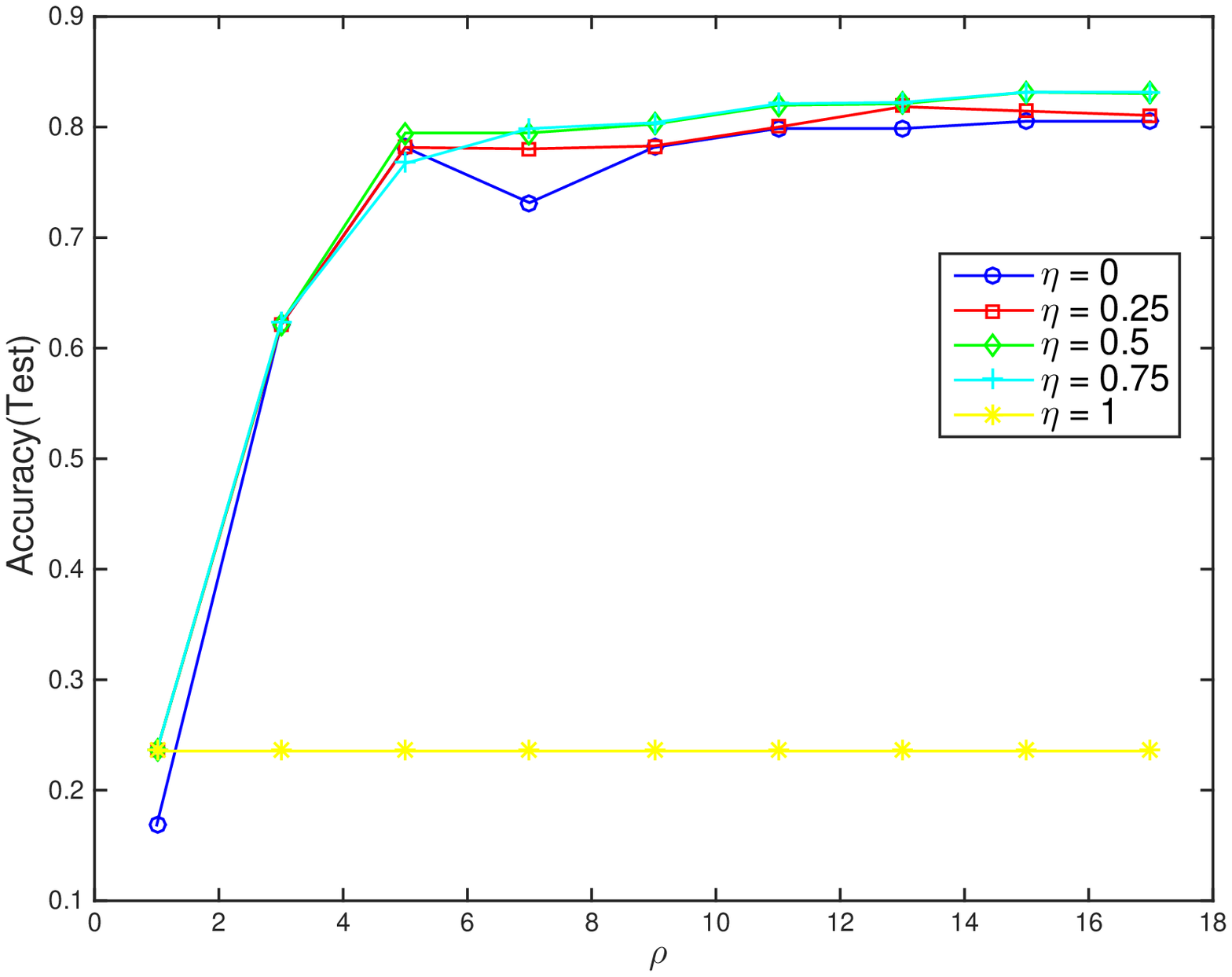}  \\
 (g) $\q = 4\, , \; \l = 0.01 $ & (h) $\q = 4\, , \; \l = 0.007 $ &  (i) $\q = 4\, , \; \l = 0.004 $ \\
 \includegraphics[scale=0.4]{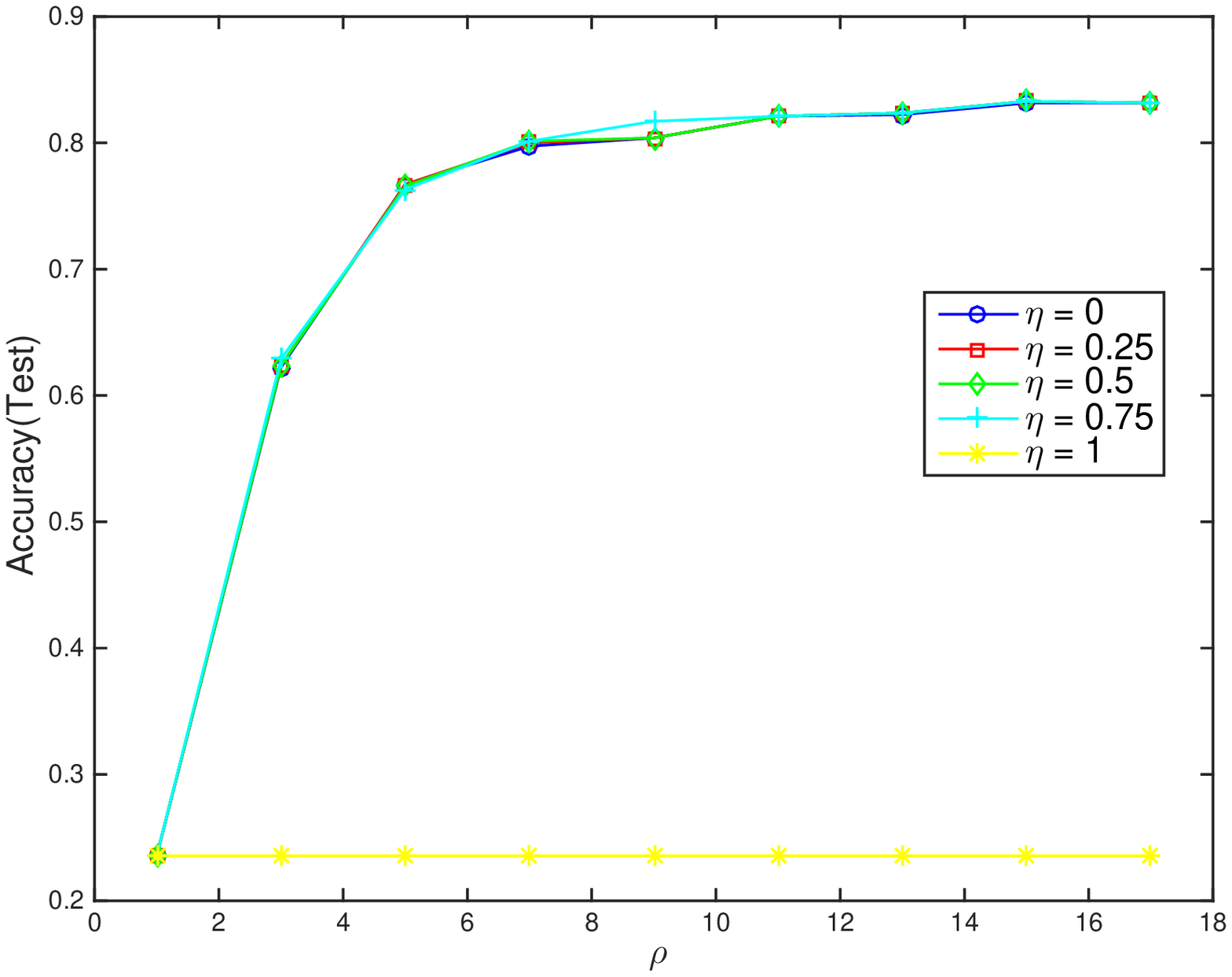}  & \includegraphics[scale=0.4]{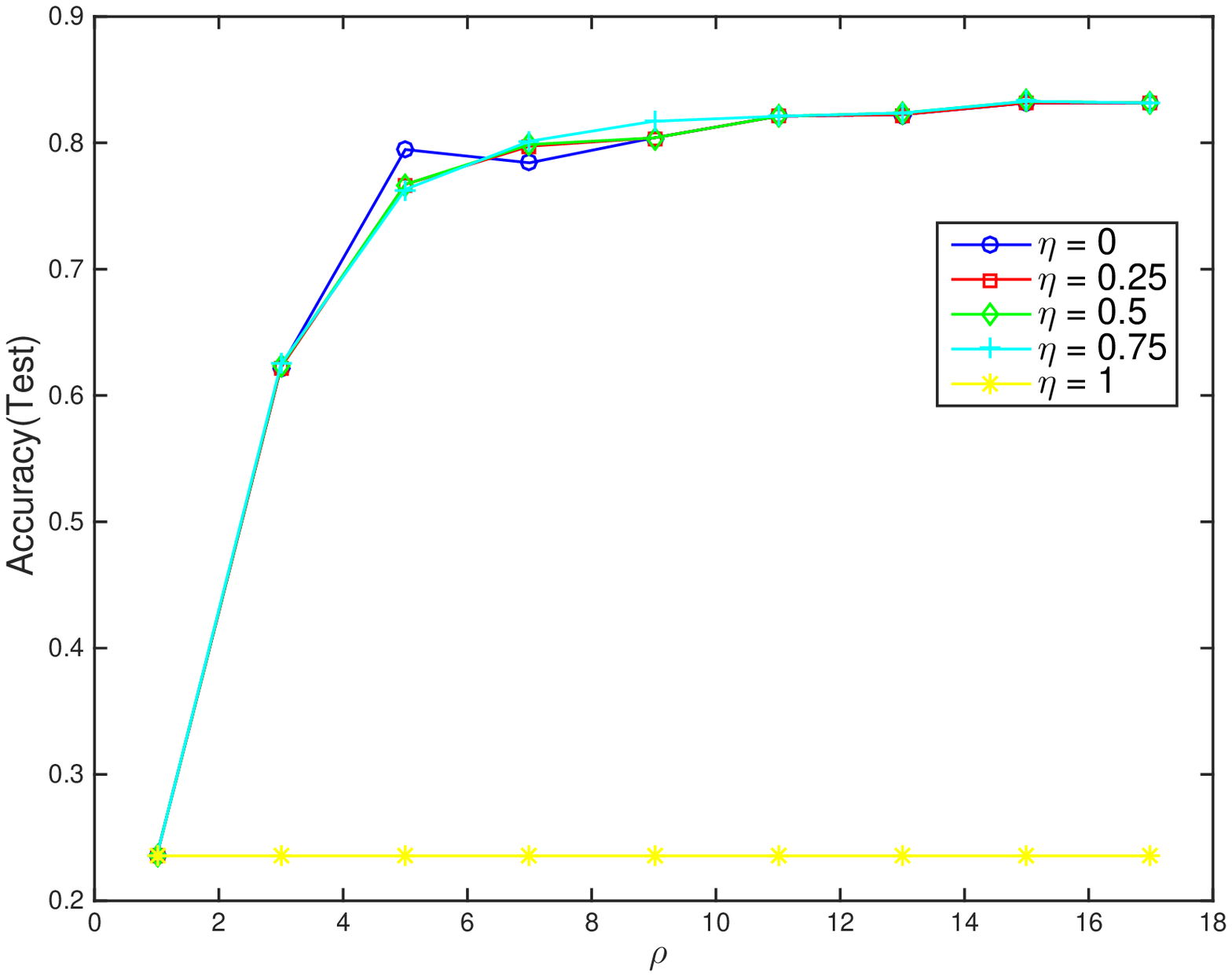}   & \includegraphics[scale=0.4]{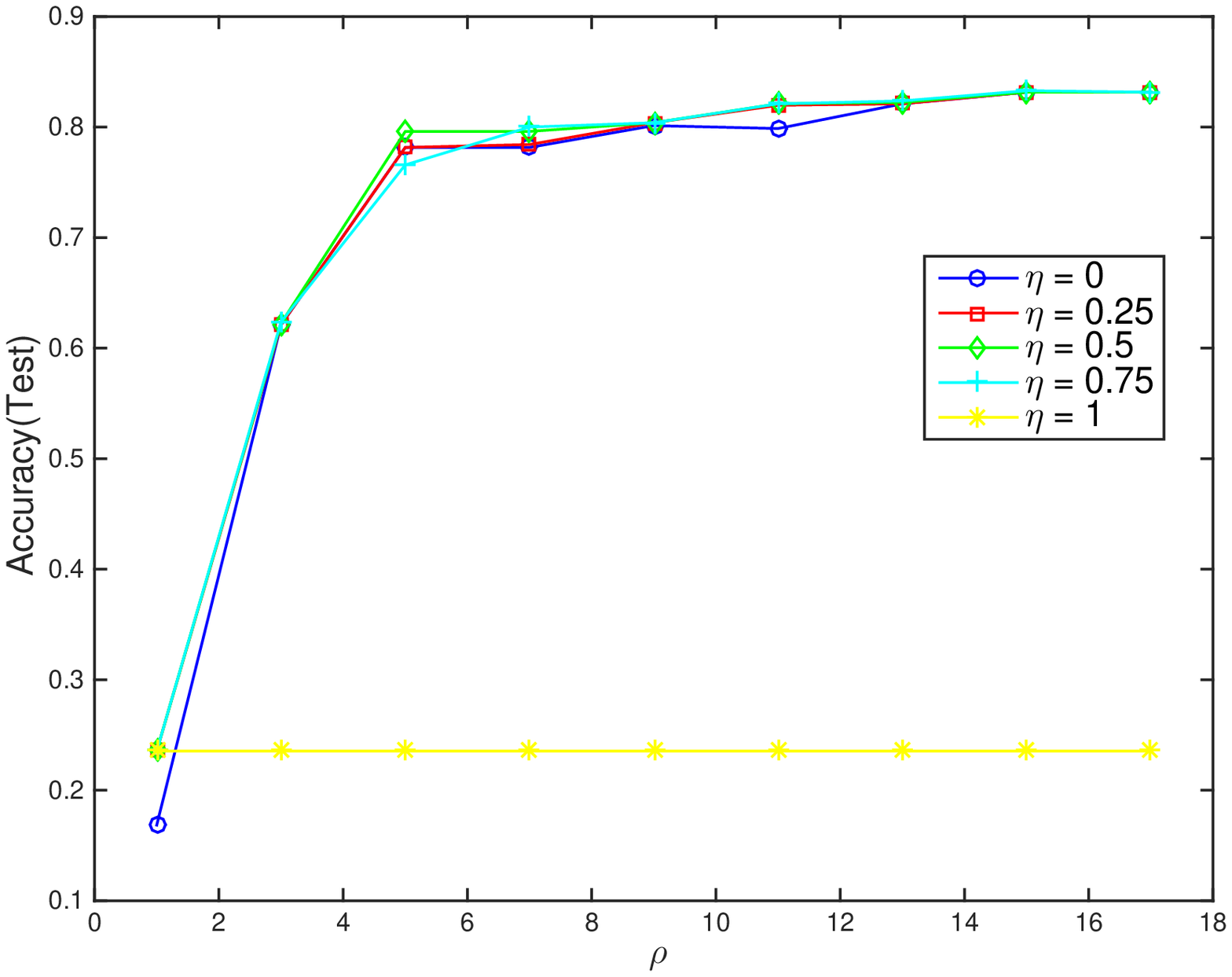}  \\
 (l) $\q = 6\, , \; \l = 0.01 $ & (m) $\q = 6\, , \; \l = 0.007 $ &  (n) $\q = 6\, , \; \l = 0.004 $ \\
\end{tabular}}
\caption{Accuracy vs. $\rho$ (number of spatial neighbors used) for the vowels classification task, $5$ supervision per class.
In each plot, different line correspond to different values of $\eta$ (balancing between temporal and spatial contributions).
Different value of $\q$ in each line.
Different value of $\l$ in each column.\label{vowels1}}
\end{figure}

We train the model on 3 repetitions of the total sequence, to allow the supervisions to be propagated to the previous samples.
The results are visualized in \RefFig{vowels1}.
We plot the prediction accuracy reached on the test set, versus the number of nodes considered in the spatial contribution.
The different line's colors mean different balancing between temporal and spatial contributions (different $\eta$).
The rows we report the results with different values of $\q$, varying in $1$, the minimum step allow the impulse to respond, $3$ and $6$, from which the value of the response are approaching $0$.
Since the convergence of the system is related to the number of impulses in the period, in the left column we fix $\l \ug 0.01$, a value which guarantee convergence for every different setting uses. In the right column, $\l$ is minimized in each configuration to guarantee both convergence and best fitting performance.  

\subsubsection{0 vs 1 MNIST Experiments}

In this section we try some experiments on the MNIST dataset,
in order to compare the results with other algorithms (\cite{olap},\cite{Gold}) and to see the behavior of our model when supervisions decreases.
We start with the three class of digits 0, 1 and 2 (scaled down to $16 \times 16$ pixels following \cite{Gold}). 
The original training set provides $5923$ elements for class 0, $6742$ for class 1 and $5958$ for class 2.
 The test set contains $980$, $1135$ and $1032$ instances respectively.
Because of our assumptions on the time correlation of data,
we have to sort our images to generate a sequence which maintain a strong correlation among consecutive instance. 
For this purpose, we quickly formalized a random walk in~\cite{rws,DGM04} and create two sequences, to train and test the model respectively.
If $x_s$ is the selected point at the step $s$ of the sequence,
we assign to each instance $x_q$ the probability defined by the gaussian distance (with variance $\sigma >0$):
$$
p(s,q) = e^{-\frac{ || x_s - x_q ||^2}{2 \sigma^2}}.
$$
However, to increase the probability to explore the whole dataset,
at each step we could randomly pick up a point depending on $p(s,q)$ or move to an arbitrarily point (with probability $0.01$).
To explore more points, we also avoid to pick up one of the last points chosen.
We want to study both the 3-class problem and the 2-class problem 0vs1. 
Then we generate the sequences {\em s012} and {\em s01} described in~\RefTab{seq}.
\footnote{The video of these sequences is available at \url{https://drive.google.com/folderview?id=0B7_Mj3qkmLd9c01nMGZKUm5rV0E&usp=sharing}}

\begin{table}[h]
\begin{scriptsize}
\begin{center}
\resizebox{\textwidth}{!}{%
\begin{tabu}{lccccccc}
\tabucline[1.5pt]{-}
\multicolumn{8}{c}{\rule{0pt}{4ex} {\bfseries Sequential Data} \vspace{0.2cm}} \\
\tabucline[1.5pt]{-}
\rule{0pt}{3ex}		&				& \multicolumn{4}{|c|}{Nodes visited} 										& Class 		      &  \\
\rule{0pt}{3ex}		&\multirow{-2}{*}{Total steps}	&\multicolumn{1}{|c}{$0$}	&	$1$			& 	$2$			&\multicolumn{1}{c|}{Total}			& changes 	      & \multirow{-2}{*}{Classes} \vspace{0.1cm}\\
\tabucline[1pt]{-}
\rowcolor{gray!25}
\rule{0pt}{3ex} Short Training	&	$5\cdot 10^4$		&    $4962/5923$	&	$6358/6742$	& 	-			& $11320/12655$     & $346 \, (<1\%)$   & \cellcolor{white!50} \\
\cline{1-7}
\rowcolor{gray!15}
Medium Training	&	$1\cdot 10^5$		&    $5511/5923$	&	$6543/6742$	& 	-			& $12054/12655$     & $645 \, (<1\%)$   & \cellcolor{white!50} \\
\cline{1-7}
\rowcolor{gray!5}
Long Training	&	$5\cdot 10^5$		&    $5870/5923$	&	$6706/6742$	& 	-			& $12576/12655$     & $3069 \, (<1\%)$   & \cellcolor{white!50} \\
\cline{1-7}
\rowcolor{gray!20}
\rule{0pt}{3ex}Short Test 	&	$5\cdot 10^4$		&    $949/980$	 	&	$1122/1135$	& 	-			&   $2071/2115$          & $206 \, (<1\%)$  & \cellcolor{white!50}\\
\rowcolor{gray!10}
Long Test 	&	$1\cdot 10^5$		&    $975/980$	 	&	$1131/1135$	& 	-			&   $2106/2115$          & $631 \, (<1\%)$     &  \multirow{-5}{*}{\cellcolor{white!50} 0,1} \\
\tabucline[1pt]{-}
\rowcolor{gray!20}
\rule{0pt}{3ex} Train 	&	100000			&    $5403/5923$	&	$6641/6742$	& 	$3393/5958$	& $15437/18623$     & $2203 \, (2\%)$   & \cellcolor{white!50}\\
\cline{1-7}
\rowcolor{gray!10}
\rule{0pt}{3ex} Test	 	&	30000		&    $976/980$	 	&	$1133/1135$	& 	$1004/1032$	&   $3113/3147$          & $1028 \, (3\%)$     &  \multirow{-2}{*}{\cellcolor{white!50} {\bfseries \emph{s012}}} \\
\tabucline[1pt]{-}
\end{tabu}%
}
\end{center}
\end{scriptsize}
\caption{Description of \emph{s01} and \emph{s012}.}\label{seq}
\end{table}

We set the experiments in the same way of the previous Section.
As already said, we estimate $\ee$ and $\sigma$ on the first $10\%$ of points provided by the sequence.
Te impulsive response is the one in \RefFig{MNIR} again.
We start our experiments by training the model with only $1$ supervision per class, given on two nodes with the same number of hits in the sequences.
A remark can be done on the performance evaluation.
The sequence contains many possible repetitions of some points, then,
if we calculate the prediction accuracy on the sequence, the results can be affected by some noise.
Indeed, multiple apparitions in the sequence of a good predicted instance push up the Accuracy.
Another issues rise up from the nodes which could be classified in different way in different apparitions.
To avoid this problem we calculate an \emph{Averaged Accuracy}, 
by assign to each points a score, given by the sum of as many 1 as the good prediction on it, normalized by the total number of passes on it.
Then this Averaged Accuracy is obtained by sum the scores of each node of the test set, then divide by their total number.

\begin{figure}[htbp]
\resizebox{\textwidth}{!}{\hspace{-0.5cm}
\begin{tabular}{cc}
\includegraphics[scale=0.4]{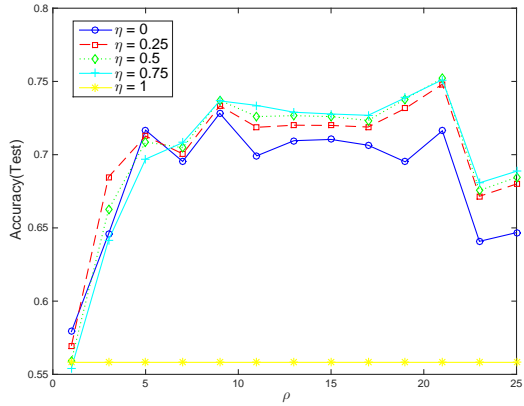} & \includegraphics[scale=0.4]{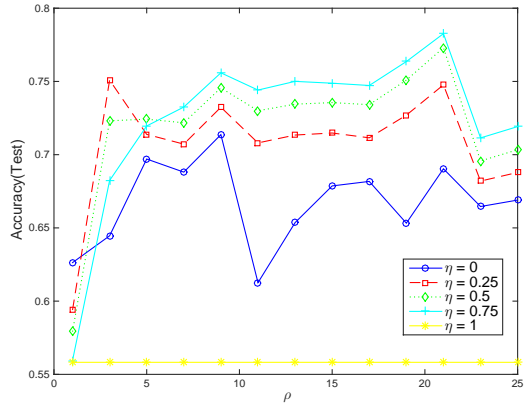}  \\
(a) $\q = 2\, , \; \l = 0.01 $ & (b) $\q = 2\, , \; \l = 0.004 $\\
\includegraphics[scale=0.4]{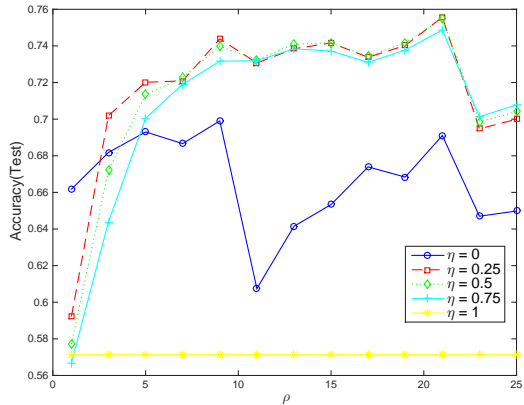} & \includegraphics[scale=0.4]{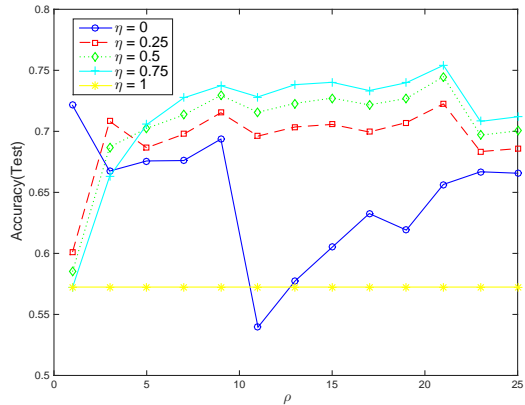}  \\
(c) $\q = 6\, , \; \l = 0.01 $ & (d) $\q = 6\, , \; \l = 0.004 $ \\
\end{tabular}
}
\caption{Accuracy vs. $\rho$ (number of spatial neighbors used) for the 0vs1 MNIST classification task.
Data used Short Training ($1$ supervision per class) and Short Test.
In each plot, different line correspond to different values of $\eta$ (balancing between temporal and spatial contributions).
Different value of $\q$ in each line.
Different value of $\l$ in each column.\label{MN01cc}}
\end{figure}

\begin{figure}[htbp]
\resizebox{\textwidth}{!}{\hspace{-0.5cm}
\begin{tabular}{cc}
\includegraphics[scale=0.4]{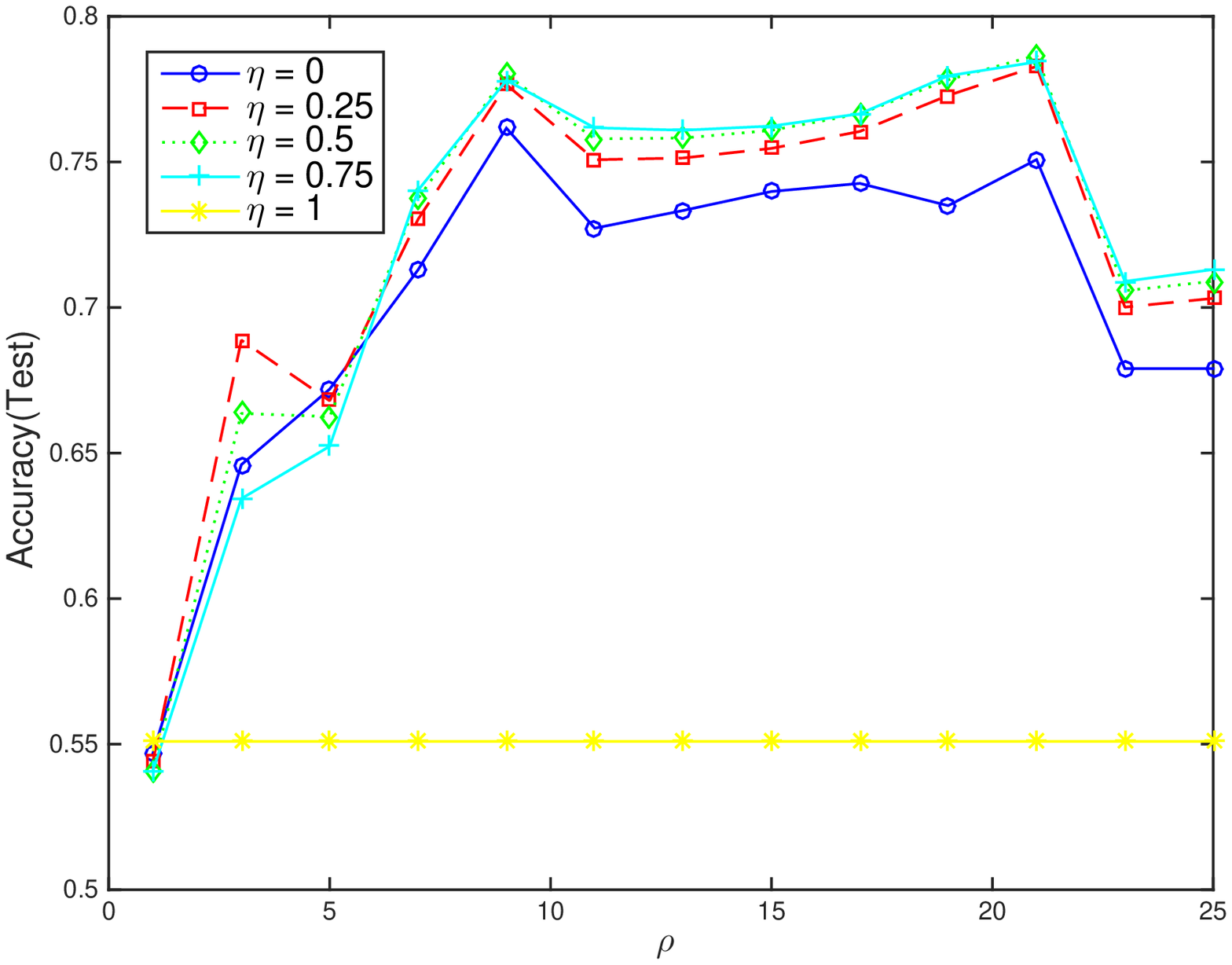} & \includegraphics[scale=0.4]{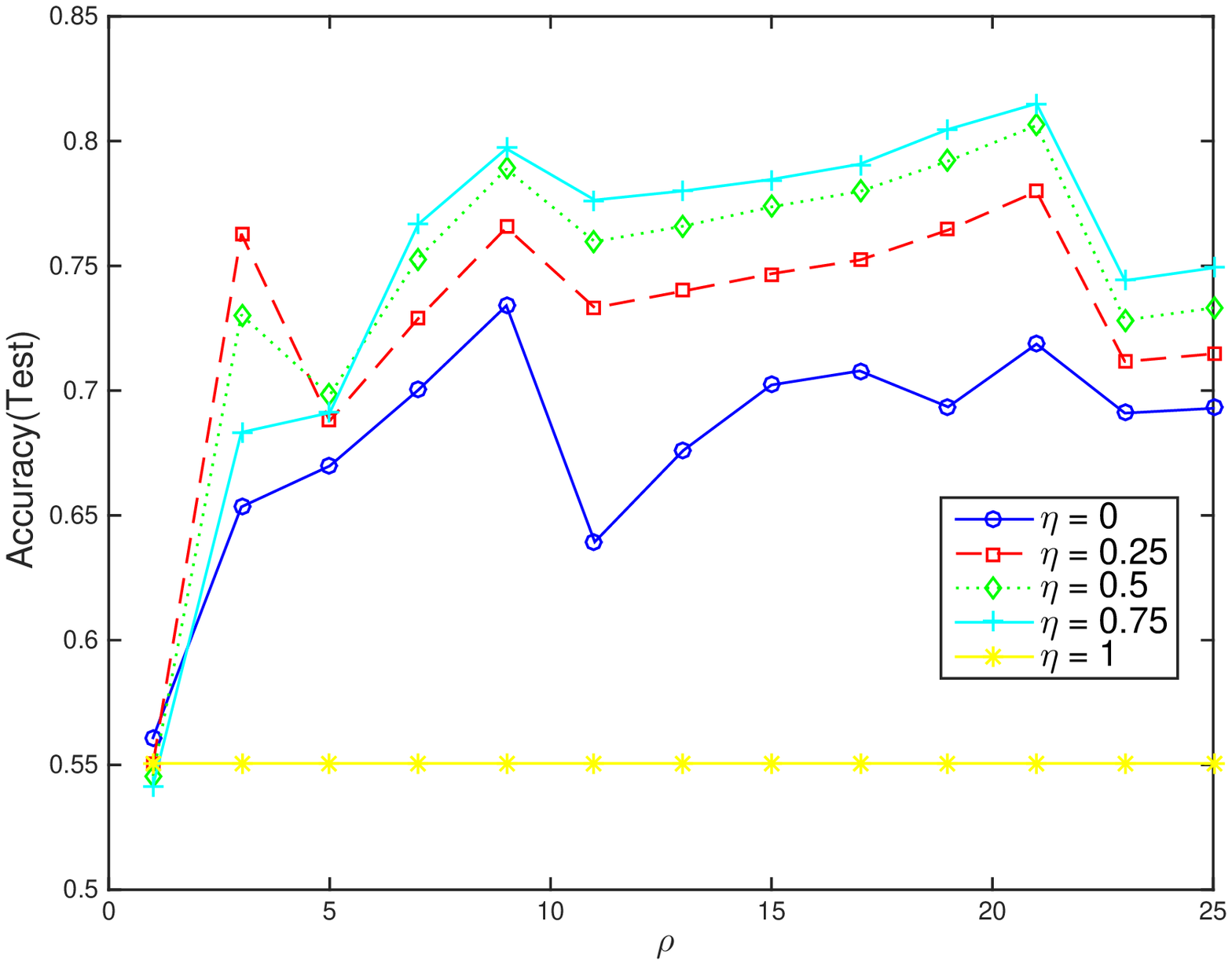}  \\
(a) $\q = 2\, , \; \l = 0.01 $ & (b) $\q = 2\, , \; \l = 0.004 $\\
\includegraphics[scale=0.4]{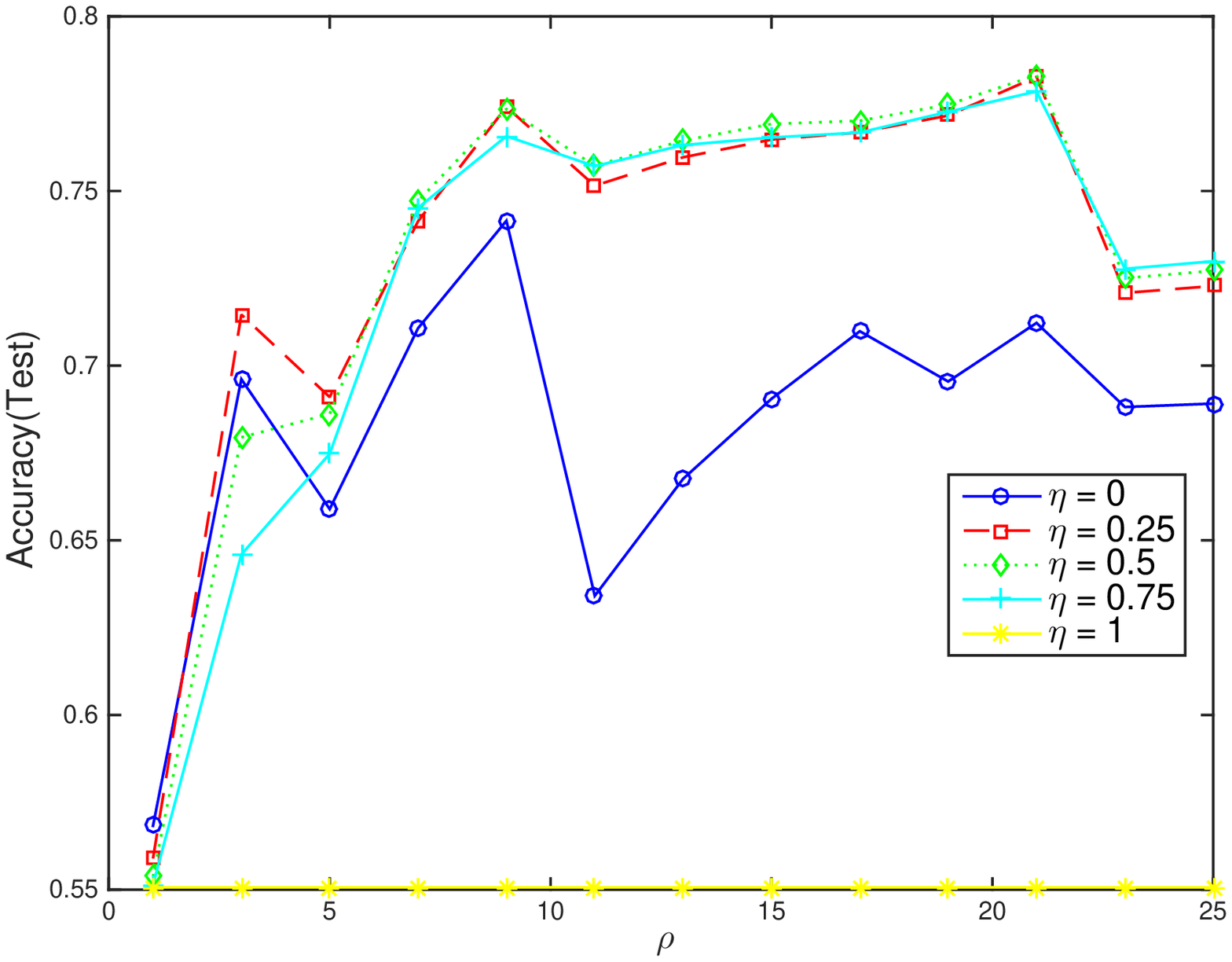} & \includegraphics[scale=0.4]{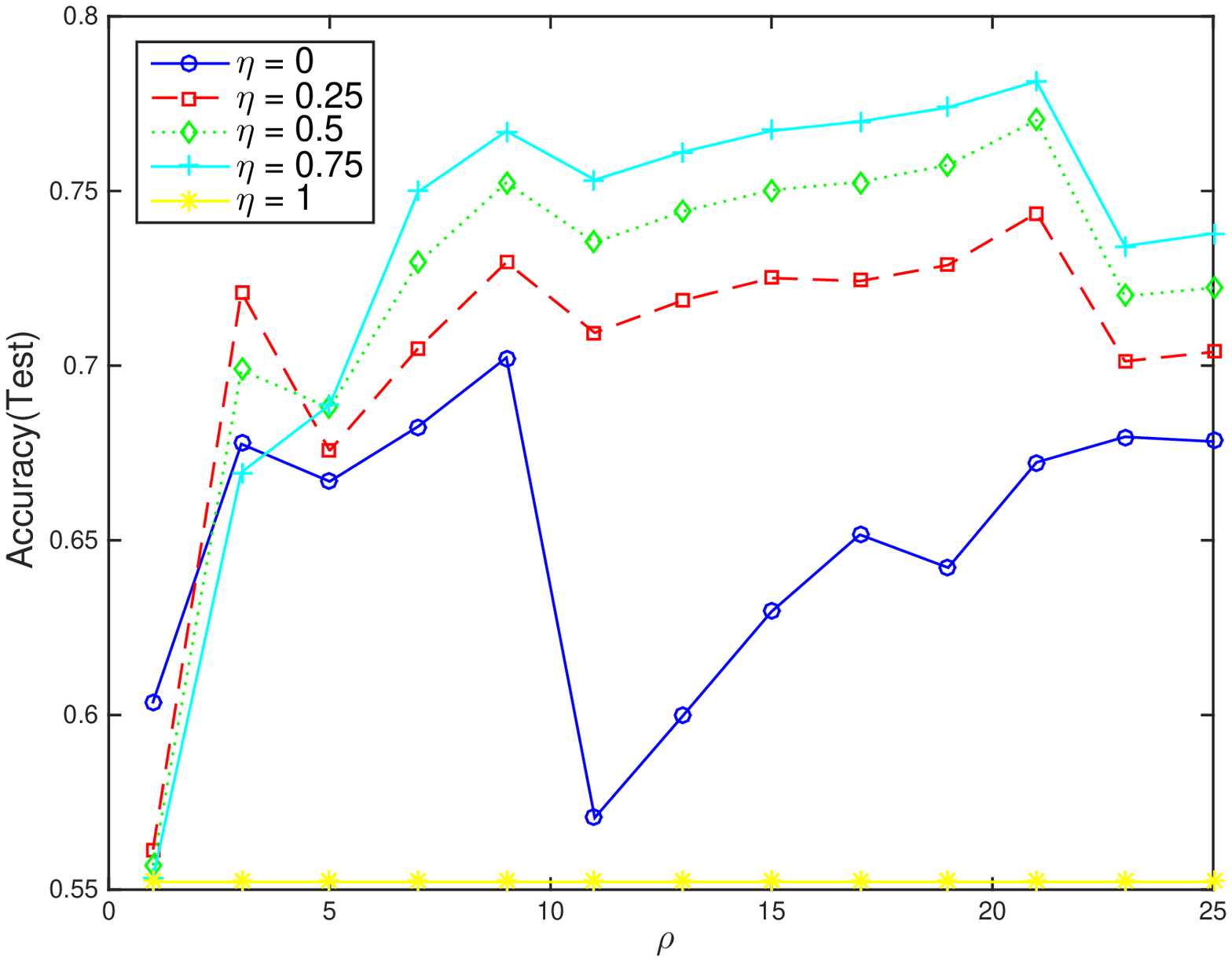}  \\
(c) $\q = 6\, , \; \l = 0.01 $ & (d) $\q = 6\, , \; \l = 0.004 $ \\
\end{tabular}
}
\caption{Accuracy vs. $\rho$ (number of spatial neighbors used) for the 0vs1 MNIST classification task.
Data used Short Training ($1$ supervision per class) and Long Test.
In each plot, different line correspond to different values of $\eta$ (balancing between temporal and spatial contributions).
Different value of $\q$ in each line.
Different value of $\l$ in each column.\label{MN01cl}}
\end{figure}

\begin{figure}[htbp]
\resizebox{\textwidth}{!}{\hspace{-0.5cm}
\begin{tabular}{cc}
\includegraphics[scale=0.4]{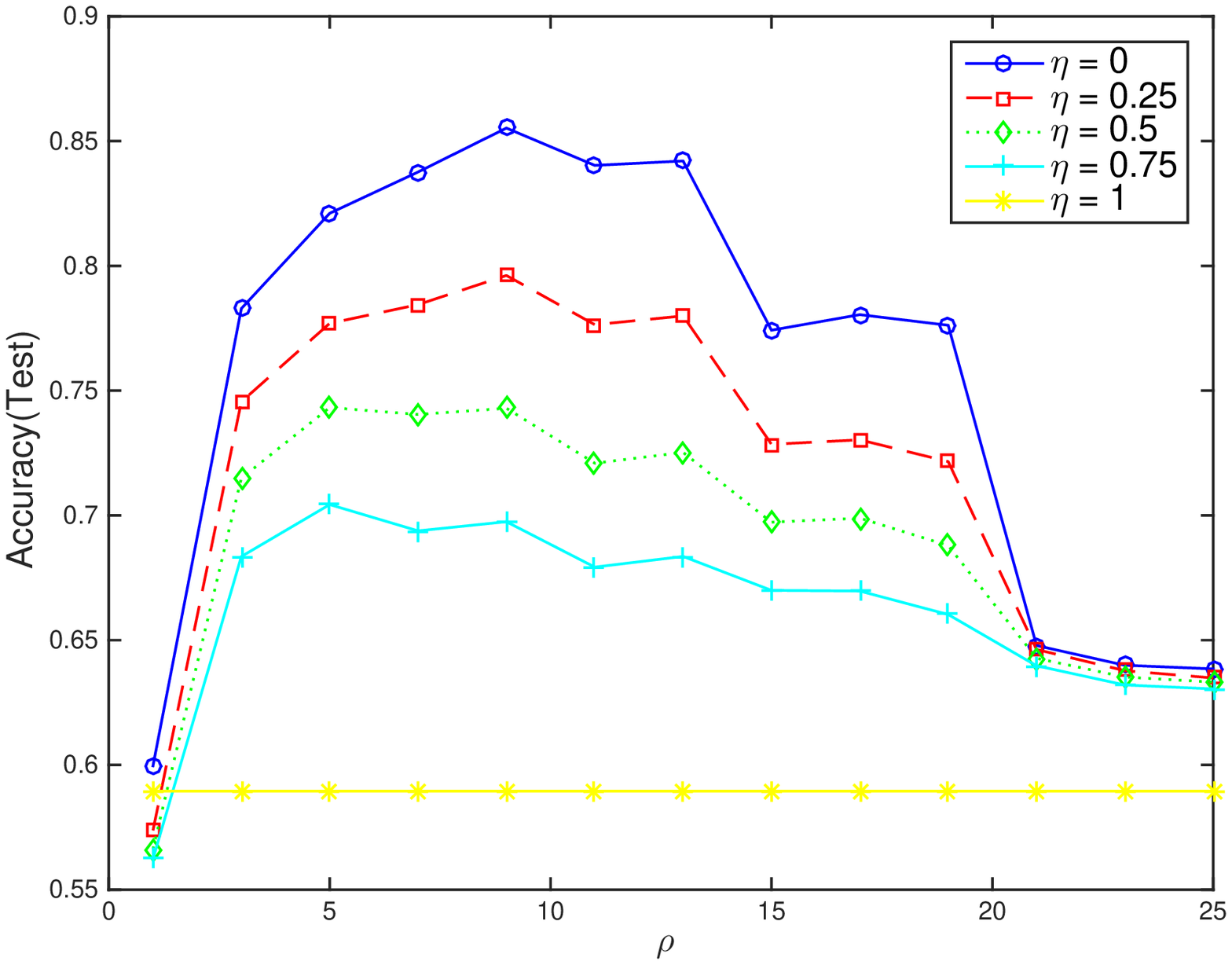} & \includegraphics[scale=0.4]{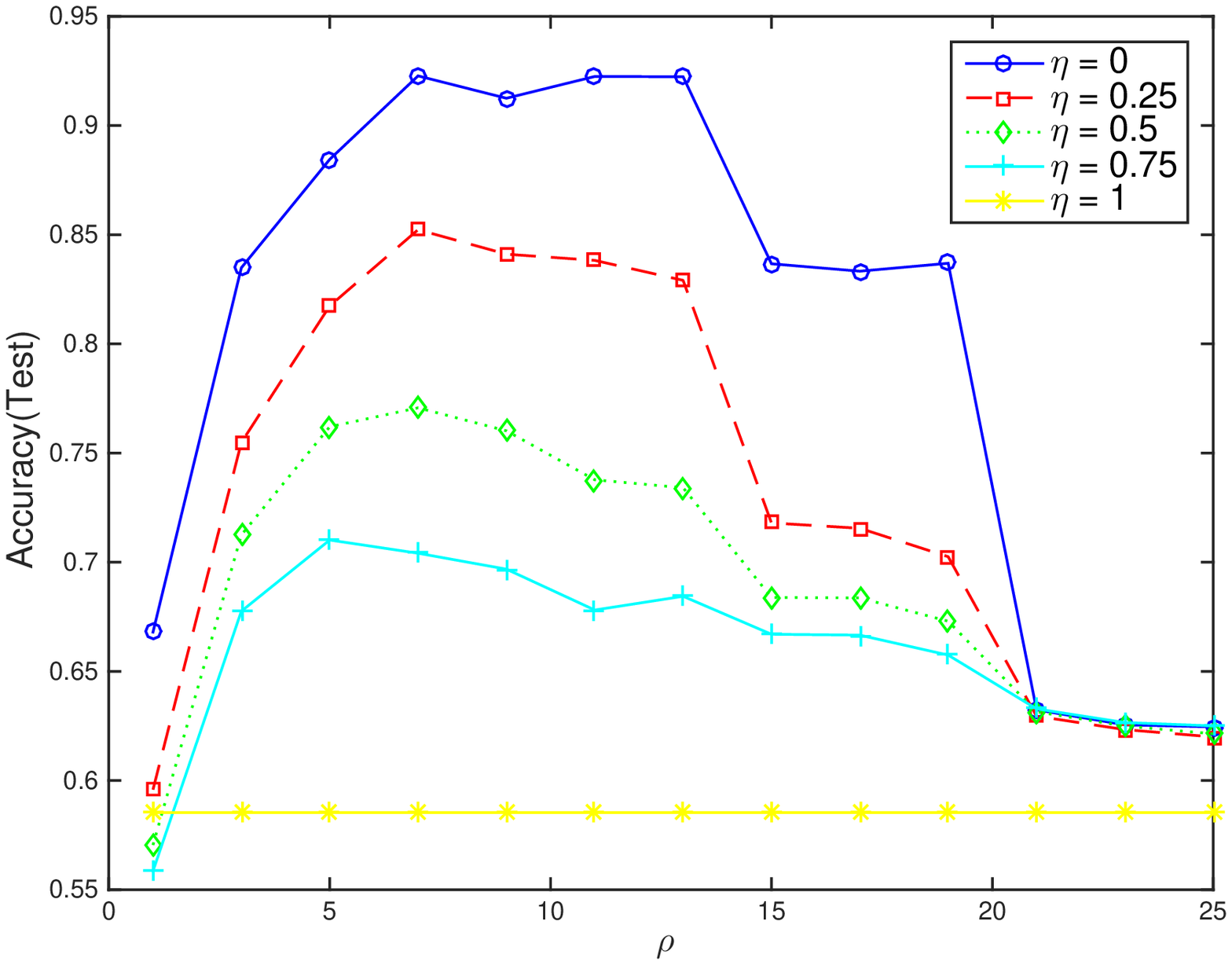}  \\
(a) $\q = 2\, , \; \l = 0.01 $ & (b) $\q = 2\, , \; \l = 0.004 $\\
\includegraphics[scale=0.4]{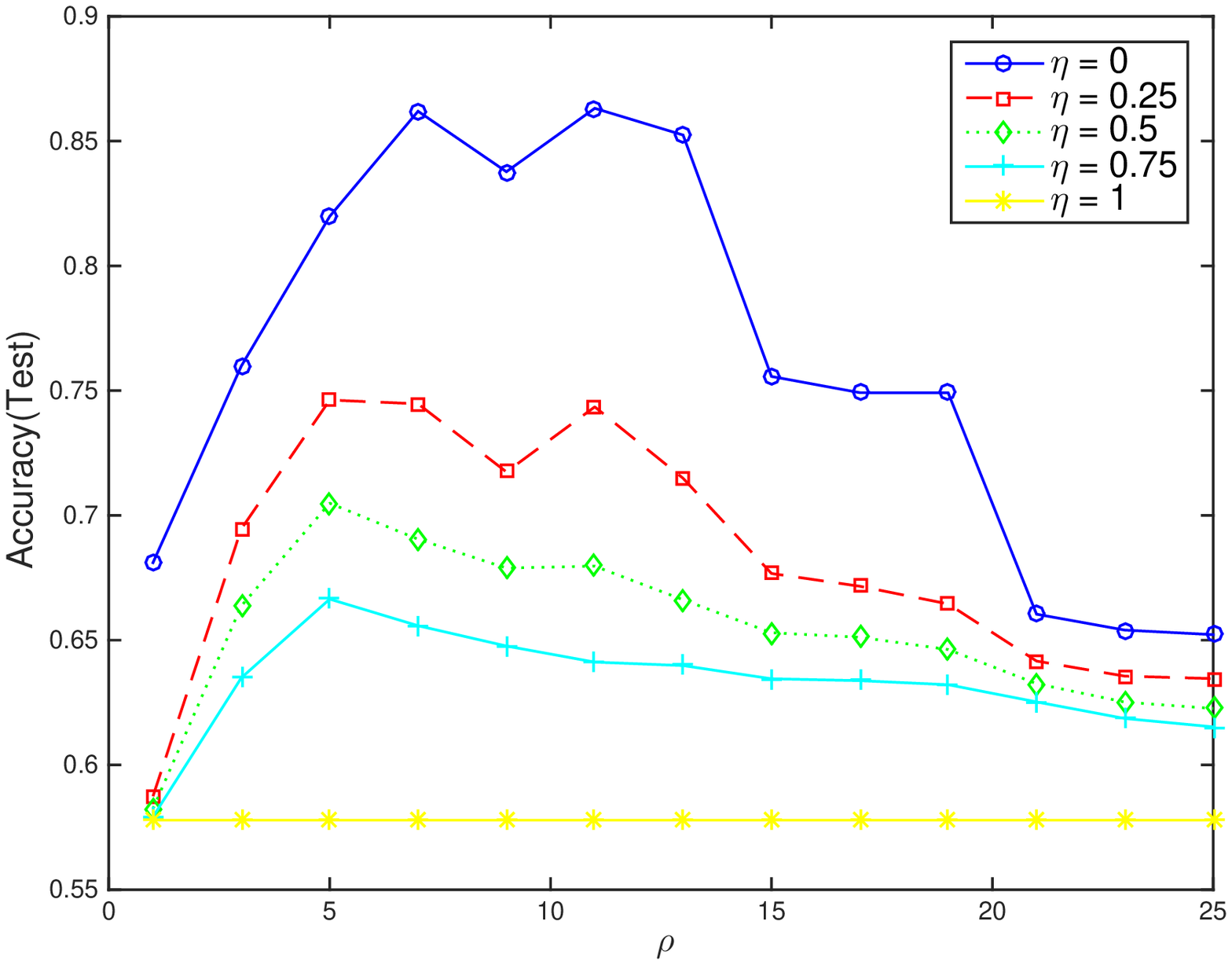} & \includegraphics[scale=0.4]{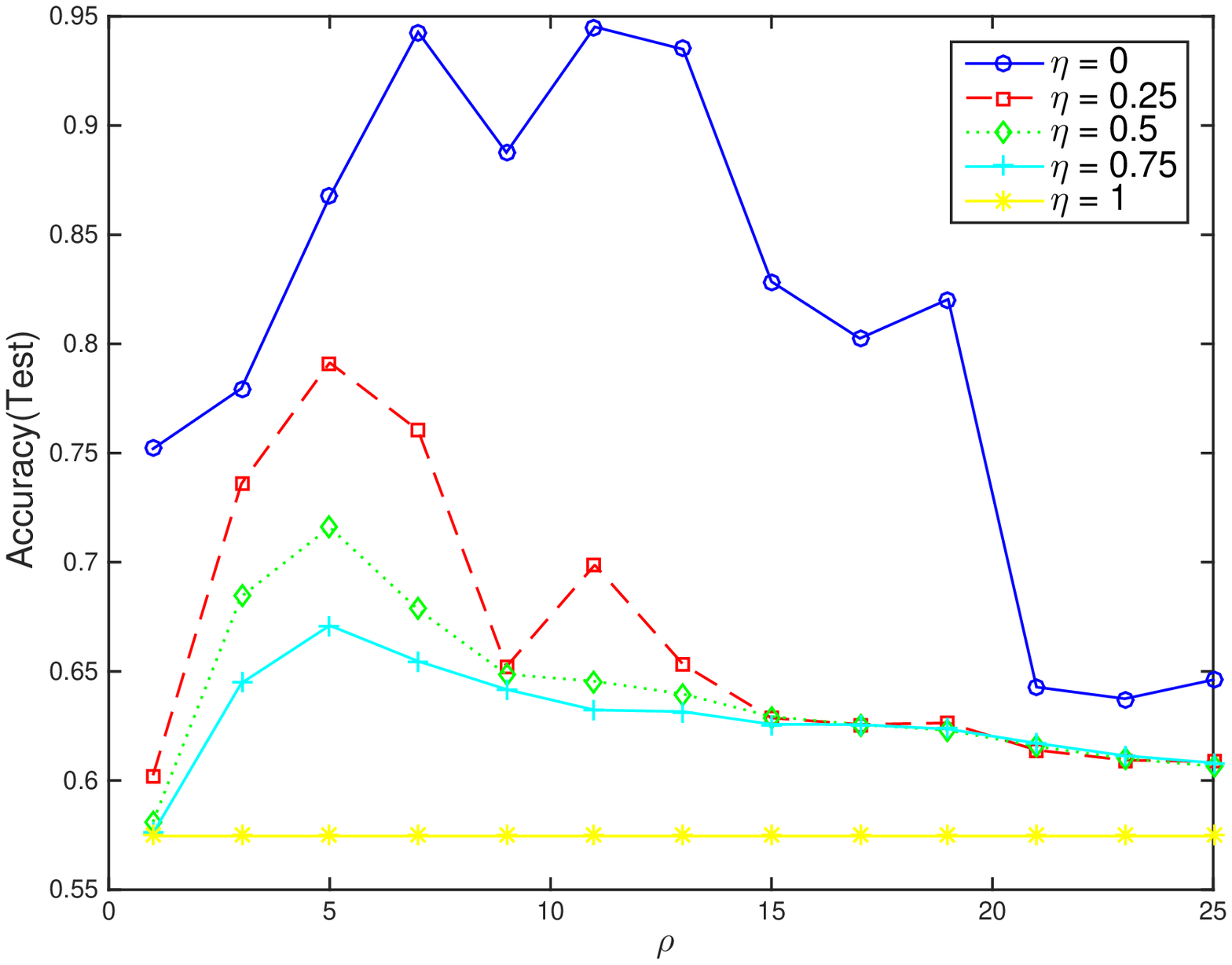}  \\
(c) $\q = 6\, , \; \l = 0.01 $ & (d) $\q = 6\, , \; \l = 0.004 $ \\
\end{tabular}
}
\caption{Accuracy vs. $\rho$ (number of spatial neighbors used) for the 0vs1 MNIST classification task.
Data used Medium Training ($1$ supervision per class) and Short Test.
In each plot, different line correspond to different values of $\eta$ (balancing between temporal and spatial contributions).
Different value of $\q$ in each line.
Different value of $\l$ in each column.\label{MN01mc}}
\end{figure}

\begin{figure}[htbp]
\resizebox{\textwidth}{!}{\hspace{-0.5cm}
\begin{tabular}{cc}
\includegraphics[scale=0.4]{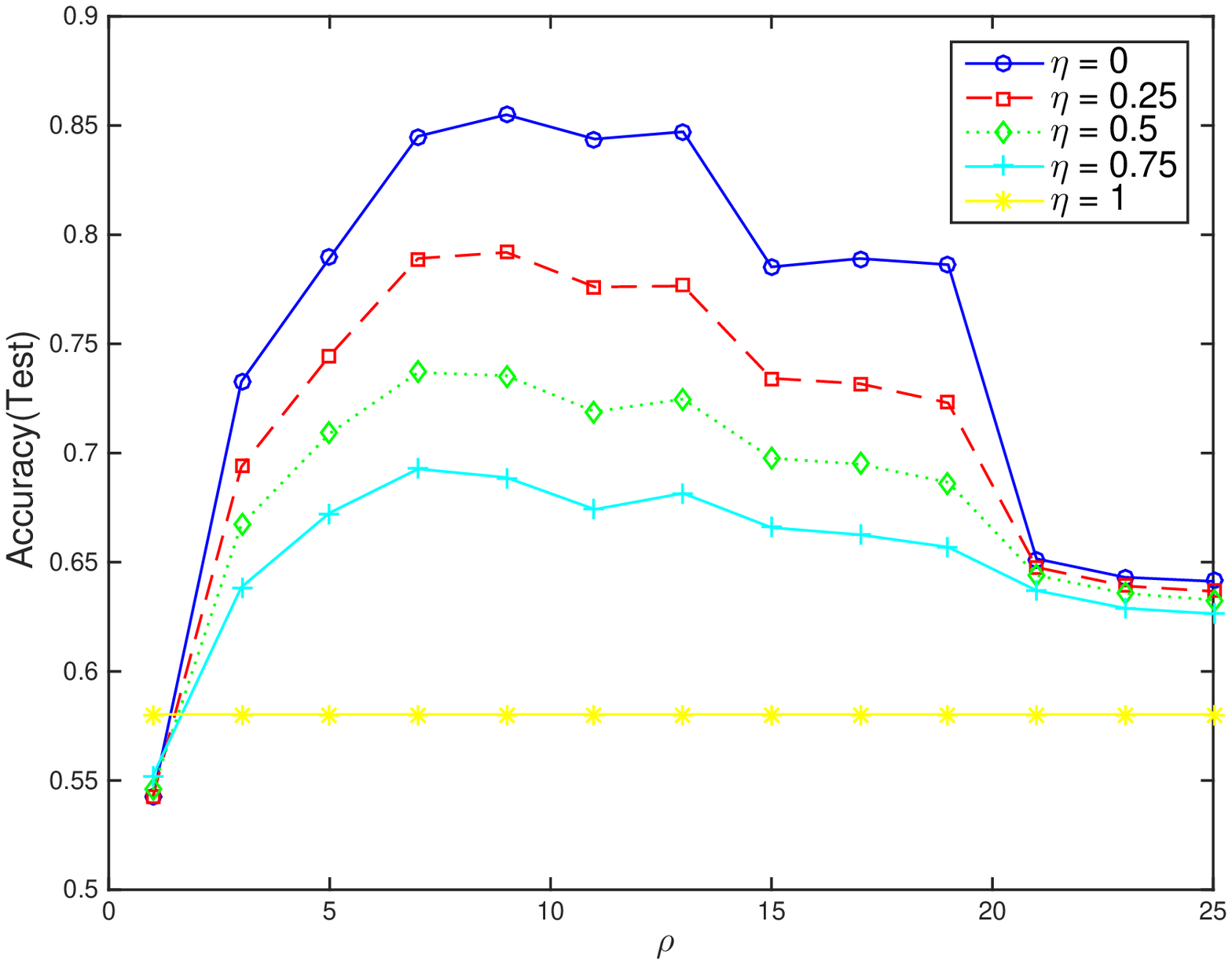} & \includegraphics[scale=0.4]{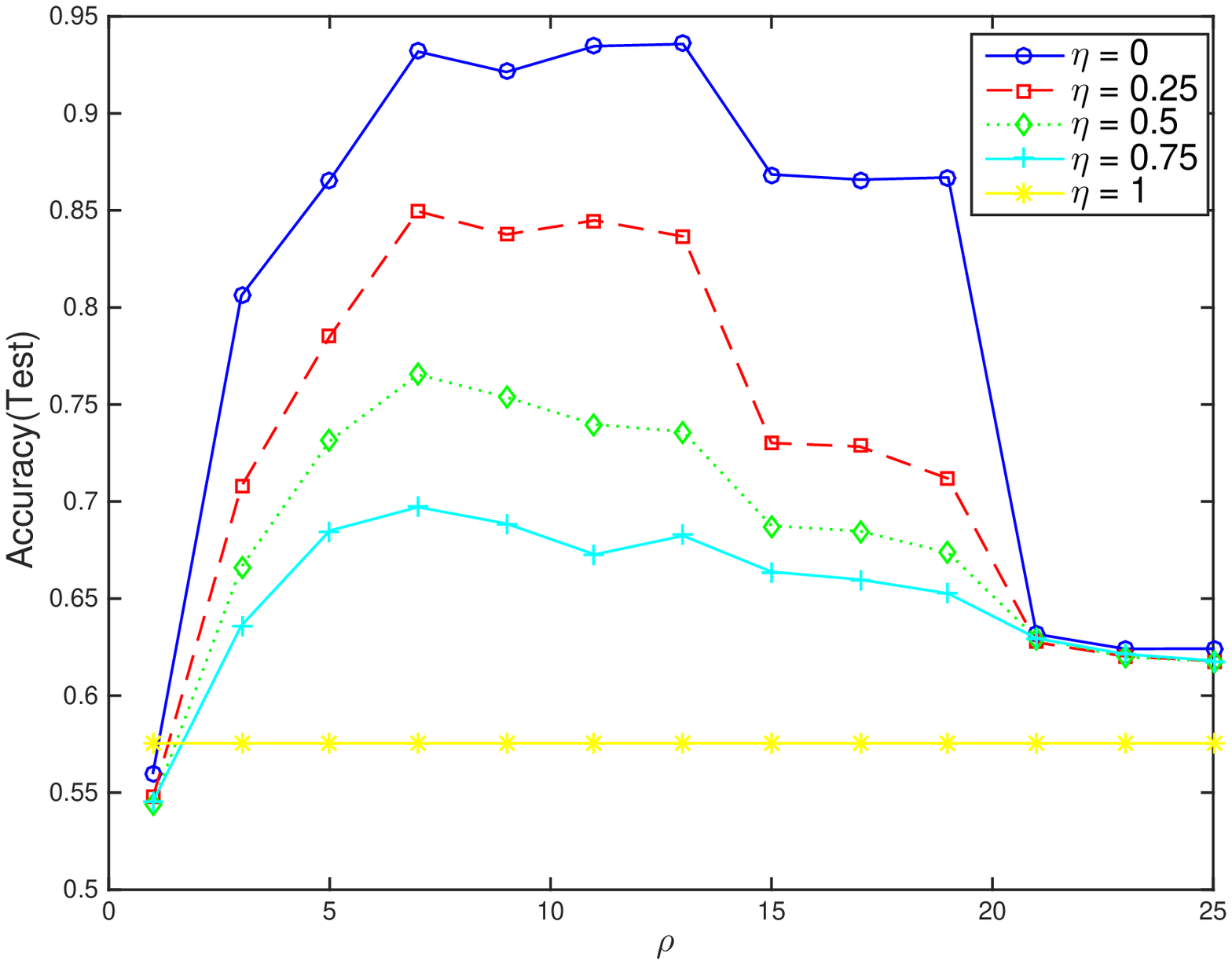}  \\
(a) $\q = 2\, , \; \l = 0.01 $ & (b) $\q = 2\, , \; \l = 0.004 $\\
\includegraphics[scale=0.4]{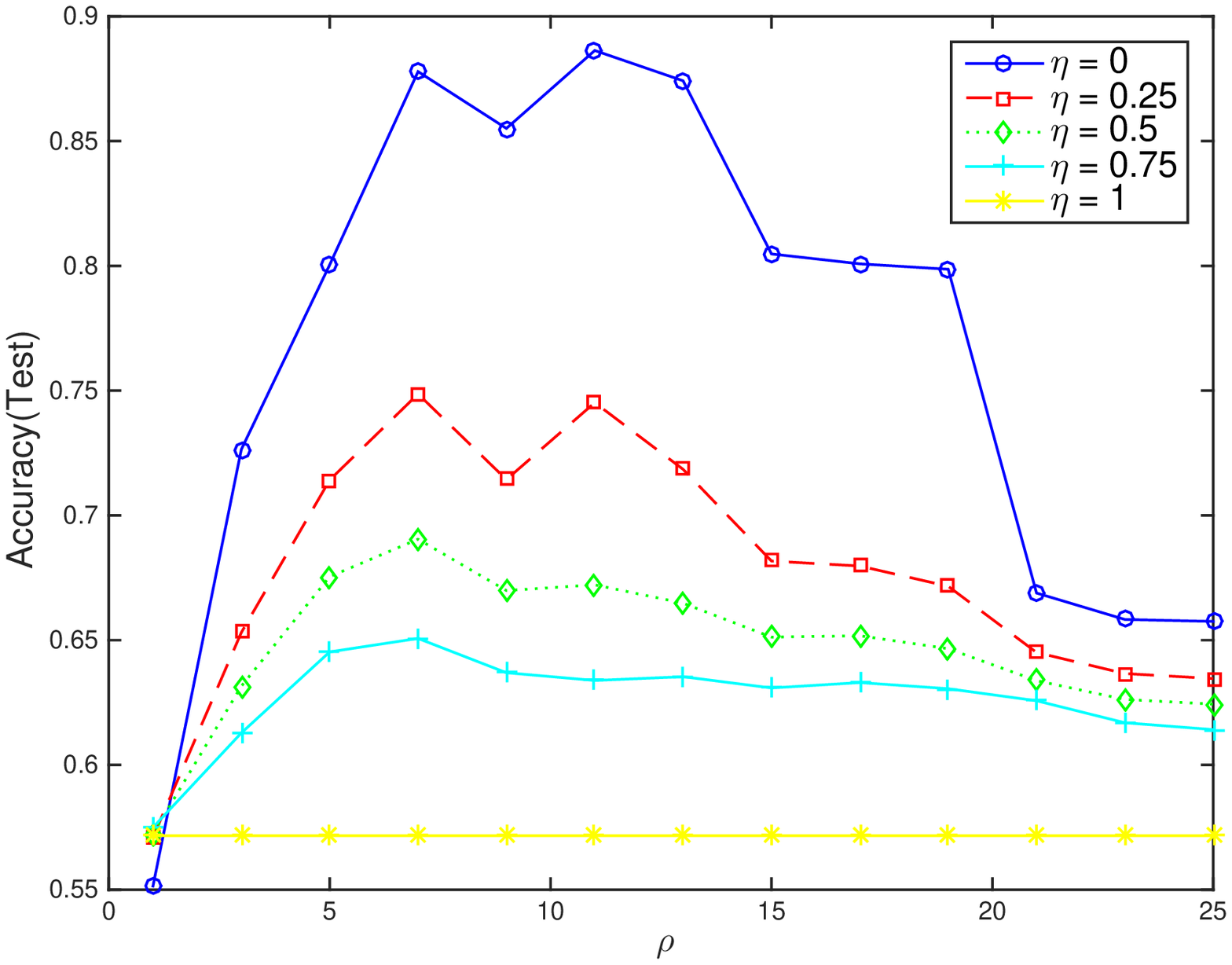} & \includegraphics[scale=0.4]{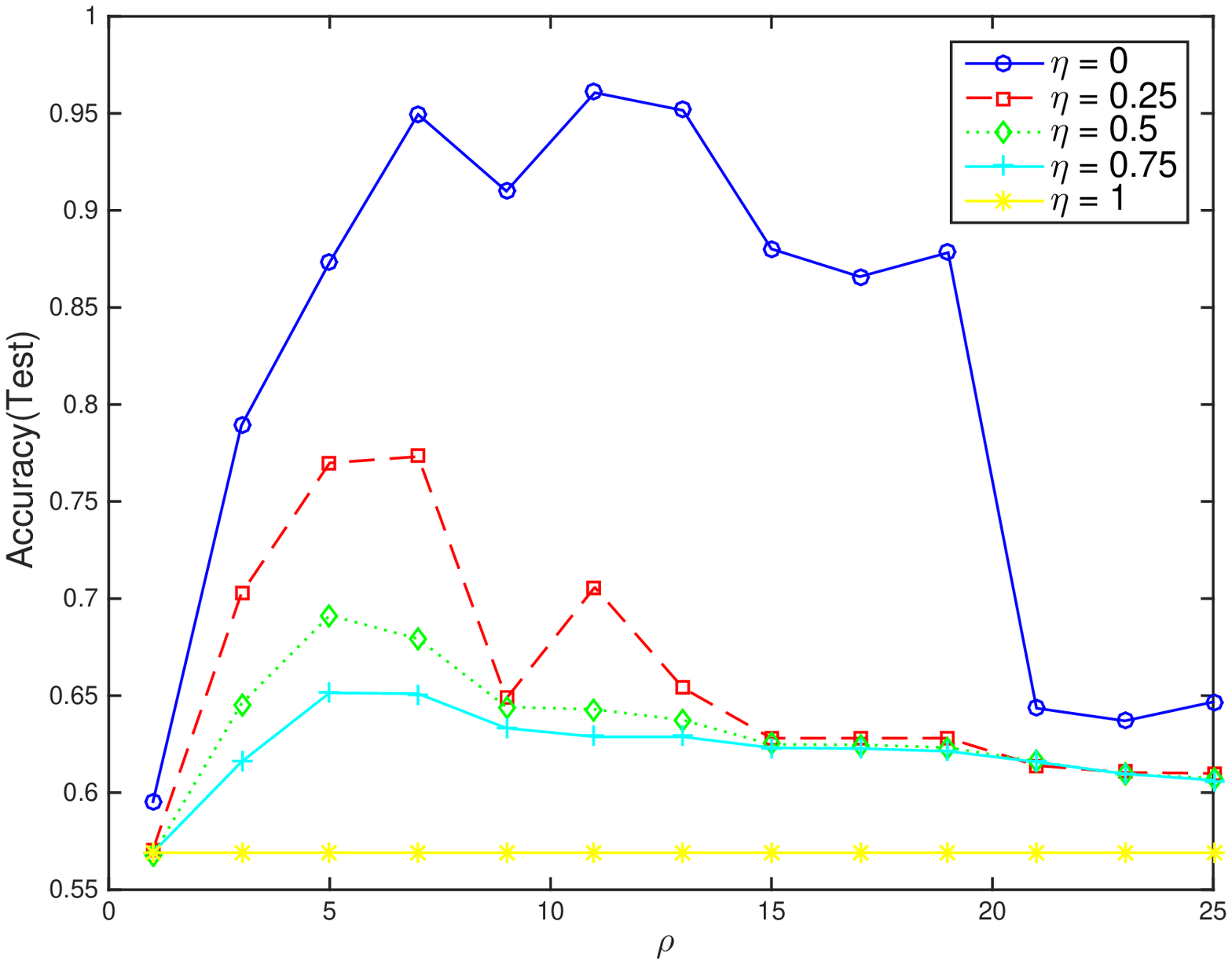}  \\
(c) $\q = 6\, , \; \l = 0.01 $ & (d) $\q = 6\, , \; \l = 0.004 $ \\
\end{tabular}
}
\caption{Accuracy vs. $\rho$ (number of spatial neighbors used) for the 0vs1 MNIST classification task.
Data used Medium Training ($1$ supervision per class) and Long Test.
In each plot, different line correspond to different values of $\eta$ (balancing between temporal and spatial contributions).
Different value of $\q$ in each line.
Different value of $\l$ in each column.\label{MN01ml}}
\end{figure}

\begin{figure}[htbp]
\resizebox{\textwidth}{!}{\hspace{-0.5cm}
\begin{tabular}{cc}
\includegraphics[scale=0.4]{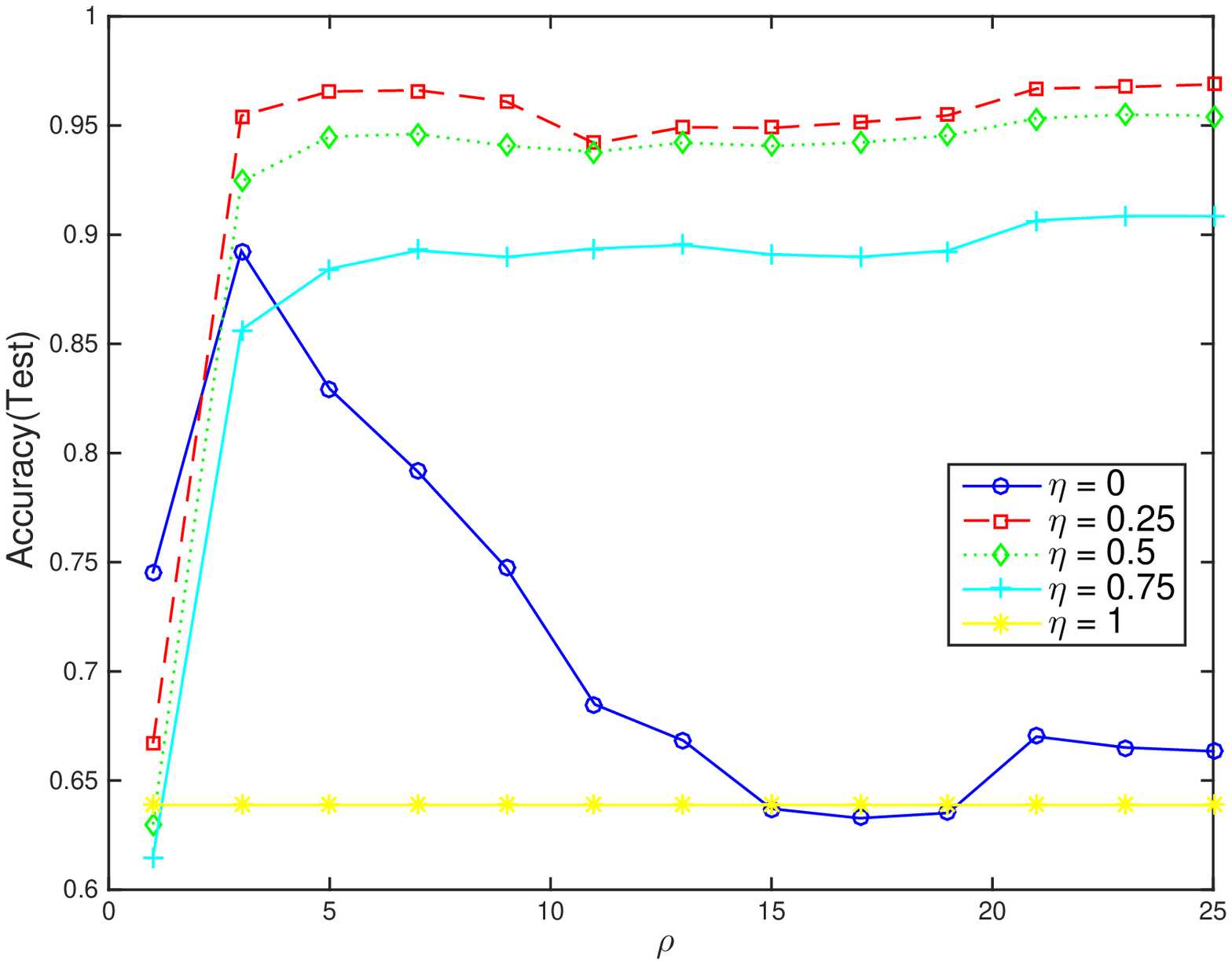} & \includegraphics[scale=0.4]{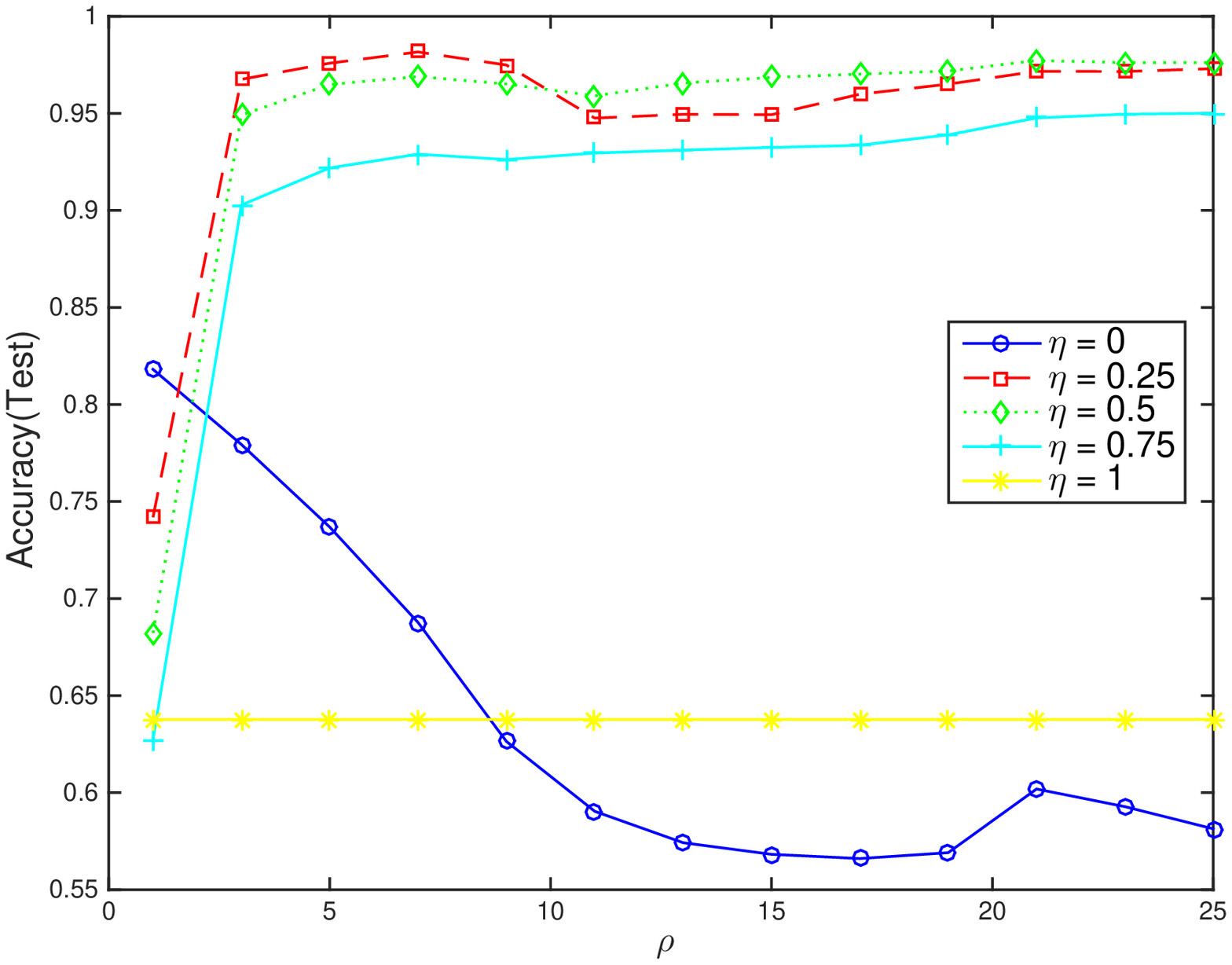}  \\
(a) $\q = 2\, , \; \l = 0.01 $ & (b) $\q = 2\, , \; \l = 0.004 $\\
\includegraphics[scale=0.4]{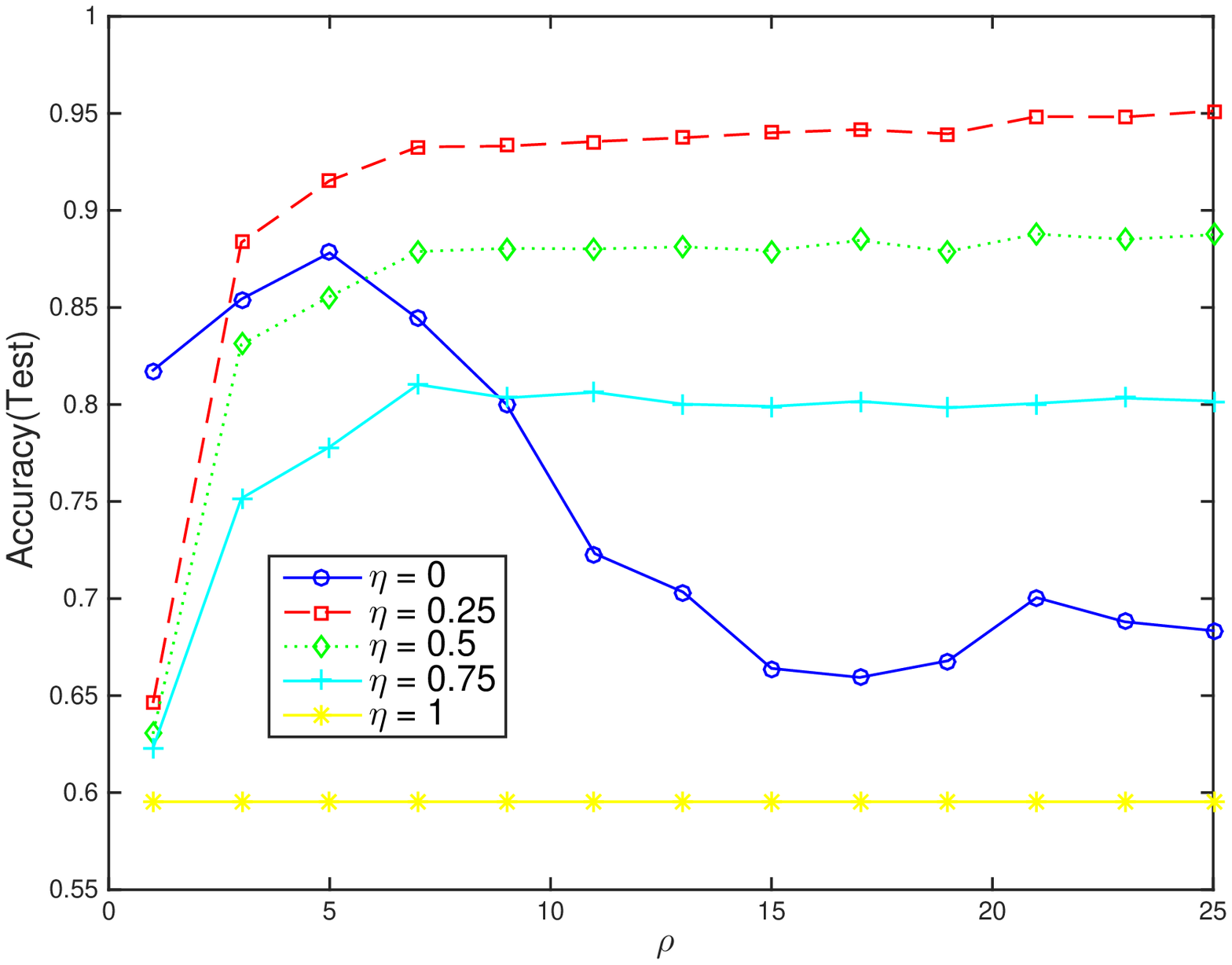} & \includegraphics[scale=0.4]{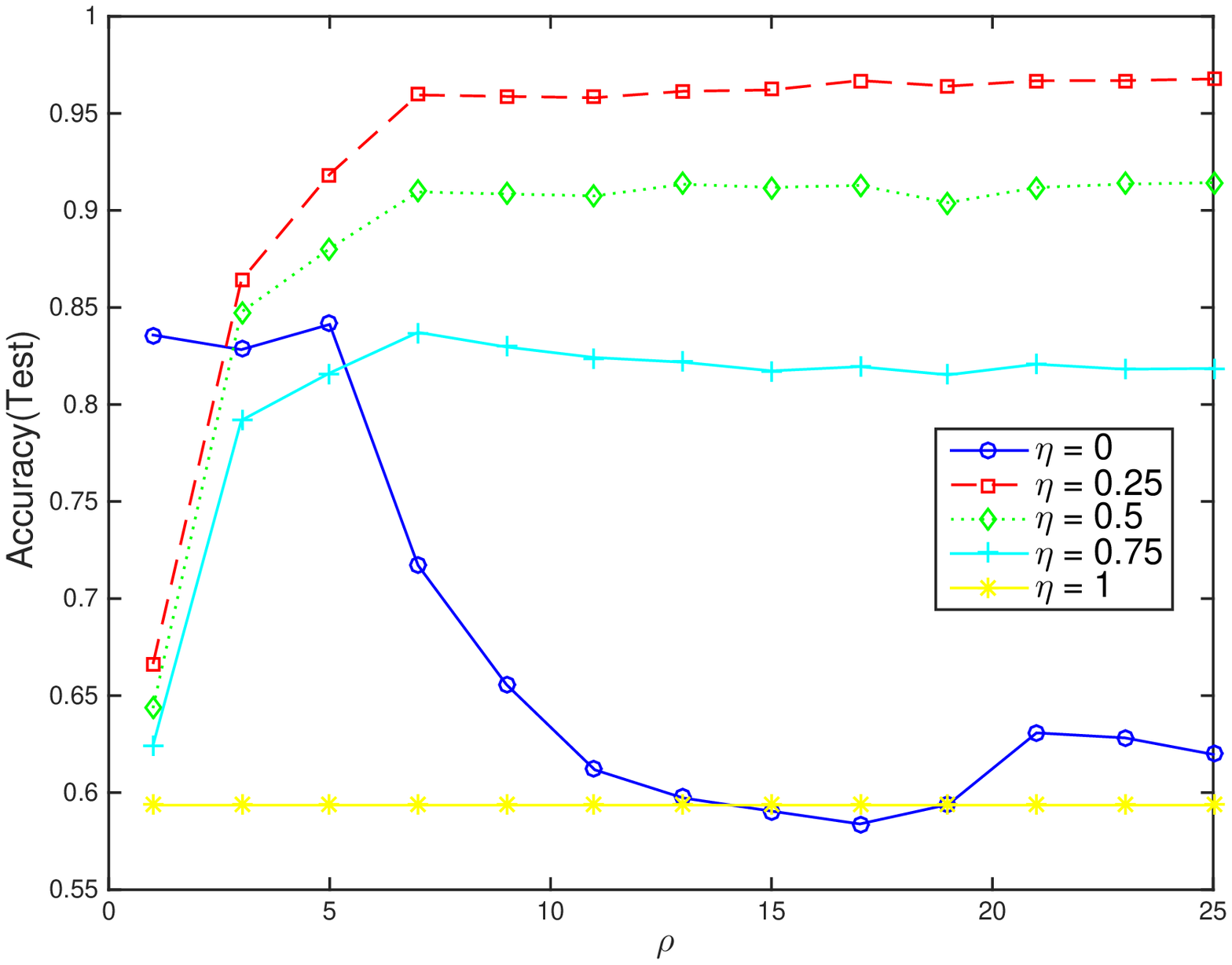}  \\
(c) $\q = 6\, , \; \l = 0.01 $ & (d) $\q = 6\, , \; \l = 0.004 $ \\
\end{tabular}
}
\caption{Accuracy vs. $\rho$ (number of spatial neighbors used) for the 0vs1 MNIST classification task.
Data used Long Training ($1$ supervision per class) and Short Test.
In each plot, different line correspond to different values of $\eta$ (balancing between temporal and spatial contributions).
Different value of $\q$ in each line.
Different value of $\l$ in each column.\label{MN01lc}}
\end{figure}

\begin{figure}[htbp]
\resizebox{\textwidth}{!}{\hspace{-0.5cm}
\begin{tabular}{cc}
\includegraphics[scale=0.4]{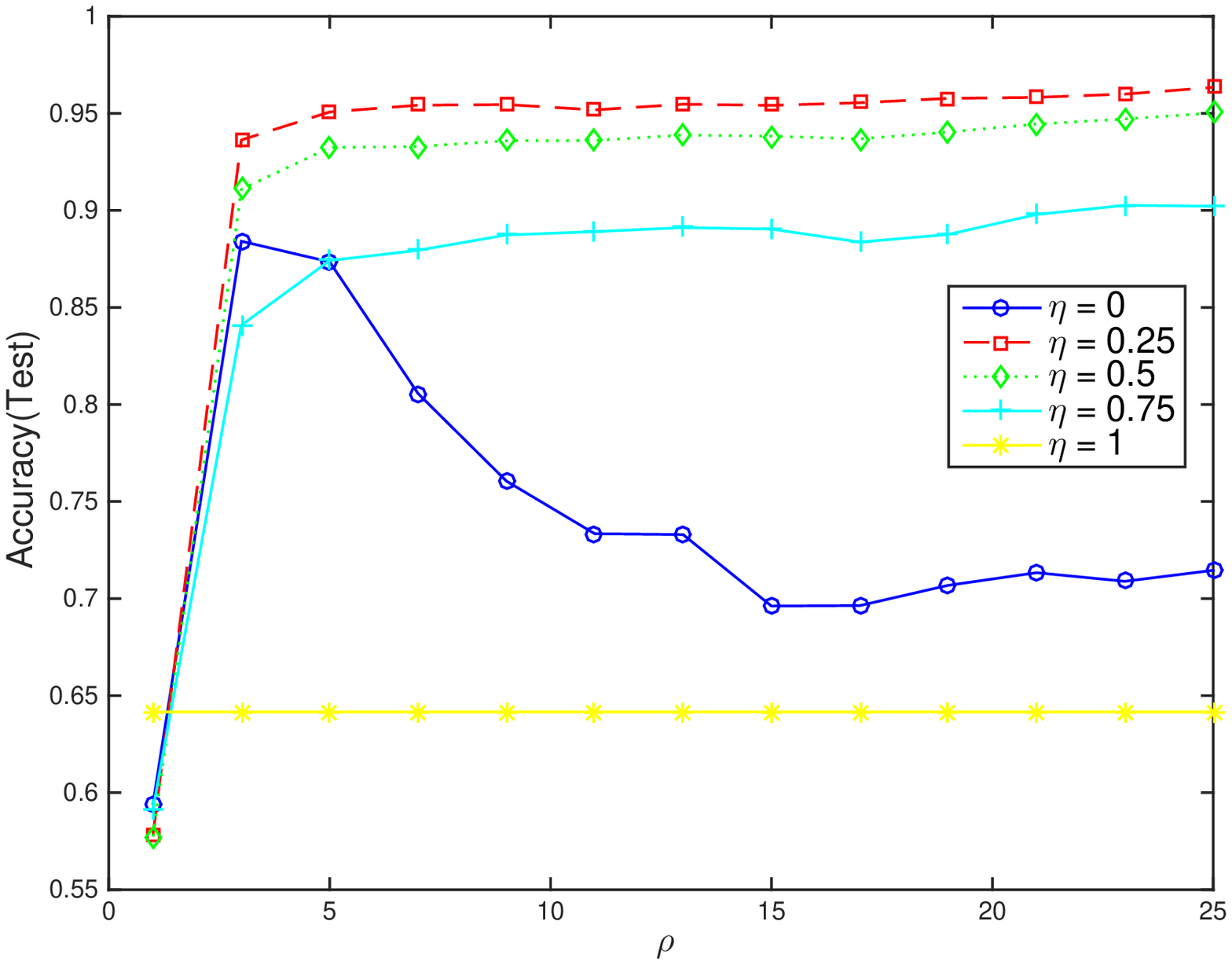} & \includegraphics[scale=0.4]{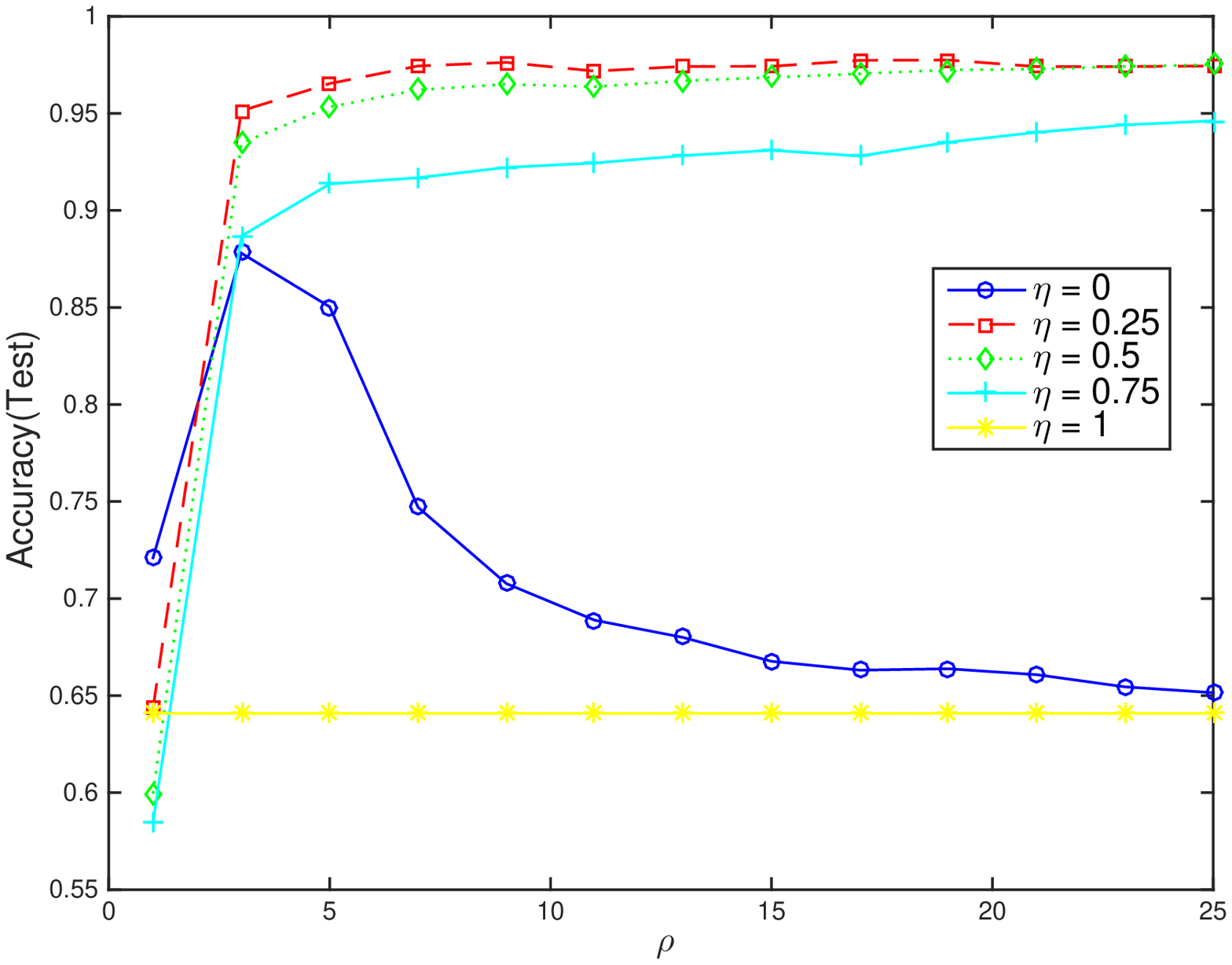}  \\
(a) $\q = 2\, , \; \l = 0.01 $ & (b) $\q = 2\, , \; \l = 0.004 $\\
\includegraphics[scale=0.4]{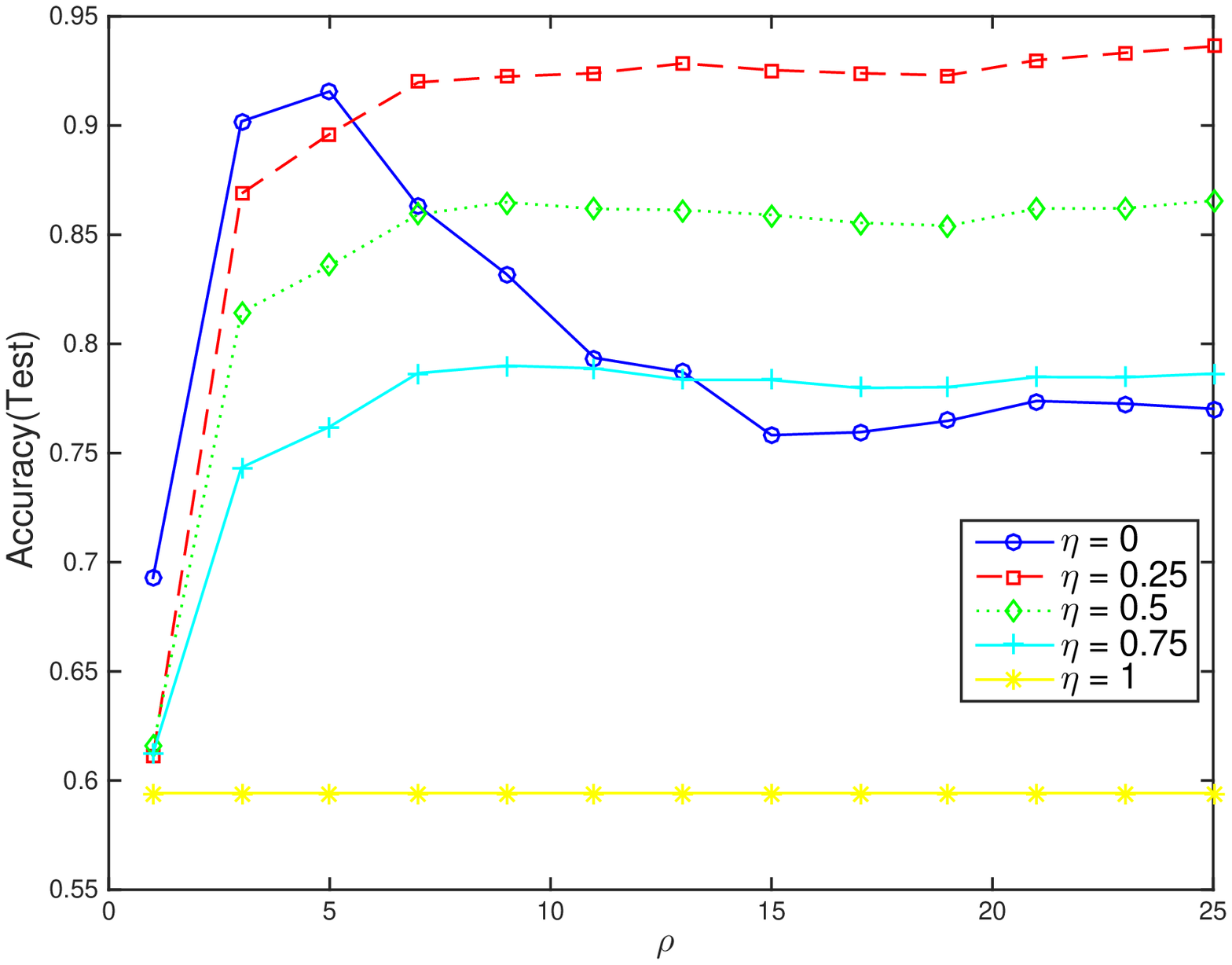} & \includegraphics[scale=0.4]{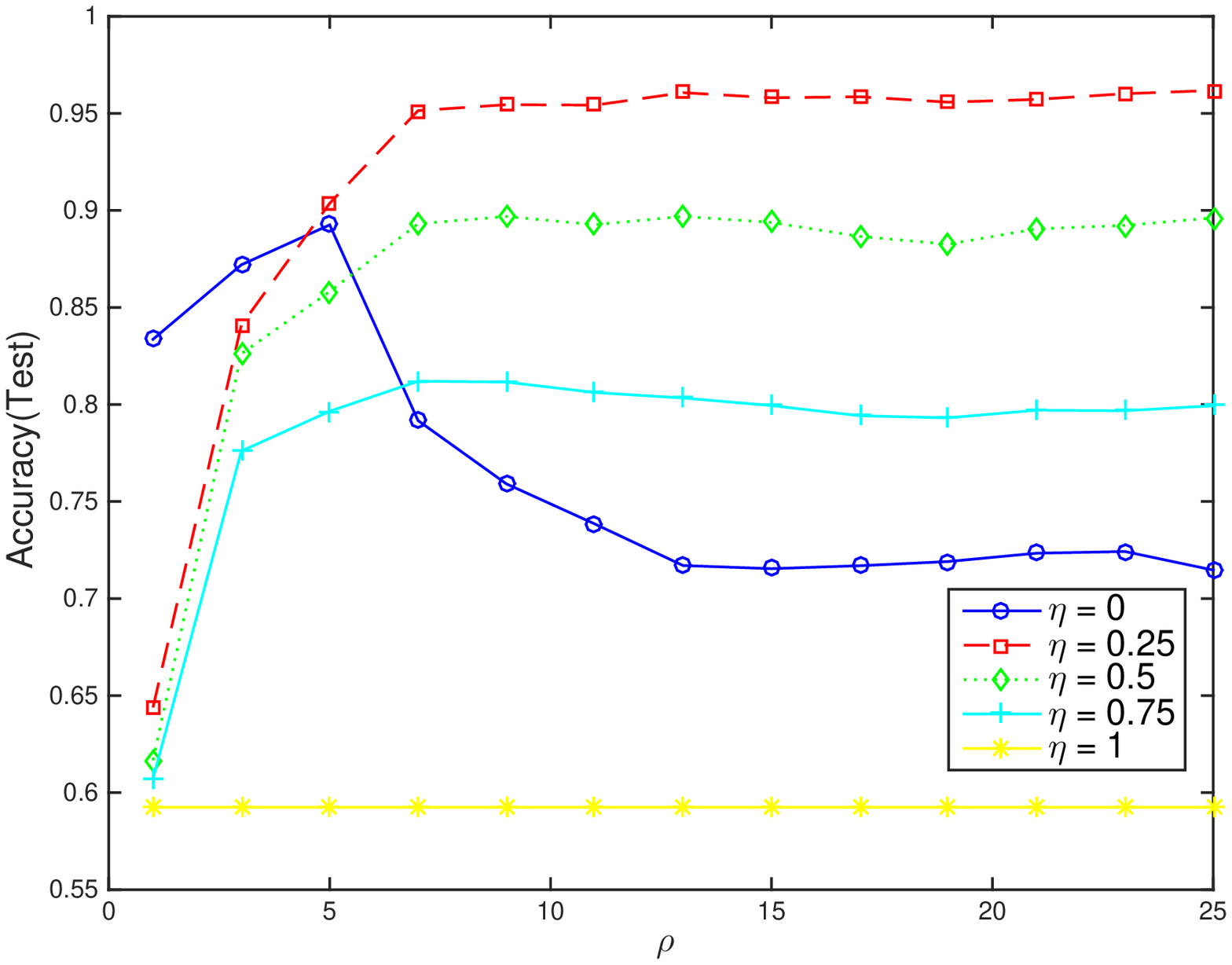}  \\
(c) $\q = 6\, , \; \l = 0.01 $ & (d) $\q = 6\, , \; \l = 0.004 $ \\
\end{tabular}
}
\caption{Accuracy vs. $\rho$ (number of spatial neighbors used) for the 0vs1 MNIST classification task.
Data used Long Training ($1$ supervision per class) and Long Test.
In each plot, different line correspond to different values of $\eta$ (balancing between temporal and spatial contributions).
Different value of $\q$ in each line.
Different value of $\l$ in each column.\label{MN01ll}}
\end{figure}

\begin{figure}[htbp]
\hspace{-1cm}
\includegraphics[scale=0.4]{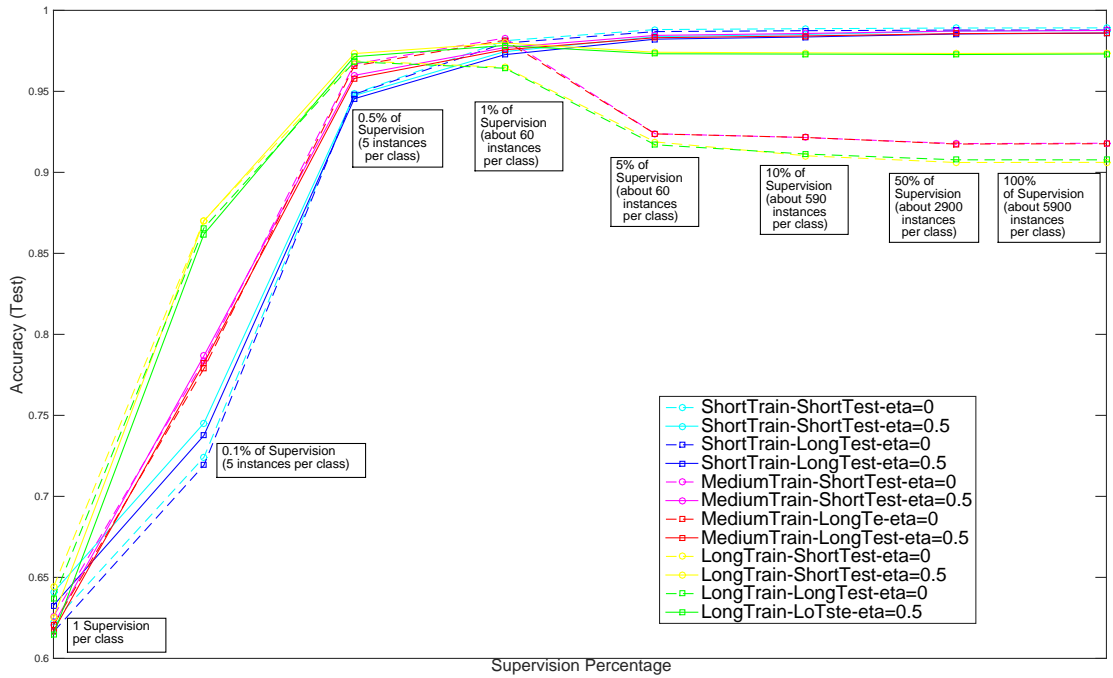}
\caption{Accuracy vs. Supervision Percentage for the 0vs1 MNIST classification task.
Number of supervisions in $\left\{1,\,5 (0.1\%),\, 30 (0.5\%),\, 60(1\%),\, 300(5\%),\, 600(10\%),\, 2900 (50\%),\, 5900 (100\%)\right\}$.
In each plot, different line correspond to different values of $\eta$ (balancing between temporal and spatial contributions).
Different value of $\q$ in each line.
Different value of $\l$ in each column.\label{ShLoavg01}}
\end{figure}

\begin{figure}[htbp]
\resizebox{\textwidth}{!}{\hspace{-0.5cm}
\begin{tabular}{cc}
\includegraphics[scale=0.4]{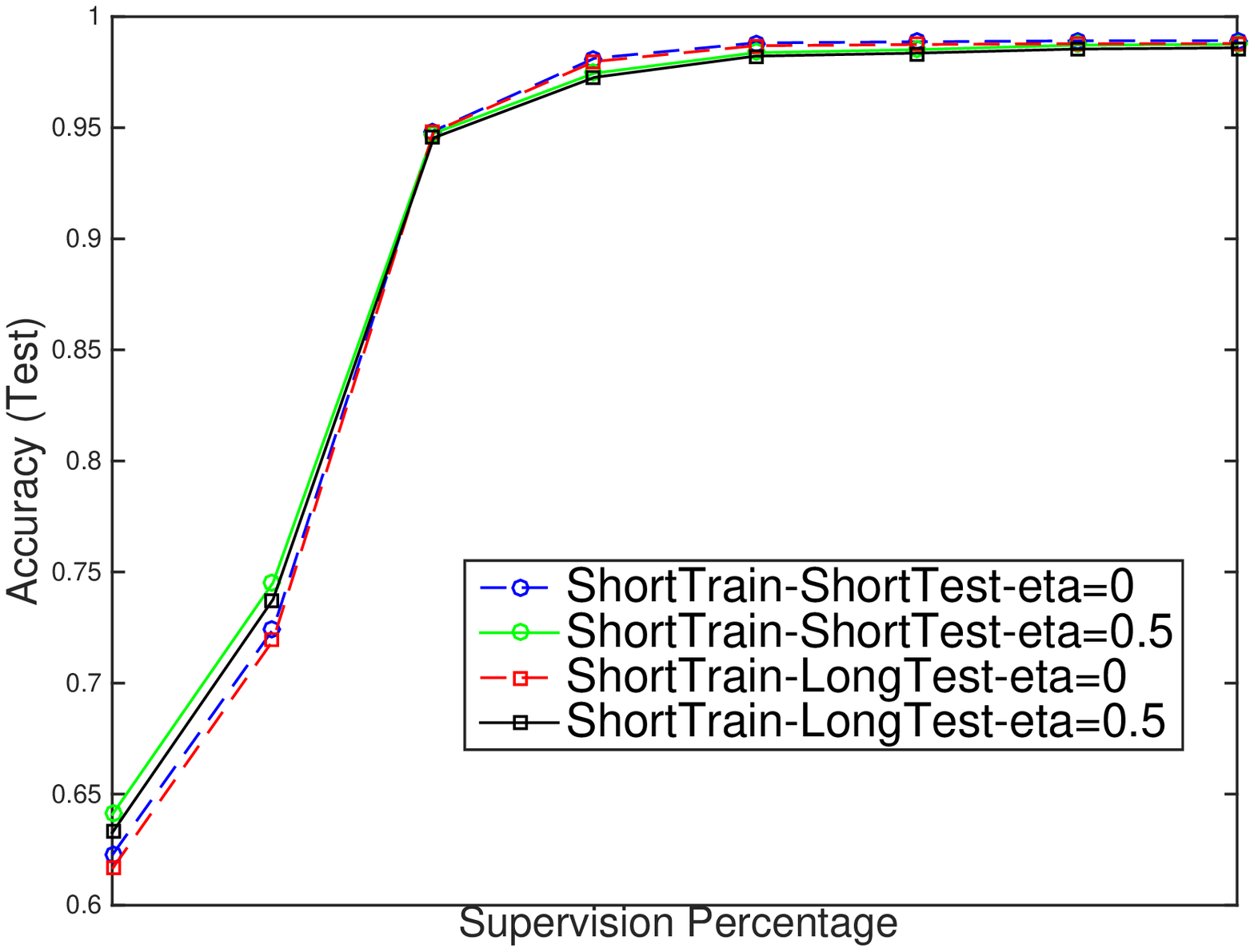} & \includegraphics[scale=0.4]{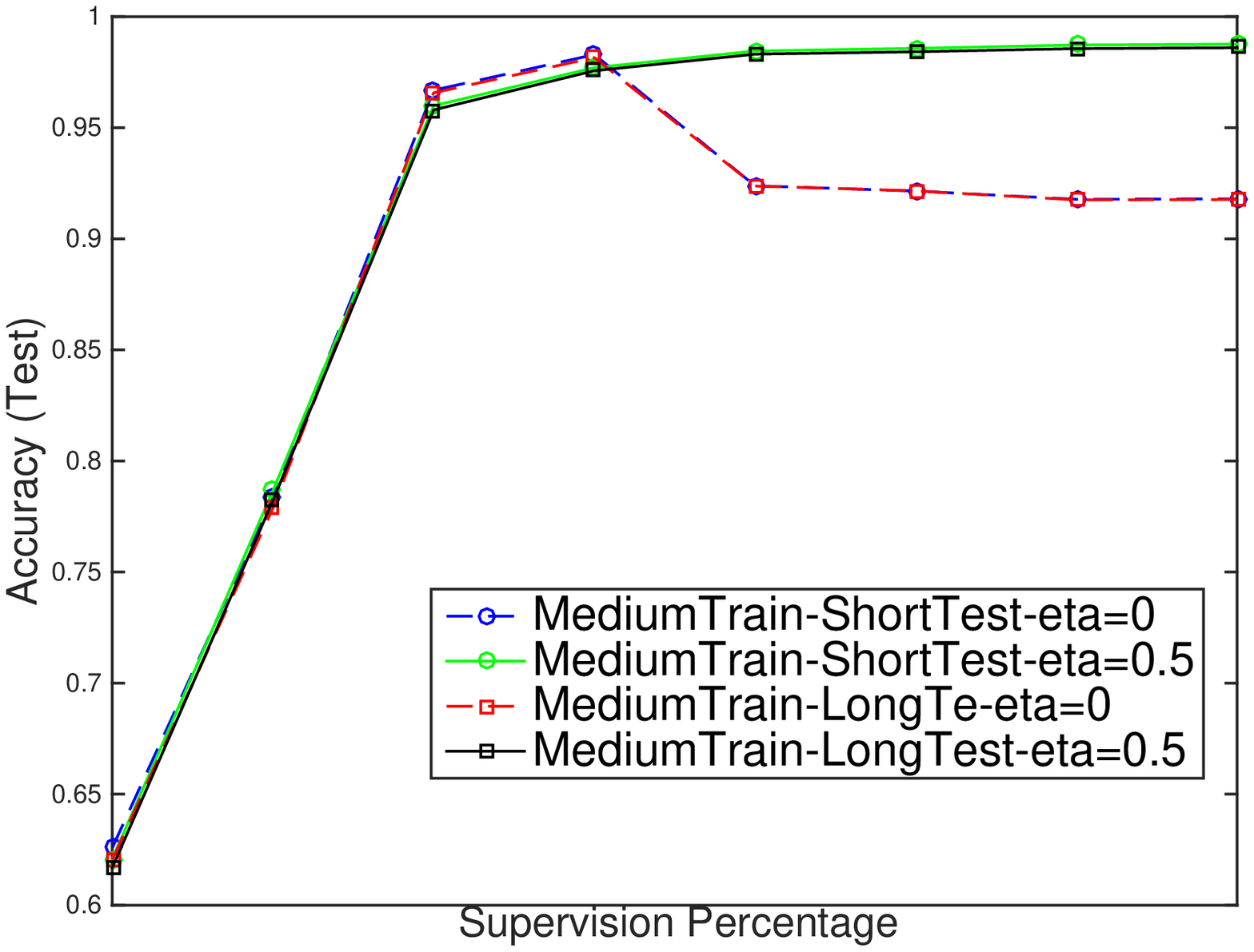}  \\
(a) Short Training & (b) Medium Training\\
\includegraphics[scale=0.4]{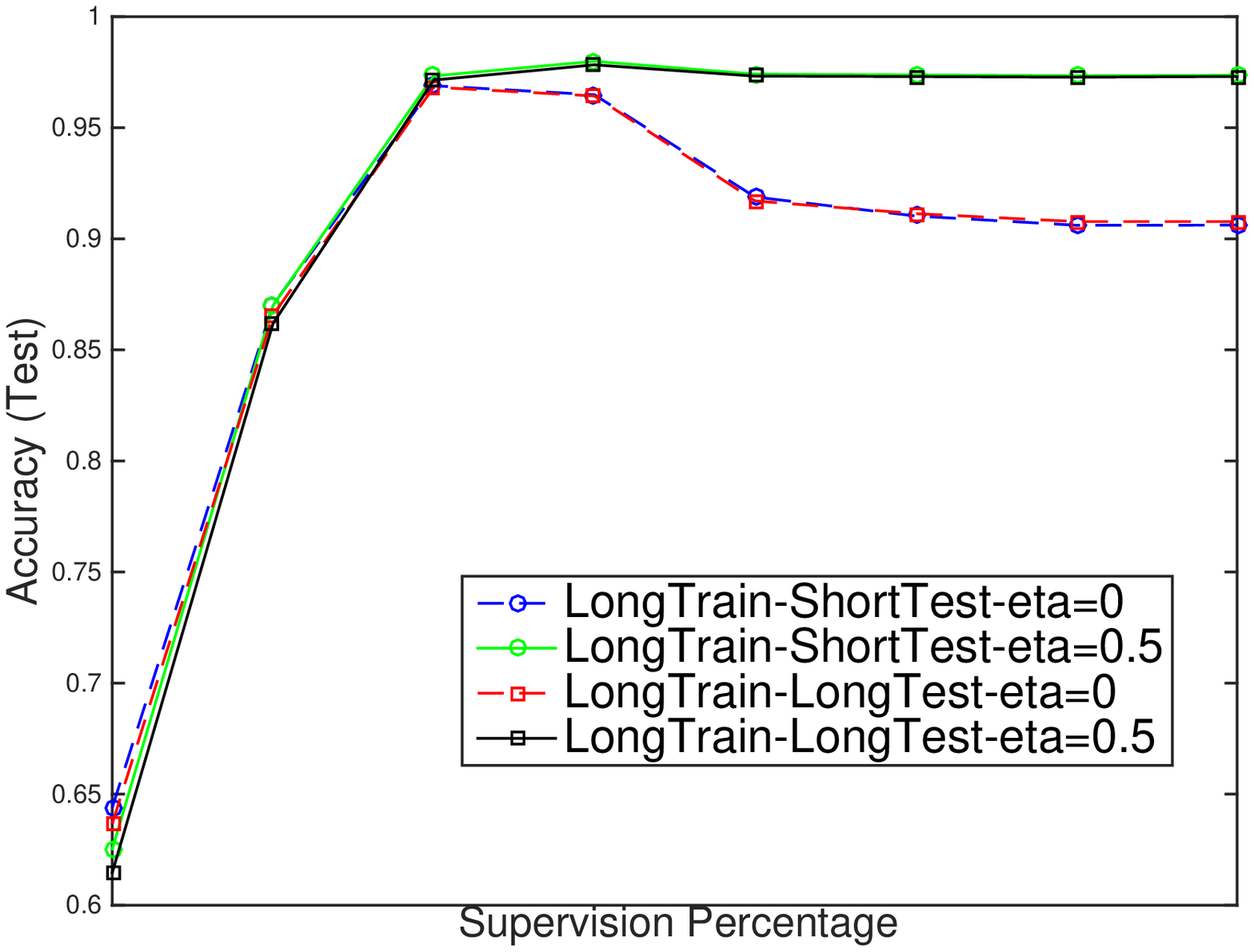} & \includegraphics[scale=0.4]{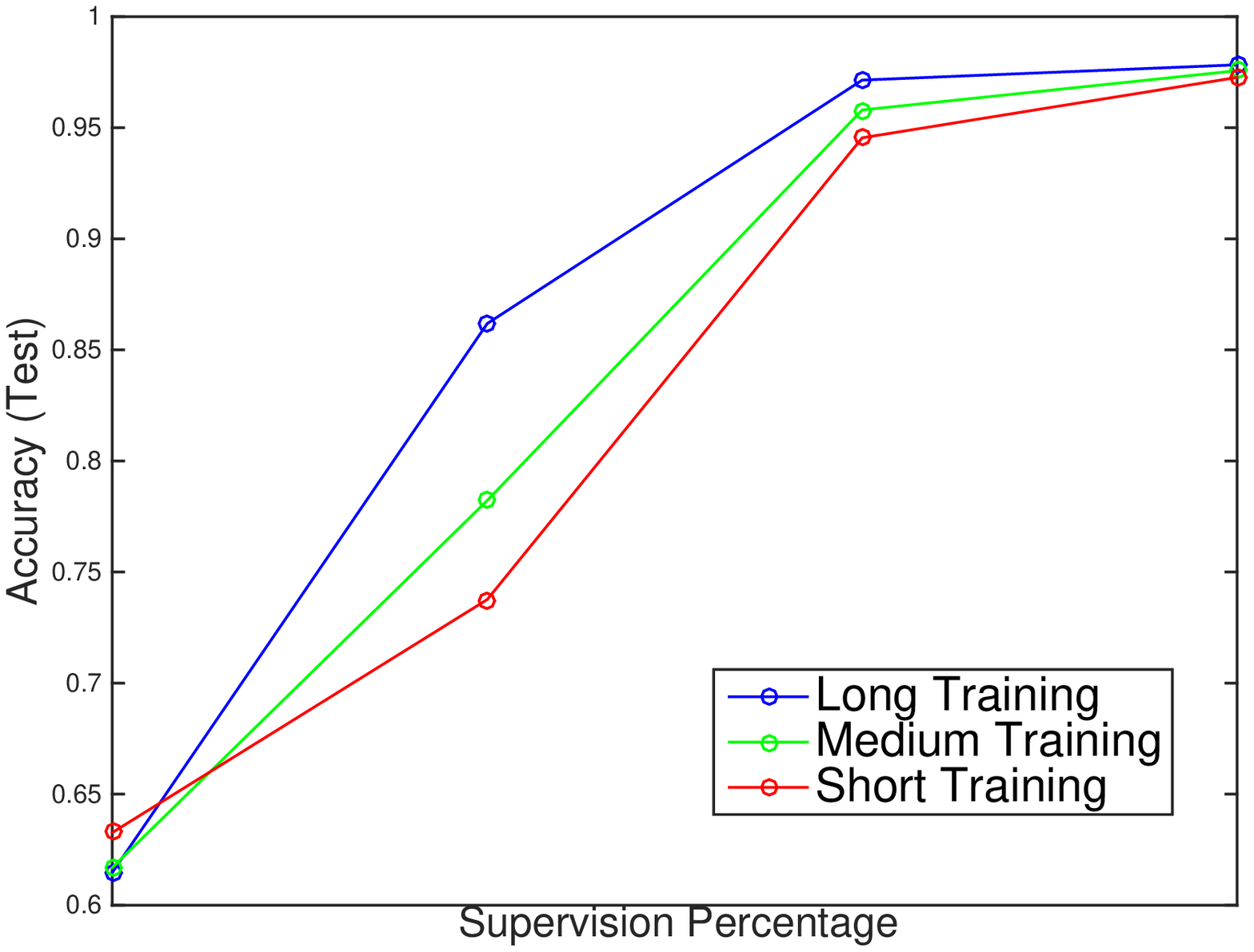}  \\
(c) Long Training &  (d) Supervision in $\left\{1,\,5 (0.1\%),\, 30 (0.5\%),\, 60(1\%)\right\}$\\
\end{tabular}
}
\caption{Accuracy vs. Supervision Percentage for the 0vs1 MNIST classification task.
Number of supervisions in $\left\{1,\,5 (0.1\%),\, 30 (0.5\%),\, 60(1\%),\, 300(5\%),\, 600(10\%),\, 2900 (50\%),\, 5900 (100\%)\right\}$.
Different sequences used for Training.
(a) Short Training Sequence for Shirt and Long Test Sequences and $\eta = 0, 0.5$. 
(b) Medium Training Sequence for Shirt and Long Test Sequences and $\eta = 0, 0.5$.
(c) Long Training Sequence for Shirt and Long Test Sequences and $\eta = 0, 0.5$.
(d) Trend with few supervision for the 3 lengths of the Training Sequences, Long Test Sequence and $\eta = 0.5$.\label{mixlong}}
\end{figure}

We test our trained model also on a simple random permutation of the original test set.
This with the aim to see if it is able to well predict random sorted data once it has been trained with sequential data.
The hope is that the training is enough to learn a solid model for the temporal links.

\newpage
\subsubsection{MNIST Video }

\begin{figure}[H]
\begin{adjustbox}{max size={\textwidth}}{}
 \begin{subfigure}{.23\textwidth}
 \centering
  \includegraphics[scale=0.2]{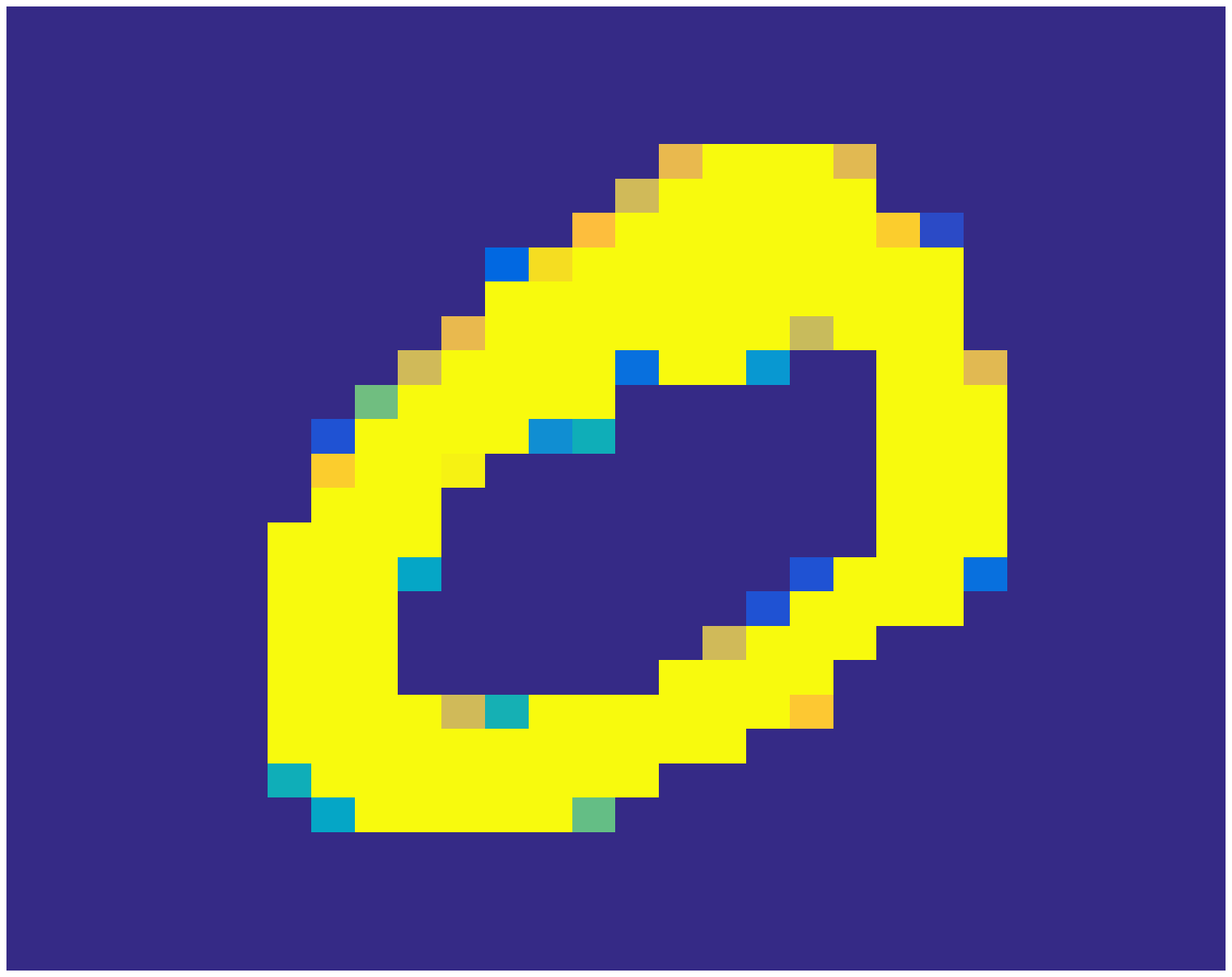}
  \caption{}\label{sample1}
  \end{subfigure}
   \begin{subfigure}{.23\textwidth}
 \centering
  \includegraphics[scale=0.2]{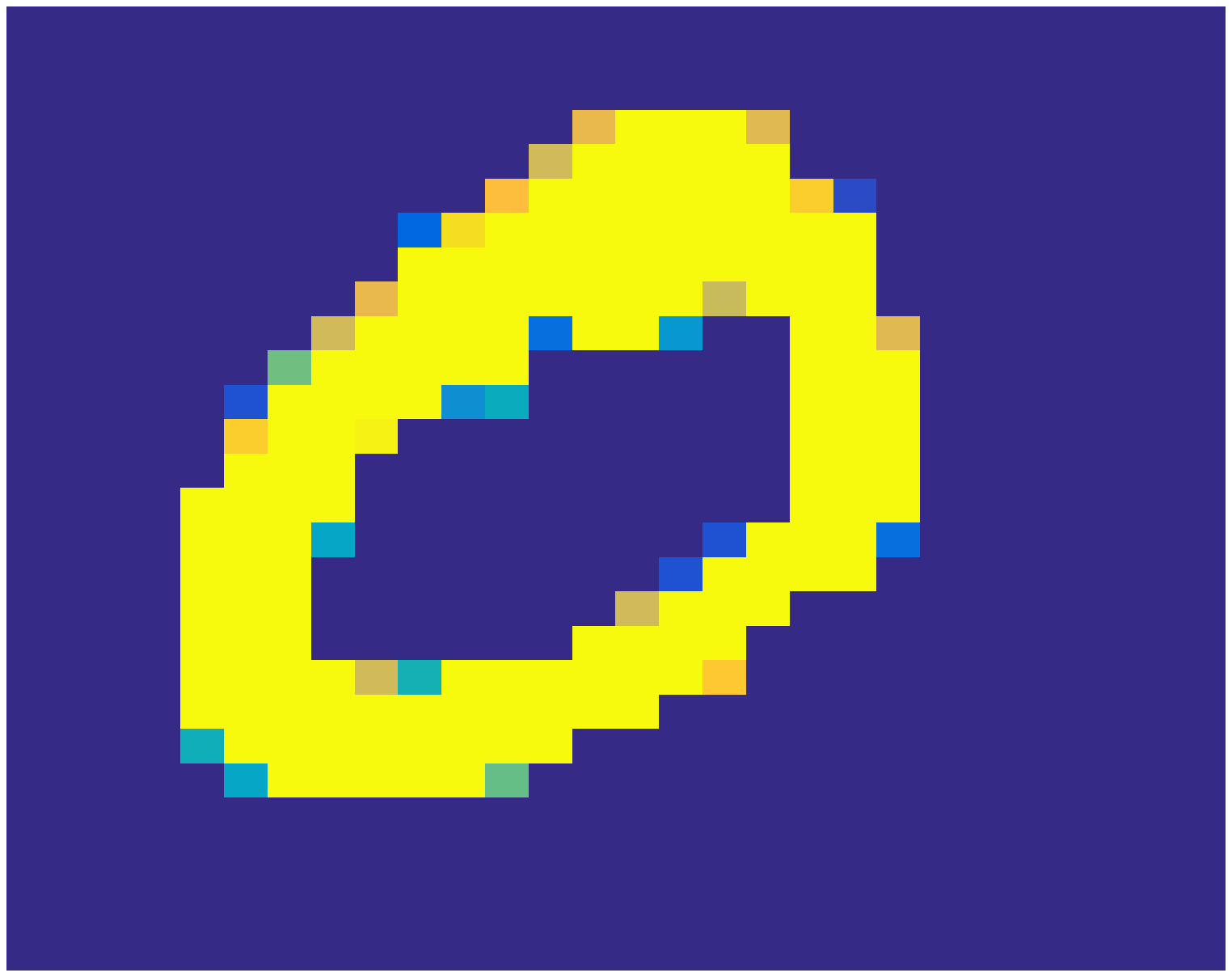}
  \caption{}\label{sample2}
  \end{subfigure} 
  \begin{subfigure}{.23\textwidth}
 \centering
  \includegraphics[scale=0.2]{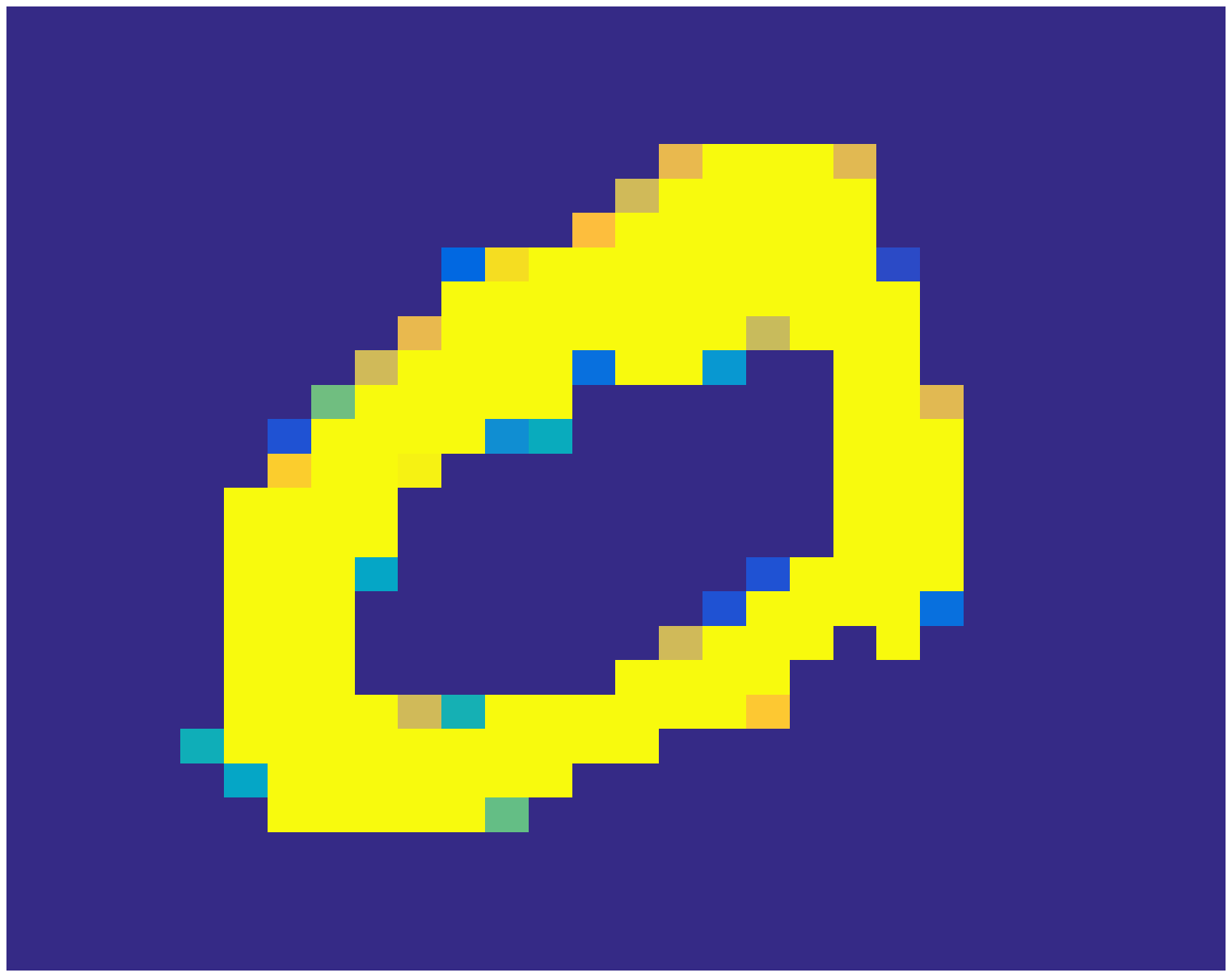}
  \caption{}\label{sample3}
  \end{subfigure}
    \begin{subfigure}{.23\textwidth}
 \centering
  \includegraphics[scale=0.2]{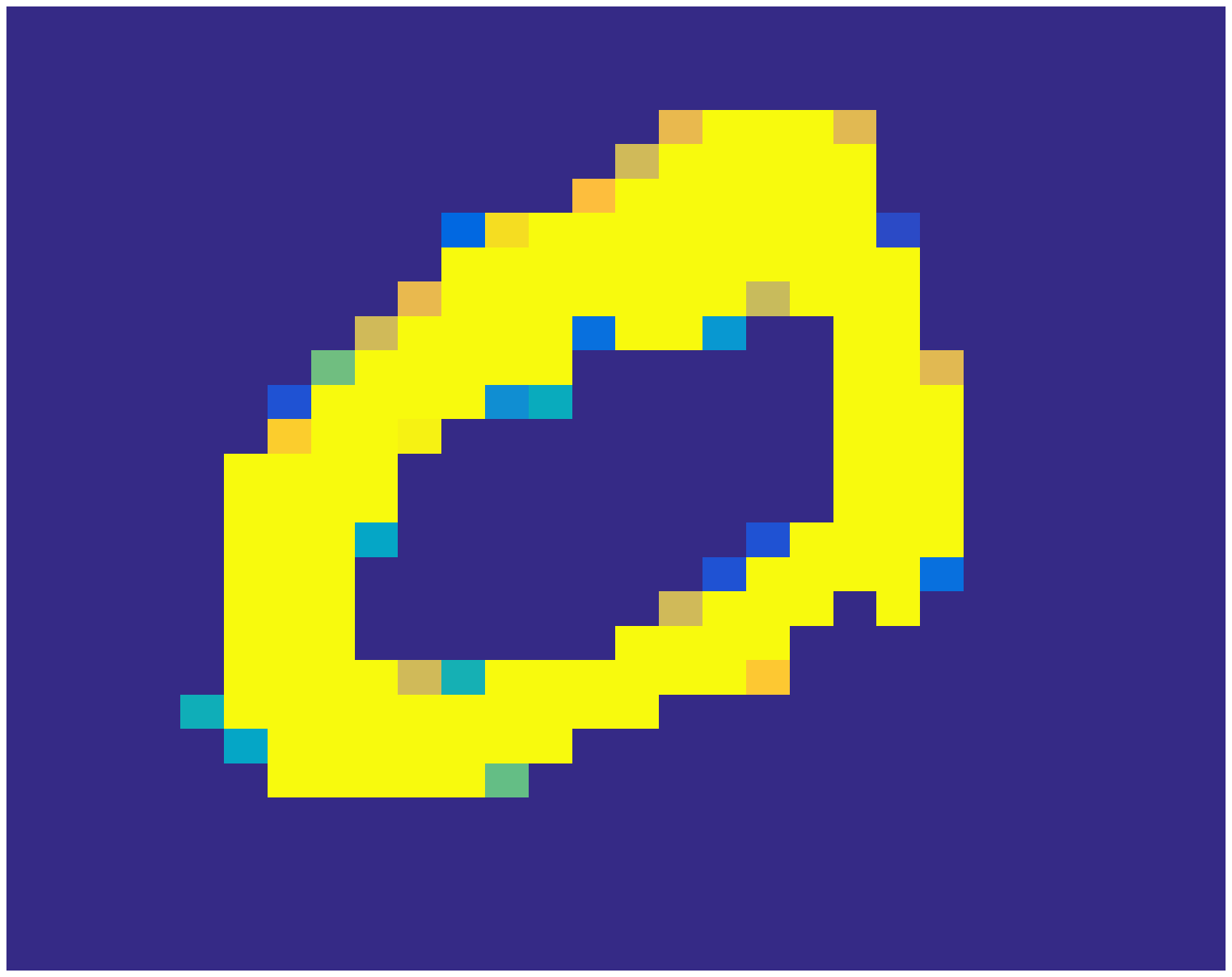}
  \caption{}\label{sample4}
  \end{subfigure} 
  \end{adjustbox}
  \caption{Examples of consecutive training images generated by random transformations: (a) is the original image of the digit 0, (b) an right horizontal translation of 1 pixel ($|| x_2 - x_1|| = 10.93$), (c) clockwise rotation of $4^¡$ ($|| x_3 - x_2|| = 8.51$), (d) left horizontal translation of 1 pixel ($|| x_4 - x_3|| = 5.53$).}\label{samples}
\end{figure}

\begin{figure}[H]
\begin{adjustbox}{max size={\textwidth}}{}
 \begin{tabular}{cc}
  \includegraphics[scale=0.4]{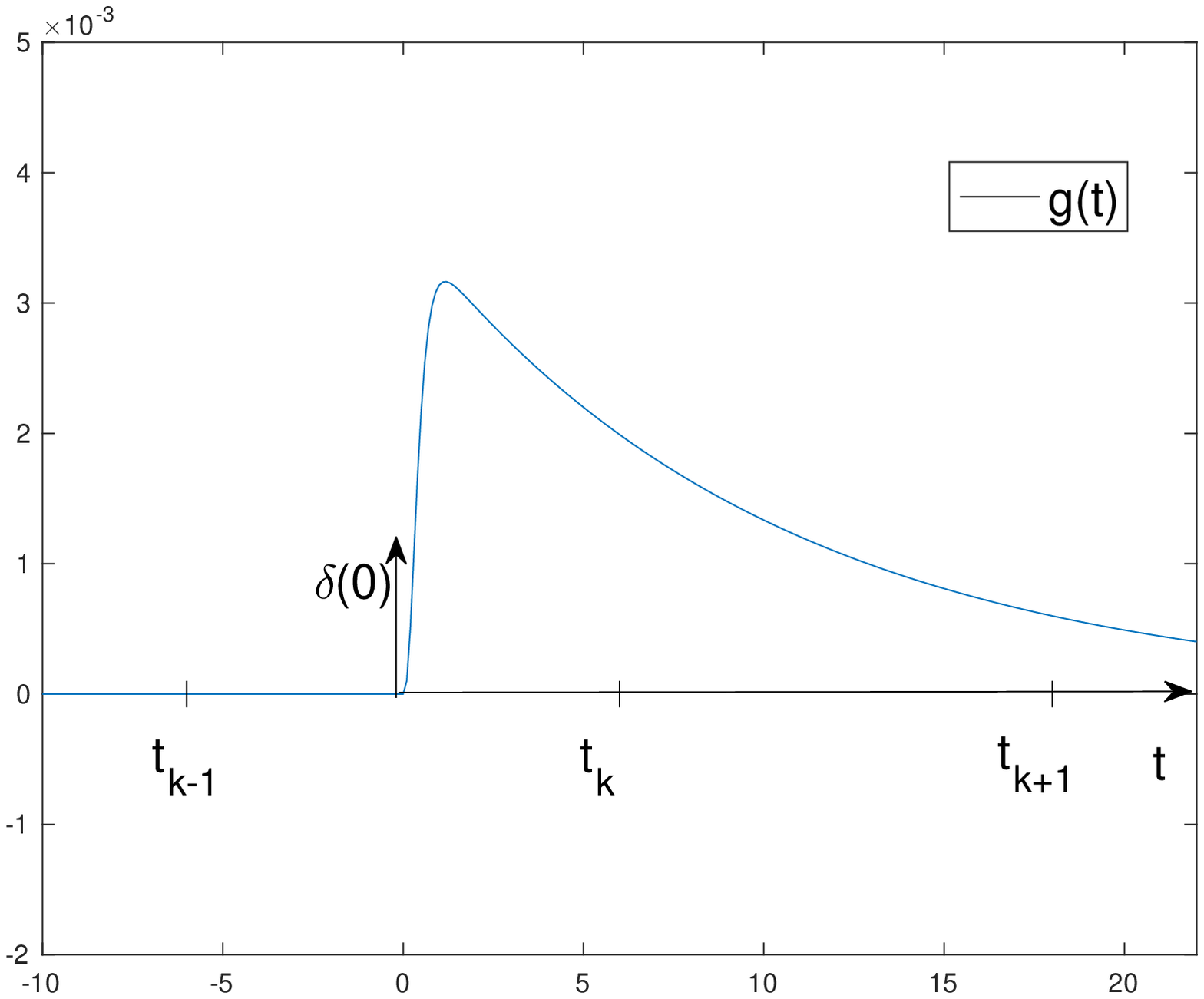} &
  \includegraphics[scale=0.41]{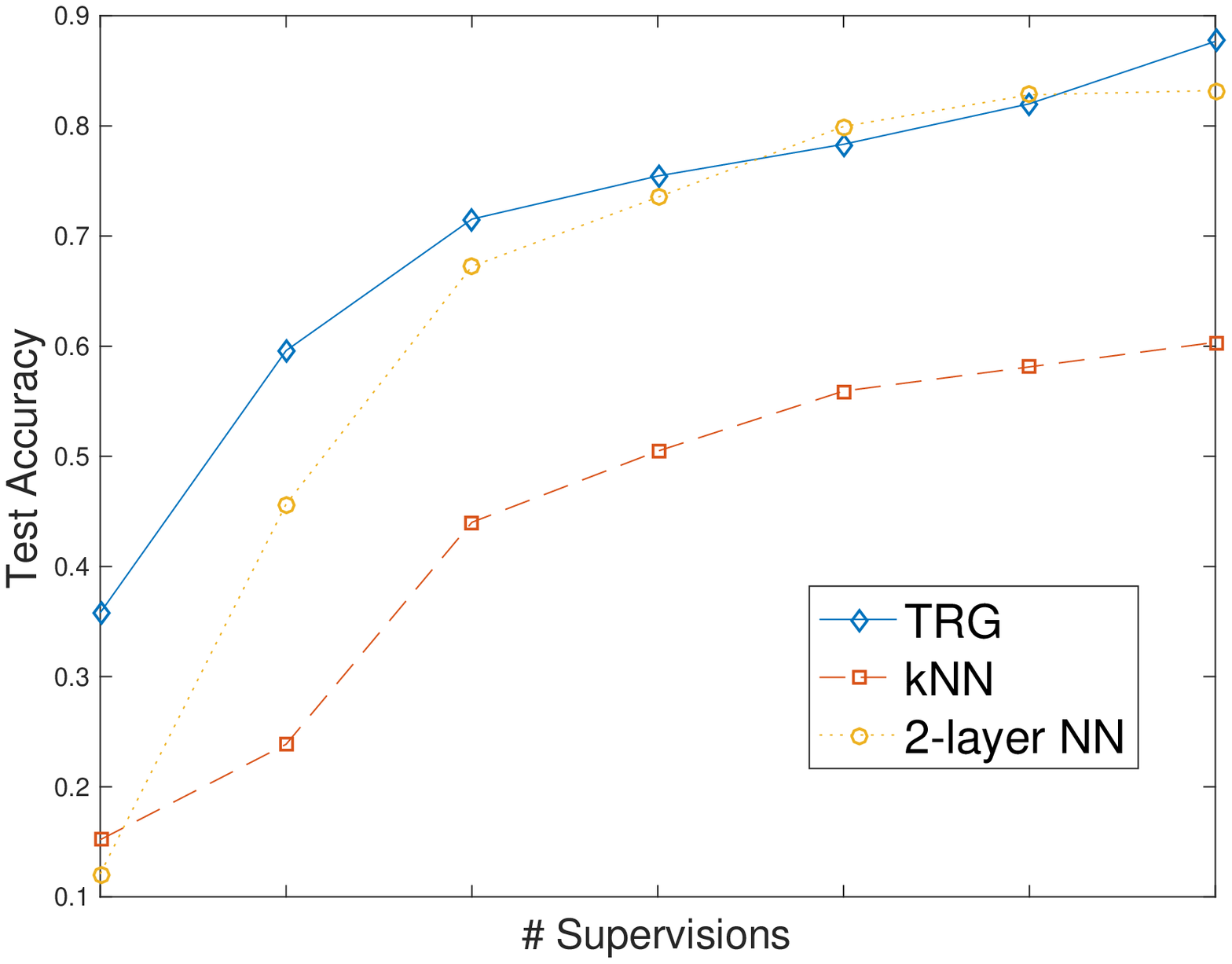}\\
  (a) & (b) \\
  \end{tabular}
  \end{adjustbox}
  \caption{(a) Impulsive Response and relative samples distribution; (b) Results of \RefTab{res}.}\label{mnvres}
\end{figure}

In this section we test our algorithm on a video created from the MNIST dataset. We randomly select $5240$ points from the original training data (equally distributed per class) and generate a sequence by operating $60$ consecutive small transformations on each sample, so as to obtain a video (sampled at 30 frames per second) in which each samples appear for 2 seconds. The sequence of possible transformations is created by randomly select among: rotating (angle $\{-20:4:20\}$), translating (maximum 3 pixels in each direction), scaling (only few pixels bigger or smaller), blurring (Gaussian filter with parameter $\{0.25:0.25:1\}$). Since the image are $28 \times 28$ pixels and we feed our algorithm with the 1-D vector of grey levels, a  small transformation produce a quite different descriptor. We obtain a training sequence of $32400$ samples. An example of few consecutive generated images is showed in \RefFig{samples}. The changes are not so clear at view, but is relevant at pixels level, as indicated from the reported Euclidean distances. We train our graph with the sequence varying the number of supervisions. The test is carried out on the original test set by a simple Nearest Neighbor search among the nodes of the trained graph. We report the comparison with standard algorithm as Nearest Neighbor and a 2-layer Artificial Neural Network. For both of them we exploit the relative Matlab functions (\texttt{fitcknn} and \texttt{patternnet}) which is high performing. We set $k=10$ (which we found to achieve the best test performance in $\{1,\,5,\,10,\,25,\, 50,\,100,\,1000\}$) in the first one and $300$ hidden units in the second one (as reported in~\cite{lecunn98} for the same architecture). To set the parameters of our algorithm we choose $\t = 10$ (which is arbitrarily since we have to adapt $\q$ consequently ) and a fourth order system with roots $r_1=-0.1,\, r_2= -6,\,r_3=  -6.5,\,r_4=  -7.4$, which give the impulsive response of \RefFig{MNIR}. The idea is that an impulse should propagate the information over few incoming samples before to goes to $0$. We fix $\ee=3$, which allow to store in the graph about $8000$ nodes ($\approx 24\%$), however its role is clear since more nodes improve the prediction performance but affect the computational cost. The remaining parameters are chosen in a validation phase on the first $10\%$ of the sequence (a $90\%$ of the portion is used for training and the remainder for the test). A small regularization parameter $\l$ amplifies the response (improving the performance) but lead to divergence (because of an accumulation of the delay of the impulsive response). A good balance is found at $\l = 0.01$. The others $\sigma = 3$, $\q = 12$, and $\rho = 5$ are chosen so as to achieve the best accuracy. The balance between temporal and spatial contribution is tested for $\eta = \{0,\,0.5\}$ but does not produce notable differences and we report only $\eta = 0.5$. The results are reported in \RefTab{res} and \RefFig{mnvres} decreasing the number of supervisions. We report also the computational time since our algorithm used also the unsupervised sample and is slower when we have few supervised points to feed to ANN or kNN. Each test is carried out 3 times with supervisions on different samples (randomly chosen) and accuracy is averaged.

\begin{table}[h]
\begin{center}
\begin{tabu}{cc||cc|cc|cc}
\multicolumn{2}{c||}{\rule{0pt}{2ex} Supervisions} & \multicolumn{2}{c|}{\rule{0pt}{2ex} TRG} & \multicolumn{2}{c|}{\rule{0pt}{2ex} kNN ($k=10$)} & \multicolumn{2}{c}{\rule{0pt}{2ex} 2-layers NN (HU$= 300$)} \\
\% & \# & Accuracy & Time (sec.) & Accuracy & Time (sec.) & Accuracy & Time (sec.) \\
\tabucline[1pt]{-}
100 & 32400 &0.87    &  320.05   &  0.60  &  98.11  &    0.83  &  640.01 \\
  \rowcolor{gray!10}50 & 16200 &  0.82 &     414.35  &    0.58 &    62.43 &     0.82 &   373.59 \\ 
    \rowcolor{gray!20}25 & 8100 & 0.78 &     431.24 &    0.55 &    34.70  &    0.79 &    204.87 \\
    10 & 3240 & 0.75  &    421.67   &  0.50 &   13.64    & 0.73 &  73.98   \\
  \rowcolor{gray!10}5 & 1620 & 0.71 &     329.42   &  0.44 &   4.81   &    0.67 &  17.09  \\
    \rowcolor{gray!20}1 & 320 & 0.59 &    306.91   &  0.23 &   1.26 &    0.45 &    4.05 \\
    0.1 & 30 & 0.35 &       360.95  &   0.15 &    0.44 &   0.12 &   2.14 \\
\end{tabu}
\end{center}
\caption{Classification accuracy on the MNIST test set for our algorithm (TRG), the k-Nearest Neighbors (kNN) and a 2-layer Artificial Neural Network (2-layers NN) decreasing the number of supervisions.}
\label{res}
\end{table}


\newpage

\bibliographystyle{plain}
\bibliography{references}

\begin{thebibliography}{1}

\bibitem{olap}
Salvatore Frandina, Marco Lippi, Marco Maggini, and Stefano Melacci.
\newblock In {\em ICANN}.

\bibitem{Gold}
Andrew~B. Goldberg, Ming Li, and Xiaojin Zhu.
\newblock Online manifold regularization: A new learning setting and empirical
  study.
\newblock In {\em ECML/PKDD (1)}, pages 393--407, 2008.

\bibitem{report}
Marco Gori, Marco Maggini, and Alessandro Rossi.
\newblock The principle of cognitive action , preliminary experimental
  analysis,
  \url{https://drive.google.com/open?id=0B7_Mj3qkmLd9fjZYQUJwcEdWT3Q2aTFMck1iTlNJVkMwY1cxT3I5Z1E2NXlVTWxjWG1KLWs}.
\newblock Technical report, University of Siena, DIISM, 2015.

\bibitem{rws}
Marco Gori, Marco Maggini, and Alessandro Rossi.
\newblock Random walks,
  \url{https://drive.google.com/open?id=0B7_Mj3qkmLd9UWhTdUE1TmZHMVk}.
\newblock Technical report, University of Siena, DIISM, 2015.

\bibitem{papini}
Duccio Papini.
\newblock Variational laws of cognition.
\newblock Technical report, 2015.

\end{thebibliography}

\end{document}